\documentclass[pdflatex,sn-mathphys-ay]{sn-jnl}

\usepackage[utf8]{inputenc} 
\usepackage[T1]{fontenc}    
\usepackage{url}            
\usepackage{nicefrac}       
\usepackage[table]{xcolor}         
\usepackage{nicematrix}
\usepackage{enumitem}
\usepackage{graphicx}
\usepackage{multirow}
\usepackage{parskip}
\usepackage{amsmath, amssymb, amsfonts}
\usepackage{amsthm, thmtools, thm-restate}
\usepackage{booktabs}%
\usepackage{subcaption}
\usepackage[hang,flushmargin]{footmisc}
\usepackage[frozencache,cachedir=minted-cache]{minted}
\usepackage{algorithm}

\usepackage[title]{appendix}%

\begin{document}

\title{Few-shot Classification as Multi-instance Verification: Effective Backbone-agnostic Transfer across Domains}


\author*{\fnm{Xin} \sur{Xu}}\email{xinxu75@gmail.com}
\author{\fnm{Eibe} \sur{Frank}}\email{eibe@waikato.ac.nz}
\author{\fnm{Geoffrey} \sur{Holmes}}\email{geoff@waikato.ac.nz}

\affil{\orgdiv{Department of Computer Science}, \orgname{University of Waikato}, \orgaddress{\city{Hamilton}, \country{New Zealand}}}

\abstract{We investigate cross-domain few-shot learning under the constraint that fine-tuning of backbones (i.e., feature extractors) is impossible or infeasible---a scenario that is increasingly common in practical use cases---and show how representing few-shot classification as a series of multiple instance verification (MIV) tasks enables competitive classification performance using the static embeddings produced by frozen, ``black-box'' backbones. More specifically, based on this novel representation, we introduce a new approach to few-shot domain adaptation named the ``MIV-head'': a classification head that is backbone-agnostic and computationally efficient. We show that the core components of the MIV-head, when trained on few-shot data from a target domain, collectively yield strong performance on test data from that domain. Importantly, they do so \emph{without} fine-tuning the backbone, requiring only the ``meta-testing'' phase. In experiments under various settings on an extension of the Meta-Dataset benchmark for cross-domain few-shot image classification, using representative off-the-shelf convolutional neural network and vision transformer backbones pretrained on ImageNet1K, we demonstrate that the MIV-head achieves highly competitive accuracy when compared to state-of-the-art ``adapter'' (or ``partial fine-tuning'') methods applied to the same backbones, while incurring substantially lower adaptation cost. Moreover, compared to our approach, well-known ``classification head'' approaches achieve a substantially lower level of accuracy. We share our code at: \url{https://github.com/xxweka/MIV-head}.
}

\keywords{Cross-Domain Few-Shot Learning, Multiple Instance Learning, Image Classification, Transfer Learning, Backbone-Agnostic Domain Adaptation}




\maketitle

\section{Introduction} \label{sec:intro}

With the emerging popularity of cross-domain few-shot learning (CDFSL), typically in image classification~\citep{closerfewshot19, md20, mdmore20, mdvtab21, closerlook23}, numerous approaches have been proposed to tackle the key challenge of how to effectively transfer knowledge from the 
source domain(s) to yield an accurate classifier for the target domain using few-shot data. A setup that is particularly relevant in practical applications is to use a generic dataset such as ImageNet1K (ILSVRC-2012 \citep{imagenet_cvpr09, ILSVRC15}, henceforth referred to as ILSVRC) as the source domain, and a small set of labeled data available for the target domain (henceforth called the ``few-shot support set''), to train a classifier. The task is to learn to classify new examples in the target domain,  called ``queries'', based on the information in the labeled support set and the source domain. In practice, the latter is often given in the form of a publicly available feature extractor (or ``backbone'') that has been pretrained on ImageNet1K. This is the scenario we consider in this paper.\footnote{We focus on the ``cross-domain'' setting in FSL, \emph{not} the ``in-domain'' scenario used by some \citep{matchingnet16}, where the source and target domains comprise different sets of classes from the same dataset.}

\begin{figure}[t]
    \centering
    \includegraphics[width=\linewidth]{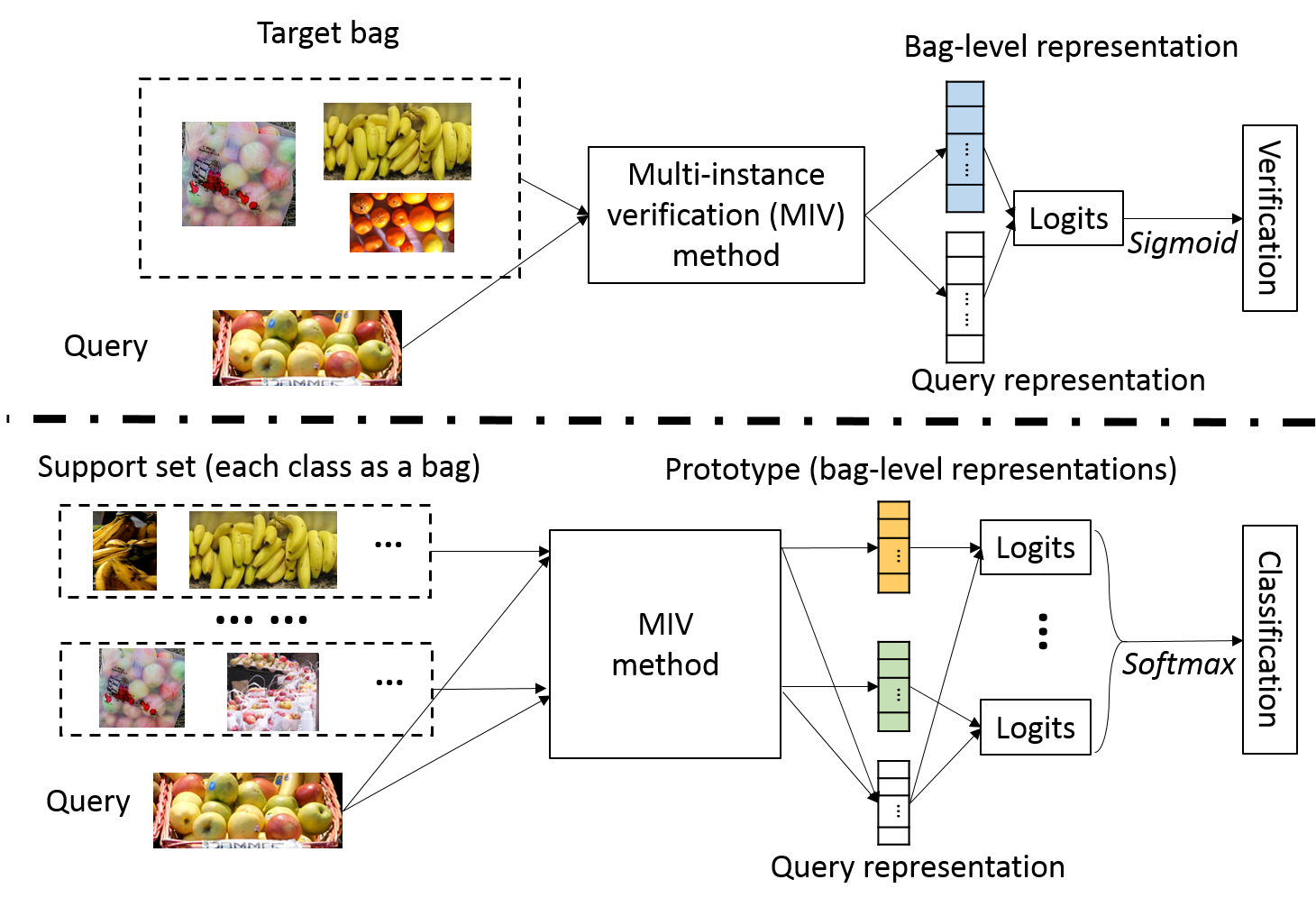}
     \caption{Few-shot classification (FSC) represented as a series of multi-instance verification (MIV) tasks. The upper panel illustrates a standard MIV task, where a target bag and a query are paired for binary classification. The lower panel shows that FSC can be represented as a series of MIV tasks: the support set forms a set of target bags, with each class as a bag. All the bags (or classes) are paired with a query, forming a series of MIV tasks, to perform multi-class classification.} \label{fig:task_fscmiv}
\end{figure}

The state-of-the-art (SOTA) in the CDFSL literature has been predominantly achieved by approaches that fine-tune the backbone, either fully or partially: these approaches modify backbone model weights and/or architecture based on the new target domain. One such approach is called an ``adapter'' approach. It adds adapters with learnable parameters \emph{inside} the backbone and thus is able to adjust feature vectors for domain adaptation \citep{flute21, tsa22, ett22, adaptformer22, case22, fit23}. According to recent comparisons~\citep{closerlook23, vitbaseline24}, two of the best-performing CDFSL methods are adapter methods, namely ``Task-Specific Adapters'' (TSA,~\citealp{tsa22}) and ``efficient Transformer Tuning'' (eTT,~\citealp{ett22}). These two methods are designed for different families of backbones: convolutional neural networks (CNNs) and vision transformers (ViTs), respectively. Impressive classification accuracy has been obtained by them, exceeding that of other domain-adaptation methods---excluding carefully configured full fine-tuning methods like PMF \citep{pmf22}, which may be impractical because they are computationally costly.

The motivation for the work presented in this paper is that adapter methods and, more generally, methods that perform some form of ``backbone fine-tuning'',  have drawbacks in practical applications. First, they are unable to cope with situations where the backbone is \emph{non-modifiable} or even \emph{unknown}. This is a practically relevant constraint, important enough to merit the attention of the research community for several reasons: (1) there exists a plethora of off-the-shelf, pretrained ``foundation models'' that serve their users in a ``black-box'' manner---with frozen weights and architecture---producing outputs through cloud-based APIs; (2) the prevalence of vector databases creates demand for methods that can operate on static embeddings without details of underlying feature extractors; (3) the surging popularity of ``large models'' makes backbones increasingly more difficult to fine-tune. Second, ``fine-tuning'' approaches are inflexible regarding the choice of backbone, and their application to commonly used backbones can sometimes be challenging: we found that adapter methods frequently lead to out-of-memory (OOM) errors (\emph{cf.} Section~\ref{sec:experiments}). 
This inflexibility is particularly problematic considering the fundamental importance of the choice of backbone \citep{embed_linear20, pmf22} and off-the-shelf models available as candidate backbones \citep{closerlook23}. Finally, the fine-tuning processes are computationally expensive: they need to pass through the backbone, both forward and backward, at \emph{every} training step to update the adapters' or backbones' parameters.

Our goal is to handle black-box feature extractors in a lightweight, versatile, backbone-agnostic fashion, \emph{and} to achieve accuracy comparable to the SOTA established by adapter methods. This is a challenging problem. When traditional classification heads, typically with learnable parameters, are applied to the output of a frozen backbone, they fail to achieve competitive performance close to the SOTA, as we demonstrate in Appendix~\ref{app:fh}, despite being API-friendly and computationally efficient. This is not surprising: the embeddings produced by the backbone are not adapted to the target domain, and hence their relationship to the (unseen) class labels may be ambiguous and potentially unreliable. To alleviate this issue, unlike existing methods that implicitly treat embeddings of all support samples as equally connected to their class labels, we explicitly model the support set pertaining to a class as a bag of ``noisy'' instances and let them compete for a bag representation given the query. 

Inspired by the ``multiple instance verification'' (MIV) problem \citep{miv25}, which involves comparing a query instance with a bag of heterogeneous target instances of unknown relevance (henceforth ``target bag'') to verify if the target bag contains instance(s) of the same class as the query, we represent a few-shot classification (FSC) problem as a series of MIV tasks, and tackle this problem within the paradigm of ``verification with noise''. Figure \ref{fig:task_fscmiv} illustrates MIV and its application to FSC. The essential observation is that the data pertaining to a class in the support set can be viewed as a target bag, in which the ambiguous relevance of the bag's instances to the class label is induced by low-quality embeddings produced by black-box backbones. Hence, classification of a query can be formulated as a series of MIV tasks, with one task per bag (or class). 

In the new ``classification head'' approach we propose in this paper, the bag representations are analogous to ``prototypes'' \citep{protonet17} in FSC. The query's feature vector is paired with each prototype through a Siamese network to compute a similarity metric yielding the logits of the corresponding classes, to which a $softmax$ function is applied across all classes. The resulting method, named ``MIV-head'', can be used like a classification head---i.e., in conjunction with black-box backbones that expose multi-block, patch-level feature maps\footnote{Throughout this paper, we assume the underlying model API of the backbone provides access to such representations, as is common when using foundation models such as DINO-ViT \citep{dino21}.}---and achieves SOTA accuracy in our experiments. To realize the MIV-head, we use a variant of ``cross-attention pooling'' (CAP)---the MIV solution proposed by~\cite{miv25}---as a key component (Component 2). We also introduce two additional components to address the challenges brought by (1) patch-level feature maps retrieved from a backbone's API, and (2) inadequate embeddings from a single block of the backbone. More specifically, we propose a ``pooling by attention'' mechanism on patch-level embeddings (Component 1) and a strategy to extract features from multiple blocks of the backbone (Component 3). 
These three components form the core of the MIV-head, see Section~\ref{sec:arch}. We emphasize that, while each individual component in the MIV-head may not be new in the literature \citep{miv25, fusion20}, they have not been applied collectively to CDFSL. Viewing an FSC problem as a set of MIV tasks makes this possible.

\begin{figure}[t]
    \centering
     \begin{subfigure}[t]{0.66\linewidth}
        \begin{subfigure}[b]{0.49\linewidth}
            \includegraphics[width=\textwidth]{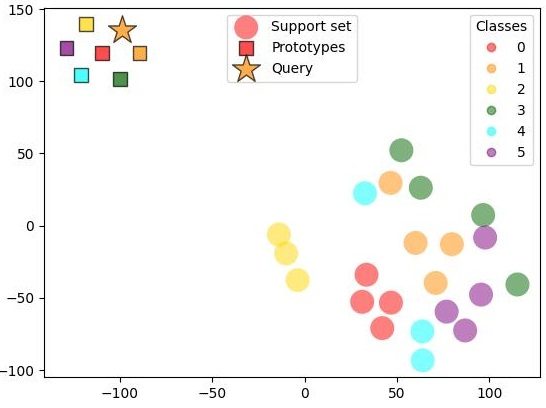}
            \vspace*{-.5cm}
            \caption*{\footnotesize Last block of backbone}
        \end{subfigure}
        \begin{subfigure}[b]{0.49\linewidth}
            \includegraphics[width=\textwidth]{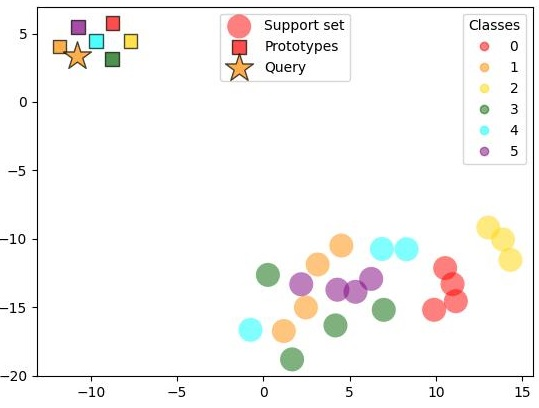}
            \vspace*{-.5cm}
            \caption*{\footnotesize Second-to-last block of backbone}
        \end{subfigure}
        \vspace*{-2mm}
        \caption{MIV-head (Ours)}\label{fig:tsne_miv_a1}
     \end{subfigure}
     \begin{subfigure}[t]{0.33\linewidth}
     \vspace*{-3.8cm}
     \includegraphics[width=\textwidth]{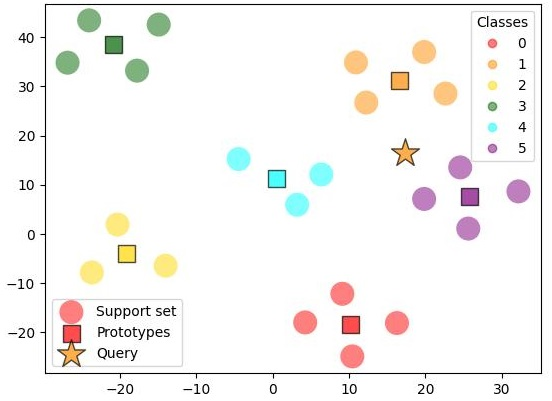}
     \vspace*{0mm}
     \caption{TSA}\label{fig:tsne_tsa_a1}
     \end{subfigure}
     \caption{Embedding visualizations of a support set (circles) from the Aircraft dataset, obtained with t-SNE. All embeddings are colored according to their class labels, as specified by the legend. The left and right panels of Figure~\ref{fig:tsne_miv_a1} are visualizations of embeddings from the last and second-to-last block of a frozen, off-the-shelf ResNet-50 backbone, respectively, and also show the prototypes (squares) and the query (star) after processing by the MIV-head. The third panel (i.e., Figure~\ref{fig:tsne_tsa_a1}) shows the embeddings of the support set, the prototypes and the query from the same backbone when applying trained TSA adapters. }\label{fig:tsne_full}
\end{figure}

To illustrate the challenges imposed by a frozen, black-box backbone, Figure~\ref{fig:tsne_full} shows t-SNE visualizations of embeddings of a support set sampled from the Aircraft dataset \citep{aircraft13}, based on a ResNet-50 backbone. After training adapters for the backbone, embeddings of the support samples created by TSA (Figure~\ref{fig:tsne_tsa_a1}) are of high quality: well clustered w.r.t. their ground-truth class labels. Therefore, centroids of clusters (or classes) can be used as prototypes to classify the query. In contrast, ``classification head'' approaches such as the MIV-head are faced with low-quality embeddings of the support set retrieved from the black-box backbone---as illustrated by Figure~\ref{fig:tsne_miv_a1}: they are less well clustered w.r.t. their ground-truth class labels, irrespective of whether embeddings are obtained from the last or second-to-last block of the backbone (\emph{cf.} left and right panels).

This example illustrates that embeddings from generic backbones may be problematic and yield low classification accuracy. As shown in Figure~\ref{fig:tsne_miv_a1}, our approach projects all prototypes near the query (as opposed to the support set) and frequently succeeds in pulling the prototype of the ``query class'' \emph{nearest}, thereby leading to improved classification despite the low quality of the support-set embeddings. Appendix~\ref{app:tsne} includes more t-SNE visualizations. 

We demonstrate the benefits of our approach using experiments on an extended version of the well-known Meta-Dataset (MD) benchmark \citep{md20, mdvtab21}, based on off-the-shelf CNN and ViT backbones pretrained on ILSVRC by supervised and self-supervised pre-training. Experimenting under various settings, we show that the MIV-head achieves similar, or higher, accuracy compared to the SOTA adapter methods, TSA and eTT, applied to the same backbone, while incurring substantially lower adaptation cost (and latency). We draw similar conclusions when comparing the MIV-head to the recent fine-tuning method ``LN-Tune'' \citep{vitbaseline24}, which has been shown to be among the best-performing. We also find that established ``classification head'' approaches perform substantially worse in terms of accuracy, suggesting that MIV-head may be the first backbone-agnostic method capable of achieving this level of performance. Furthermore, a comprehensive ablation study (in Section~\ref{sec:ablation}) demonstrates that all core components of the MIV-head collectively contribute to its superior performance, empirically validating our design.

\begin{table}[t]
\centering
\renewcommand{\arraystretch}{1.5}
\begin{tabular}{lll}
\textbf{Regarding ...} & \multicolumn{1}{c}{\textbf{Previous literature}} & \multicolumn{1}{c}{\textbf{Novelty of our work}} \\ \toprule
\rowcolor{lightgray!50!white}Backbone& Fine-tuning backbone to & NEW: No modification of backbone model\\
\rowcolor{lightgray!50!white}models&achieve SOTA accuracy&weights or architecture, while being competitive\\
&1) Classification problem&1) ``Verification with noise'' (MIV) problem\\
Representations &2) Support samples are&2) Same-class support samples are treated as bag\\
and solution& \ \ \ \ treated as ``noise-free''& \ \ \ \ of ``noisy'' instances with unknown relevance\\
design& \ \ \ \ \& with equal relevance & 3) NEW: Introduction of competitions to\\
& &\ \ \ \ ``soft-select'' embeddings in the design\\
\rowcolor{lightgray!50!white}Pooling of patch- & Average-pooling (or pooled&NEW: ``pooling by attention'', of patches and\\
\rowcolor{lightgray!50!white}level embeddings&by ``[CLS]''-token in ViT)&w.r.t. spatial dimensions of patches, hierarchically\\
&1) Working jointly with&1) NEW: Working on top of frozen, black-box\\
Cross-attention&\ \ \ \ backbone fine-tuning&\ \ \ \ backbones in CDFSL application\\
pooling (CAP) &2) No ``in-attention skip-&2) NEW: ``In-attention skip-connection'' as an\\
&\ \ \ \ connection''&\ \ \ \ additional module in CAP\\
\rowcolor{lightgray!50!white}Aggregation from&Non-CDFSL usage, usually&CDFSL usage, aggregating logits from multiple\\
\rowcolor{lightgray!50!white}mid-level blocks&aggregating embeddings&blocks \& based on other MIV-head components\\
\bottomrule
\end{tabular}
\caption{Summary of the innovations in this paper, compared to existing practices from the prior literature.}\label{tab:novelty}
\end{table}

Our main contributions are threefold: First, we tackle CDFSL when ``backbone fine-tuning'' is impossible and, by using black-box backbones, achieve accuracy competitive with, or exceeding, SOTA fine-tuning methods. Second, we propose a new representation of, and solution to, CDFSL using a ``verification with noise'' paradigm (MIV) that addresses challenges brought by black-box backbones. Third, we introduce a backbone-agnostic method akin to a classification head that incurs lower adaptation cost than fine-tuning approaches; it comprises three core components that collectively solve the problem.

To further articulate our innovations and differentiate this paper from the previous literature, Table~\ref{tab:novelty} contrasts our work (rightmost column) with existing practices in the literature (middle column), under five themes (leftmost column). We believe there is particular novelty in two aspects: (1) the backbone-agnostic nature of our approach (the first theme in Table~\ref{tab:novelty}), and (2) the introduction of competitions in our design (the second theme, more precisely point 3, in Table~\ref{tab:novelty}). The former deals with black-box, frozen backbones, which induce the low quality of embeddings, or ``noise'', when backbones are used in a new domain---justifying our representation of a CDFSL problem through the lens of MIV. The latter is the guiding principle of our design for the MIV-head's three core components.

In addition, while our work is strongly motivated by~\citep{miv25}, there are two key distinctions---as shown in the fourth theme of Table~\ref{tab:novelty}: One, \cite{miv25} used CAP in conjunction with ``backbone finetuning'', while we adopt CAP on top of frozen, black-box backbones in backbone-agnostic CDFSL. Two, the CAP component of the MIV-head differs from the one used by~\cite{miv25} because the original CAP does not have the novel ``in-attention skip-connection'' mechanism. We also highlight those differences in Section~\ref{sec:cap}. Additionally, we introduce pooling by attention in this paper.

Although some elements in the MIV-head appear in the prior literature, they have not been appropriately combined and successfully adopted in the CDFSL application. We believe it is important to carefully orchestrate all elements to tackle the challenges of backbone-agnostic CDFSL. As shown in our ablation study (Section~\ref{sec:ablation}), combining elements mechanically or simplistically fails. Therefore, it is a nontrivial contribution to demonstrate how to achieve empirical success by well-conceived strategies, under the guiding principle of ``competition'', as done in this paper. Moreover, the MIV-head \emph{does} have elements that are entirely new---particularly, the  ``pooling by attention'' mechanism in Component 1 and the ``in-attention skip-connection'' mechanism in Component 2.

While most components of the MIV-head are primarily justified empirically, they are implementations of well-established principles in AI. In particular, the principle of introducing competition to solve problems of an ``assignment'' nature, typically through some form of attention or optimization mechanism, is frequently applied in machine learning and computer vision. Representative examples include ``slot assignment'' in object-centric learning using slot attention \citep{slot20, ocl23} and ``cluster assignment'' in unsupervised or self-supervised learning using the Sinkhorn-Knopp algorithm \citep{swav20}.


\section{Related Work} \label{sec:background}
Two bodies of research are closely related to our paper: work on the core components of the MIV-head and work on CDFSL methods, with the latter broadly categorized into transfer learning (or adaptation), meta-learning, and hybrid approaches. Representative work on CDFSL includes that of~\citet{bit_fs21},~\citet{fit23}, and~\citet{case22}.

\subsection{Multi-instance verification (MIV) and mid-level blocks}
In MIV \citep{miv25}, a query instance is verified against a bag of instances with heterogeneous, unknown relevance. \citet{miv25} showed that naive combinations of multi-instance learning (MIL), a form of learning from bags of data introduced by~\citet{mil97}, and standard verification methods such as Siamese neural networks may fail in this setting. \citet{miv25} proposed a new pooling framework named ``cross-attention pooling'' (CAP), in which all instances within the target bag \emph{compete} to represent the bag in a Siamese-twins architecture. The outputs of CAP are two dense feature vectors: a bag representation and a transformed query. The approach we present in this paper is based on the observation that the two vectors are well suited to represent a prototype and the query in FSC, rendering CAP a key component of our MIV-head. In addition to the attention mechanisms used by standard transformers \citep{transformer17}, CAP applies two novel attention functions proposed by~\citet{miv25}, one of which is used in the MIV-head presented here. For more details, we refer the reader to Section~\ref{sec:cap}, Appendix~\ref{app:cap_maths} and~\cite{miv25}.

It is worth highlighting one property of the bag representation produced by CAP: it is query-dependent at inference time, meaning that the representation varies with the query instance. This property is appealing when the bag representation is used as a prototype in FSC, as it allows the prototype to adapt efficiently not only to new domains, but also to new queries.

One of the characteristics of the MIV-head is that it retrieves embeddings from multiple mid-level blocks of the backbone (sometimes called ``intermediate layers'' in the literature). This procedure is common in computer vision, including in semantic segmentation \citep{segmentation_survey20}, pretraining \citep{blip23}, and FSL \citep{fusion20, midlevel21}. Our application of this idea differs from existing approaches in three ways. First, as we will show, the usefulness of this component depends on the presence of other components of the MIV-head. As shown in Tables~\ref{tab:intx} and \ref{tab:intx2} (Section~\ref{sec:abl_dependence}), the contribution of retrieving embeddings from mid-level blocks (i.e., Component 3) would be negligible in the absence of the other components. This highlights the importance of attributing the performance of the MIV-head to all components collectively, rather than assessing the contribution of each component in isolation. Second, our method does not process patch-level embeddings using ``global average pooling'', which is the operation typically used in the literature. Third, in contrast to approaches that aggregate all embeddings from the mid-level blocks into new embedding(s), we aggregate \emph{logits} computed from the embeddings of each block.

\subsection{Few-shot domain transfer}
The MIV-head falls into the category of CDFSL methods that aim to adapt to target domains at the ``meta-testing'' stage, based on few-shot samples. This category consists primarily of two types of approaches. 

The first type conducts backbone fine-tuning, either fully (\cite{pmf22}) or partially (\cite{tsa22, ett22, dipa24}), and has generally produced the strongest results. Among them, two adapter methods, TSA~(\cite{tsa22}) and eTT~(\cite{ett22}), are chosen as baselines. We chose these baselines because they do not require meta-training to train adapter parameters based on source domains: their adapters can be trained from scratch during ``meta-testing'' based on the support set from the target domains, just like the MIV head. This allows us to use them as pure adaptation baselines in backbone-aligned comparisons. They are also known for their impressive performance, see \citep{closerlook23, vitbaseline24}.

The second type, comprising ``head'' approaches, does \emph{not} modify the backbone. Much of the early work on CDFSL belongs to this category, focusing on suitable choices for the classification head, or ``classifier'' \citep{survey20}, such as linear classifiers~\citep{baseline20, embed_linear20, cnaps19, case22}, cosine classifiers~\citep{closerfewshot19}, nearest centroid classifiers (NCC)~\citep{ncc13, protonet17, cos21}, EMD-related classifiers (e.g., DeepEMD of~\citet{emd20}), and Gaussian naive Bayes~\citep{fit23}. Appendix~\ref{app:fh} shows that two of the best-performing classifiers, Baseline++~\citep{closerfewshot19} and ``FiT Head'' \citep{fit23}, when used standalone, are not competitive with our approach. There also exist hybrid methods (see, for example, \citep{tipadaptor22, dipa24}), but their good performance was shown to be primarily due to fine-tuning: ``freezing'' the backbone performed substantially worse than fine-tuning. 

Our approach belongs to the second type but can achieve performance comparable to that of the first type. Note that our method can also be viewed as a kind of ``few-shot reprogramming'' of black-box models~\citep{blackbox24}, an emerging area where, to our knowledge, no approach can currently match SOTA fine-tuning methods.

\subsection{Meta-learning}
Meta-learning approaches to CDFSL use meta-training to train backbones and/or other learnable parameters in a ``learning-to-learn'' manner, see, for example, \citep{ctx20, distill21, fdmixup22}. Among them, the work on ``CrossTransformers'' (CTX) by~\citet{ctx20} is related to our work because it also applies a cross-attention mechanism akin to CAP. However, although both CTX and the MIV-head create prototypes using a cross-attention mechanism inspired by transformers~\citep{transformer17} and relate to MIL~\citep{mil97}, they are very different approaches: CTX's cross-attention mechanism flattens all patches of images in the query-set and support-set, and CTX trains all parameters, including backbone weights, during meta-training, which makes it a meta-learner. In contrast, we aggregate patches and images in a hierarchical manner through different components (Components 1 and 2) of the MIV-head and use frozen backbone weights, making our approach a ``meta-testing'' adaptation framework.

There are other meta-learners that also use cross-attention \citep{can_fsc19} or MIL \citep{mil_fsc22}, but they focus on ``in-domain'' settings and backbone enhancements, as opposed to cross-domain classification heads. Notably, many meta-learning algorithms employ adapters, including adapters for CNNs~\citep{fwtransform20, tsa22, fit23} and ViTs~\citep{ett22, adaptformer22, vitbaseline24}. The work of \citet{case22} and~\citet{lite21} highlights and addresses the optimization challenges of such methods. Methods whose adapters are trained through meta-training, including FiLM~\citep{flute21}, CaSE~\citep{case22}, and FiT~\citep{fit23}, are orthogonal to our approach and can be used jointly with the MIV-head.\footnote{While FiT~\citep{fit23} has its own classification head (see Appendix~\ref{app:fh}), its backbone is still orthogonal to the MIV-head. } There are also ensemble methods such as stacking \citep{fes24} that could be applied in conjunction with the MIV-head because it produces multiple candidate logits (\emph{cf.} Section~\ref{sec:arch}).

\section{The MIV-head} \label{sec:arch}

The MIV-head comprises three core components: the ``pooling-by-attention'' mechanism, the CAP mechanism, and ``multi-block'' computation/aggregation of logits. Before delving into the details of the MIV-head---which will be discussed in Sections~\ref{sec:comp1}, \ref{sec:cap}, \ref{sec:comp3}---we first illustrate in Section~\ref{sec:dataflow} the end-to-end data flow that goes sequentially through the three components, including some concrete examples of input/output data shapes, and the main purpose of each component.

\begin{figure}[t]
    \centering
    \includegraphics[width=\textwidth]{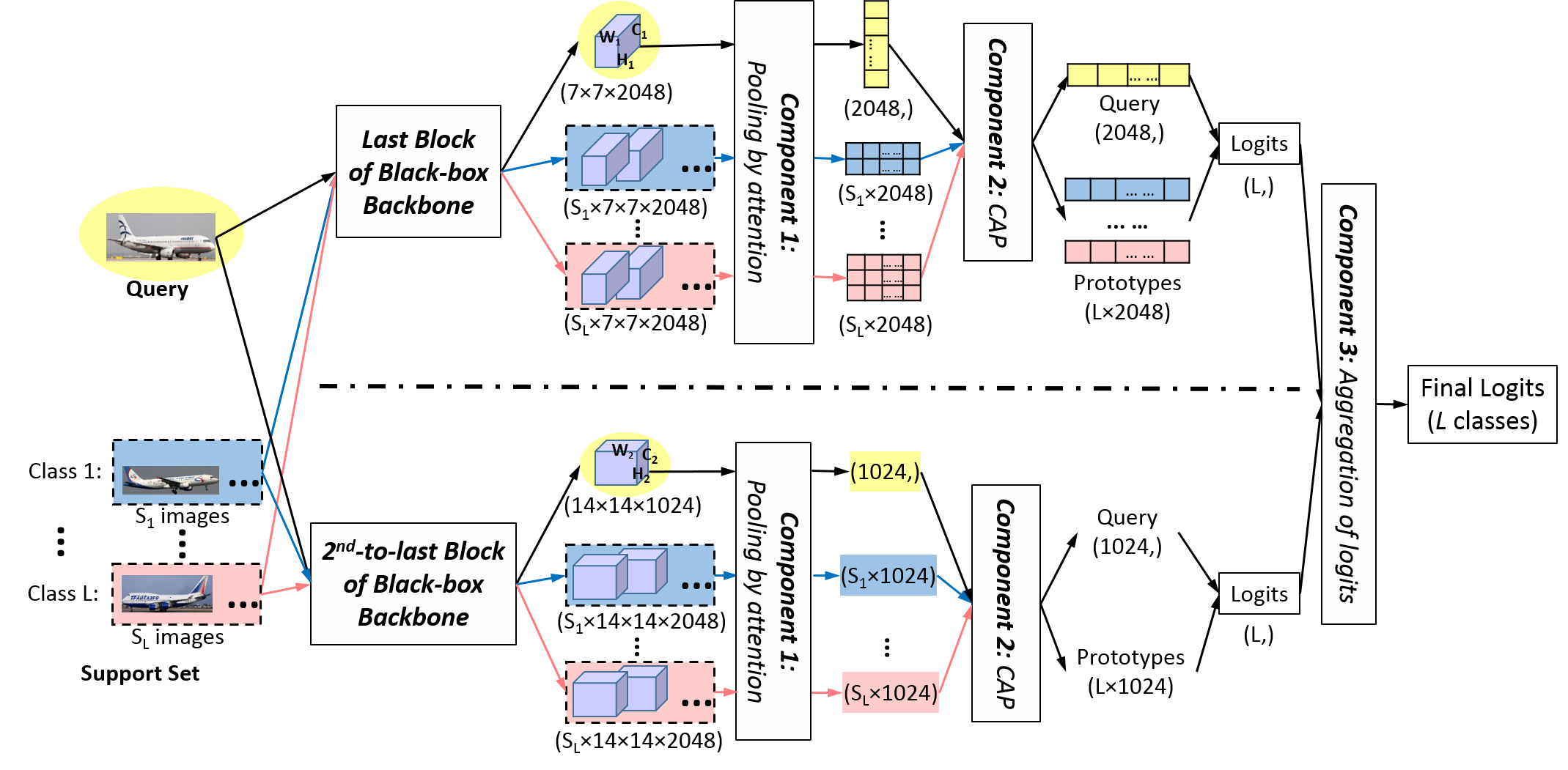}
    \caption{Illustration of the end-to-end data flow through the MIV-head (described in Section~\ref{sec:dataflow}), including the shapes of inputs and outputs for each core component of the MIV-head. This data-flow diagram assumes a hypothetical backbone that, given an input image, produces an output in the shape of $(7\times7\times2048)$ from the last block, and $(14\times14\times1024)$ from the second-to-last block. $L$ denotes the number of classes (i.e., ``number of ways'') in the support set, and $S_l\,(l=1,\ldots,L)$ denotes the number of support samples (i.e., ``number of shots'') within the $l^{th}$ class.} \label{fig:dataflow}
\end{figure}

\subsection{End-to-end data flow}\label{sec:dataflow}

The data flow through the MIV-head is delineated by Figure~\ref{fig:dataflow}, starting with a query image and all images in the support set involving $L$ classes. We denote $S_l\:(l=1,\ldots,L)$ as the number of support images within each class. For any input image, the raw embedding retrieved from the $n^{th}$ block of a backbone's API is a feature map in the shape $(H_n\times W_n\times C_n)$: it represents $(H_n\times W_n)$ patches that constitute this image, where $H_n$, $W_n$ denote the spatial dimensions (i.e., height and width) of the image in terms of number of patches, and each patch is represented by a feature vector of size $C_n$ (i.e., channel size). 
Our ultimate goal is to classify the query image as one of the $L$ classes based on the support set, using the frozen feature maps retrieved from the API of a ``black-box'' backbone.

To make our illustration more concrete, Figure~\ref{fig:dataflow} assumes a hypothetical backbone that produces an embedding for an input image, in the shape of $(7\times7\times2048)$ from its last block and $(14\times14\times1024)$ from its second-to-last block. Also assuming that, as a hyperparameter, we would utilize those two blocks' embeddings as the inputs to the MIV-head, we obtain the input data shapes, for both the query and support set, of the first component of the MIV-head, as depicted in Figure~\ref{fig:dataflow}.

Component 1 (``pooling-by-attention'') aims to address the question ``\textbf{How should an image's patch-level feature map be aggregated into a single, image-level vector?}'' Using the example in Figure~\ref{fig:dataflow}, Component 1 needs to convert an input of shape  $(7\times7\times2048)$ to $(1\times2048)$, and one of shape $(14\times14\times1024)$ to $(1\times1024)$, originating from the two blocks, respectively---such outputs are required as inputs to the subsequent component (Component 2: CAP). To create suitable outputs, we describe in Section~\ref{sec:comp1} a mechanism that ``soft-selects'' spatial dimensions and patches, by assigning attention weights during the pooling of the patches' feature maps into a single vector that represents the whole image.

Component 2 (``cross-attention pooling'', CAP) intends to answer the question ``\textbf{How should a prototype be generated from a bag of support images within the same class?}'' As shown in Figure~\ref{fig:dataflow}, this component accepts inputs of image-level vectors produced by Component 1, and treats each class in the support set as a bag of vectors, i.e., the input data shape is $(S_l\times2048)$/$(S_l\times1024)$ for the $l^{th}$ class. The output for each class is a bag-level representation in the shape $(1\times2048)$/$(1\times1024)$, typically termed ``prototype'' (of a class) in the literature. As stated in Section~\ref{sec:intro}, the adoption of CAP to pool $S_l$ vectors into one vector is motivated by treating this task as an MIV problem, where the bag contains unknown, ``noisy'' members---recognizing that, here, the ``noise'' is induced by a black-box backbone model transferred to a new domain, leading to the low quality of embedding vectors. Another output of Component 2 is the transformed query vector, due to the Siamese-network-style architecture of CAP that shares parameters between the bag and query. Therefore, Component 2 outputs a pair ``(Query, Prototypes)'' with shapes $((1\times2048), (L\times2048))$ and $((1\times1024), (L\times1024))$ for the two blocks, respectively.

We emphasize that, as highlighted by Figure~\ref{fig:dataflow}, Component 1 and 2 are both block-specific because each block requires a separate component owing to the difference in data shapes across blocks. This naturally leads to the question ``\textbf{How should features (i.e., outputs from Component 2) from several mid-level blocks be used?}'' This is the main purpose of Component 3 (``multi-block'' computation/aggregation of logits). It computes $L$-class logits---based on the vectors of query and prototypes---for each of the blocks, and then aggregates them across all blocks, to obtain the final logits used to predict the classification of the query image.

In what follows, we describe the details of each component of the MIV-head.

\begin{figure}[t]
    \centering
    \includegraphics[width=\textwidth]{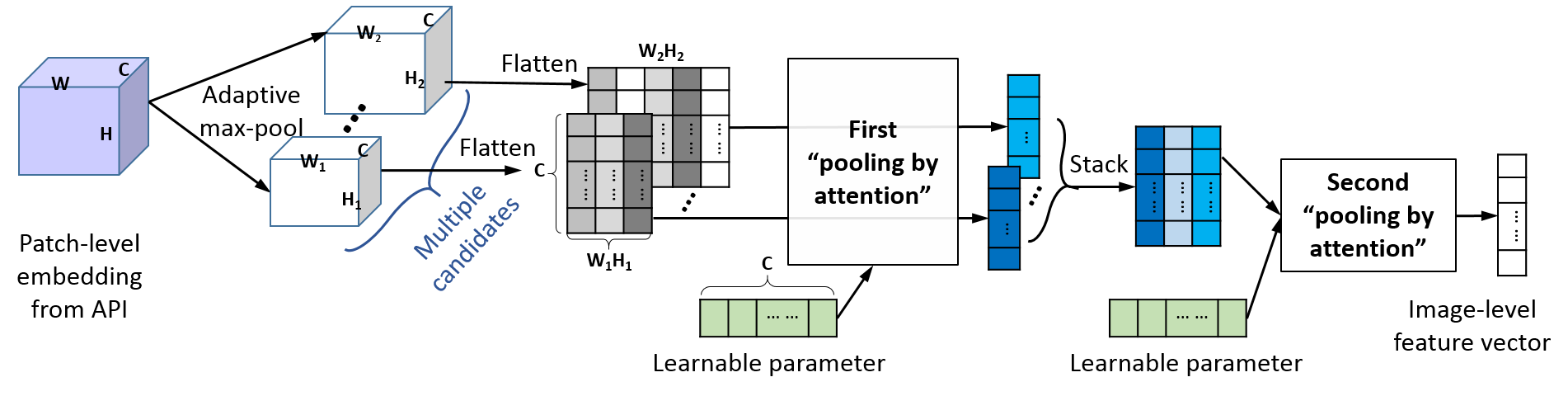}
    \caption{Component 1 of the MIV-head: ``pooling by attention'', described in Section~\ref{sec:comp1}. This component derives candidates from patch-level embeddings extracted from the backbone, and transforms them into an image-level feature map. The first and second ``pooling by attention'' denote Equations~\eqref{ppool1} and \eqref{ppool2}, respectively.}\label{fig:c1}
\end{figure}

\begin{figure}[h!]
\centering
\begin{algorithm}[H]
\caption{Component 1 processing an embedding from a single block of backbone}\label{alg:c1}
\begin{minted}{python}
import torch.nn.functional as F

# Inputs:
# 1) embedding: a raw output from single block of backbone, representing
#    an image, shape=(height, width, channel)
# 2) theta, mu: learnable parameters within Eq. (1) and (2), respectively
# 3) adaptive_shapes: hyperparameter, a list of user-specified patch shapes
# 4) tau: hyperparameter, a scaling factor of attention logits
#
# Outputs:
# i) output: a vector representation of the input image, shape=(1, channel)
        
channel = embedding.size(-1)                            # shape=(H, W, C)
channel_first = embedding.permute(2,0,1) # adapt-pool needs channel-first
temperature = (channel ** 0.5) / tau
multiple_candidates = []
for s in adaptive_shapes:
    candidate = F.adaptive_max_pool2d(channel_first, s)          # Step 1
    candidate = candidate.reshape(channel, -1) # flatten to shape=(C, HW)
    y = attn_pool(candidate, theta, temperature=temperature)      # Eq(1)
    multiple_candidates.append(y)
x = torch.cat(multiple_candidates, dim=-2)         # stack to shape=(D,C)
output = attn_pool(x.mT, mu, temperature=temperature)     # Step 2: Eq(2)

# -------- A utility function that implements Equation (1)/(2) -------- #
def attn_pool(v, param, temperature):     # shape: v=(C,seq), param=(1,C)
    attn_logits = param @ F.normalize(v, dim=-2)          # shape=(1,seq)
    attn = F.softmax(attn_logits / temperature, dim=-1)   # shape=(1,seq)
    return attn @ v.mT                                      # shape=(1,C)
\end{minted}


\end{algorithm}
\caption{Algorithm~\ref{alg:c1} lists the PyTorch-style pseudo-code that describes Component 1 of the MIV-head, where the algorithm processes an output retrieved from a single block of the backbone. Note that it uses a utility function (named ``attn\textunderscore pool'') that implements Equations~\eqref{ppool1} and \eqref{ppool2}---the definition of this function (in pseudo-code) is also included in Algorithm~\ref{alg:c1}.}\label{fig:alg}
\end{figure}

\subsection{Component 1: ``Pooling-by-attention'' to convert patch-level feature maps to an image-level vector}\label{sec:comp1}

The patch-level feature maps produced by the backbone can be useful to identify the ``Regions of Interest'' in an image, which are usually crucial in domain transfer. To make the best use of those feature maps, the following two questions must be addressed:
\begin{enumerate}
    \item What is the optimal number of patches for an image, or equivalently, the optimal spatial ``coverage'' of each patch within the image?
    \item Given the number of patches, which patches are relevant to represent an image?
\end{enumerate}

The first question rests on the observation that the spatial dimensions of the patch-level feature maps, denoted $A_n\in\mathbb{R}^{H_n\times W_n\times C_n}$, are \emph{not} unique. For example, inspired by ``spatial pyramid pooling''~\citep{spp14}, for a given $A_n$, we can apply an adaptive max-pooling and derive a different patch-level representation of the same image. The new representation will have the same number of channels $C_n$ but different spatial dimensions $(H_n'\times W_n')$ where $H_n'\leq H_n,W_n'\leq W_n$. Hence, one needs to be selective when aggregating all derived candidates of $A_n$, each with different spatial dimensions, into a single, image-level vector.

The second question can be addressed by some form of ``patch selection'' during the pooling of the patches' embeddings---simple methods that treat all patches equally, like average-pooling, tend to be suboptimal in our approach, as shown in Section~\ref{sec:abl_alone}, even though they may work well with ``fine-tuning'' methods. This is intuitive since the embedding vectors produced by the black-box backbones are \emph{static}, unlike the fine-tunable feature vectors.

To this end, we design a novel mechanism to ``soft-select'' both spatial dimensions and patches, by applying weighting functions to them in a hierarchical manner. We term this mechanism ``pooling by attention'', as the weighting factors within the weighting functions are attention scores.

More precisely, we first create multiple candidates of $A_n$ using adaptive max-pooling, and then aggregate those candidates into an image-level vector in two steps. In the first step, all patches within a candidate \emph{compete} to represent this candidate. The second step lets all candidates \emph{compete} to represent the image. Both competitions use attention mechanisms. Figure~\ref{fig:c1} illustrates the two steps graphically, and Algorithm~\ref{alg:c1} (in Figure~\ref{fig:alg}) provides PyTorch-style pseudocode. We describe both steps in more detail as follows.

In the first step, each candidate of $A_n$ is transformed into a single vector 
denoted $I_n\in\mathbb{R}^{1\times C_n}$. Concretely, we flatten $A_n$ to $\vec{A}_n$ of the shape $(H_nW_n\times C_n)$, and convert $\vec{A}_n$ to $I_n$ through the following attention function:
\begin{align}
I_n &= \underbrace{softmax\Big(\frac{\theta\cdot\vec{A}_n^T}{\left \lVert \vec{A}_n \right \rVert_2}\frac{\tau}{\sqrt{C_n}}\Big)}_{\text{Attention Score}} \vec{A}_n, \label{ppool1}
\end{align}
where $\theta\in\mathbb{R}^{1\times C_n}$ is a learnable parameter, initialized with $0$.\footnote{The initialization with zeros implies that the starting point of our search for an optimal pooling is an average-pooling.} $T$ denotes the transpose, $\left \lVert \cdot \right \rVert_2$ denotes L2-norm, $\frac{\tau}{\sqrt{C_n}}$ is a scaling factor applied to the unnormalized attention scores, and $\tau$ is a hyperparameter. Equation~\eqref{ppool1} can be directly applied to any CNN backbone, but ViTs, which already apply layer-normalization ($LayerNorm$) to their output, need a minor modification: $\frac{\vec{A}_n}{\left \lVert \vec{A}_n \right \rVert_2}$ is replaced with $\frac{\vec{A}_n}{\sqrt{C_n}}$, and a $LayerNorm$ is applied to $I_n$ subsequently.

Given an image, assuming there are $D_n$ candidates of $A_n$, the first step above would yield $D_n$ different candidates of $I_n$, denoted $I^i_n, i=1,\cdots D_n$. The second step lets all $I^i_n$ compete to represent this image. More concretely, we stack all $I^i_n$, resulting in a matrix $B_n = stack(I^1_n,\cdots, I^{D_n}_n)\in\mathbb{R}^{D_n\times C_n}$. The image-level feature vector $M_n\in\mathbb{R}^{1\times C_n}$ can be obtained by pooling $B_n$ with attention using another learnable, zero-initialized parameter $\mu\in\mathbb{R}^{1\times C_n}$, akin to Equation~\eqref{ppool1}:
\begin{align}
M_n = softmax\Big(\frac{\mu\cdot B_n^T}{\left \lVert B_n \right \rVert_2}\frac{\tau}{\sqrt{C_n}}\Big)B_n. \label{ppool2}
\end{align}
For ViTs, we again replace $\frac{B_n}{\left \lVert B_n \right \rVert_2}$ with $\frac{B_n}{\sqrt{C_n}}$, and perform one additional modification: as part of $B_n$, we include an embedding of the special token ``[CLS]'' produced by ViTs (with shape $(1\times C_n)$), which tends to be a strong predictor.


While the pseudo-code listed in Algorithm~\ref{alg:c1} (Figure~\ref{fig:alg}) explains the above two-step procedure, we provide  further clarity by illustrations using the hypothetical example in Section~\ref{sec:dataflow}. Without loss of generality, we only focus on the last block of the backbone, where the raw $A_n$ has the shape $(7\times7\times2048)$. Assuming, as a hyperparameter, one would search for optimal candidates of $A_n$ among the spatial dimensions of $(7\times7)$, $(6\times6)$ and $(5\times5)$, there are $D_n=3$ candidates: with the number of patches being $49(=7\times7)$, $36(=6\times6)$, $25(=5\times5)$, respectively. All candidates are generated using adaptive max-pooling applied to the raw $A_n$. For the first step, we apply Equation~\eqref{ppool1} to each (flattened) candidate and obtain $I_n$ of the shape $(1\times2048)$---this is the attention-weighted pooling of all patches within each candidate. In the second step, we stack three $I_n$ into $B_n$, of the shape $(3\times2048)$. Applying Equation~\eqref{ppool2} to $B_n$, we construct an image's feature vector in the shape $(1\times2048)$---this is the attention-weighted pooling of all candidates (with different spatial dimensions). 

Intuitively, the parameter $\theta$ in Equation~\eqref{ppool1} (and likewise $\mu$ in Equation~\eqref{ppool2}) can be viewed as a learnable ``query'' of an attention mechanism. 
The attention score of a patch, determined by its embedding and $\theta$, is its weighting factor in a weighted average of all patches within $\vec{A}_n$. In this sense, the attention score of each patch defines its ``share in the competition'' to represent $I_n$. As will be shown by Section~\ref{sec:abl_alone}, such competitions improve the quality of image-level representation, to be fed into the subsequent component, CAP.

\subsection{Component 2: CAP to create prototype and query representations}\label{sec:cap}

\begin{figure}[h]
    \includegraphics[width=\textwidth]{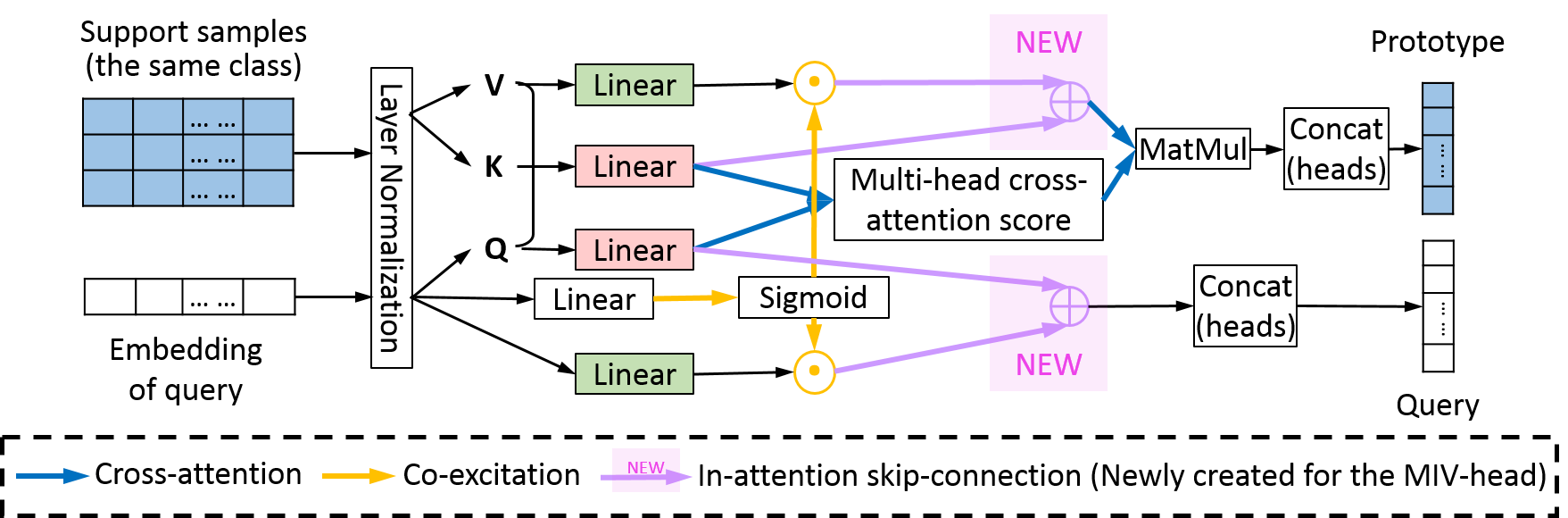}
    \caption{Component 2 of the MIV-head: ``Cross Attention Pooling'' (CAP), described in Section~\ref{sec:cap}. This component creates prototypes, one for each class, based on the feature vectors of a query image and all images of the same class in the support set. \textbf{Q}, \textbf{K}, \textbf{V} represent the ``query'', ``key'', and ``value'' elements of a multi-head attention mechanism. ``\textbf{Linear}'' stands for linear projections from channel dimension to individual heads. The red, green, white fill-colors of \fbox{\textbf{Linear}} denote $W_j^K, W_j^V, \kappa$, respectively, described by Equations~\eqref{perhead} and \eqref{sce}. The cross-attention, co-excitation, and ``in-attention skip-connection'' mechanisms are highlighted by arrowed lines in blue, yellow and purple colors (specified by the legend). We also highlight that the ``in-attention skip-connection'' mechanism is a new element created for the MIV-head, non-existent in the original CAP \citep{miv25}. $\odot$, $\oplus$, ``MatMul'', ``Concat'', ``Sigmoid'' denote element-wise multiplication, element-wise addition, matrix multiplication, concatenation (of multi-heads), and sigmoid-activation operators, respectively.}\label{fig:c2}
\end{figure}

Subsequent to Component 1, we apply CAP by treating each class in the support set as a target bag and the query as the ``query instance''---this representation fits the original CAP~\citep{miv25} structure well. More precisely, for the $n^{th}$ block we can write the formulation of a Siamese-twins architecture as $v^{P_l}_n, v^Q_n = CAP_n(P^l_n, Q_n)$, where $P^l_n\in\mathbb{R}^{S^l\times C_n}$ stands for the bag of embeddings of $S^l$ images in the $l^{th}$ class. Note that $S^l$, also known as the ``number of shots'', may vary across classes of the support set. $Q_n\in\mathbb{R}^{1\times C_n}$ denotes a query image's feature-vector; $v^{P_l}_n, v^Q_n\in \mathbb{R}^{1\times C_n}$ denote the prototype (of the $l^{th}$-class) and the query representations, respectively.

The function $CAP_n$ contains block-specific parameters and consists primarily of three mechanisms, depicted by lines of different colors in Figure~\ref{fig:c2}. For ease of exposition, the mathematical formulations of CAP are provided in Appendix~\ref{app:cap_maths}, and further details can be found in~\citep{miv25}.

We stress the key distinctions between the original CAP~\citep{miv25} and CAP used by the MIV-head here: Apart from the difference in backbones---\citet{miv25} used CAP jointly with ``backbone fine-tuning'', unlike the frozen/black-box backbones considered in this study---there is also a new mechanism named ``in-attention skip-connection'' in the variant of CAP adopted here, which did not exist in the original CAP and is highlighted by Figure~\ref{fig:c2}.

In Appendix~\ref{app:abl_single}, we show that each of the three mechanisms adds value to the MIV-head, empirically justifying those elements of CAP within our design. We describe them in what follows.

(1) \textbf{Cross-attention}: This mechanism allows image-level representations within the same bag (or class) to \emph{compete} in creation of a prototype, using the query as ``cross-reference''. To account for the relative homogeneity of a bag, whose instances all belong to the same class---compared to strong heterogeneity assumed in standard MIV tasks---we introduce a ``down-scaling'' hyperparameter on the unnormalized cross-attention scores to suppress competition within a bag. This mechanism is illustrated by blue arrowed lines in Figure~\ref{fig:c2}.

(2) \textbf{Co-excitation}: The co-excitation component provides a feature selection mechanism, inspired by the ``squeeze-and-excitation networks''~\citep{se18}, shared by both of the Siamese twins. This mechanism is illustrated by yellow arrowed lines in Figure~\ref{fig:c2}.

(3) ``\textbf{In-attention skip-connection}'': This mechanism is a new addition to the original formulation of CAP in \citep{miv25}, where there is no such ``skip-connection''. Unlike the transformer-style cross-attention~\citep{transformer17} whose parameters of the ``key'' and ``value'' vectors are disjoint, this ``in-attention skip-connection'' mechanism relates the two parameter-sets by adding the ``key''-vector to the ``value''-vector, akin to a ``skip-connection'' mechanism. To differentiate from the skip-connection in the transformer that is applied \emph{posterior to} attention, we refer to this mechanism as ``in-attention skip-connection''. As shown in the ablation study (Table~\ref{tab:more_abl} of Appendix~\ref{app:abl_single}), such a connection between ``key'' and ``value'' components in attention appears to bring a slight performance improvement (compared with ``no skip-connection''), and is thus adopted in CAP here. This mechanism is illustrated by purple arrowed lines in Figure~\ref{fig:c2}.

It is also worth noting that the linear projections from all embedding channels to individual heads are shown as ``\textbf{Linear}'' in Figure~\ref{fig:c2}. While such projections are standard in ``multi-head attention'' of transformers~\citep{transformer17}, the sharing of their weights by different mechanisms must be carefully considered in our case, as shown by different fill-colors of \fbox{\textbf{Linear}} in Figure~\ref{fig:c2} (red and green). The same set of projections is shared between Siamese twins and among all classes, but \emph{not} across blocks (as indicated by $n$ in $CAP_n$) since the channel sizes used by the projections are block-specific. Such sharing reduces the number of learnable parameters to prevent over-fitting. 

\subsection{Component 3: Computing and aggregating logits from multiple blocks}\label{sec:comp3}

\begin{figure}[t]
    \includegraphics[width=\textwidth]{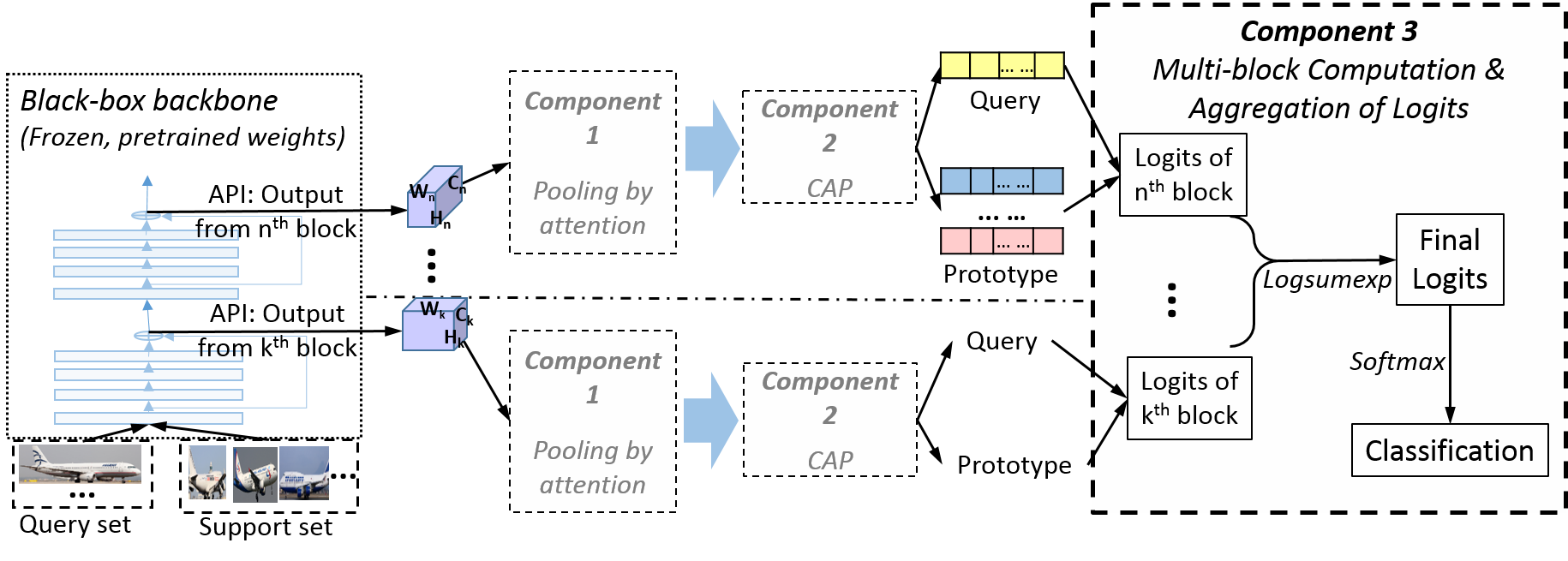}
    \caption{Component 3 of the MIV-head: ``multi-block computation and aggregation of logits'', described in Section~\ref{sec:comp3}. The block-wise logits are computed using Equation~\eqref{logits}.}\label{fig:c3}
\end{figure}

As delineated by Figure~\ref{fig:c3}, for the $n^{th}$ block ($n$=1,$\ldots$,N) and $l^{th}$ class ($l$=1,$\ldots$,L), logits can be computed using a ``centered cosine similarity with temperature'' between the two outputs of CAP: $v^{P_l}_n$ and $v^Q_n$,
\begin{equation}\label{logits}
    logits_n^l = \frac{(v^Q_n - m)\cdot(v^{P_l}_n - m)^T}{\gamma\left \lVert (v^Q_n-m)\right \rVert_2\left \lVert (v^{P_l}_n-m)\right \rVert_2}
\end{equation}
where $m$=$mean(v^P_n)\in\mathbb{R}^{1\times C_n}$ is the channel-wise mean across all classes, $v^P_n$=$stack(v^{P_1}_n,\cdots,v^{P_L}_n)\in\mathbb{R}^{L\times C_n}$, and $\gamma$(=0.1) is the ``temperature''. We aggregate the class logits from the different blocks using the $logsumexp$ function, which is a differentiable approximation of the $max$ function: $logits^l = logsumexp(logits_1^l,\cdots,logits_N^l), l=1,\ldots,L$, to obtain the final logits used to classify the query image.

The entire pipeline, including all three components, is trained end-to-end for a fixed number of steps using the support set and cross-entropy loss, prior to inference and evaluation. 

\subsection{Remarks}

The rationale behind our architectural design is to introduce \emph{competitions} in parts of the model, including ``pooling-by-attention'' (competition between patches), CAP (competition between same-class support samples), and $logsumexp$ (competition between logits from multiple backbone blocks). We believe repeated competitions in the three components collectively address the well-known difficulty in FSC of learning from a small number of support samples~\citep{tsa22, fit23}, leading to more effective representations and better performance in the target domains---a key hypothesis to be tested in our experiments.

\section{Experiments} \label{sec:experiments}

Our experiments follow the standard ``varying-way varying-shot'' (and additionally ``five-way one-shot'') protocol used in previous literature \citep{ctx20, tsa22, ett22}. Given our focus on adaptation during ``meta-testing'', within our experiments all algorithms train relevant parameters based on the few-shot support set only, before inference and evaluation on test queries.

\subsection{Experimental setup} \label{sec:setup}

\subsubsection{Data and backbones}

\begin{table}[t]
\centering
\begin{NiceTabular}{@{}lcccccccccccc@{}}\toprule
\Block[c]{2-1}{Test\\dataset}&& \multicolumn{3}{c}{Support size} && \multicolumn{3}{c}{Class count} && \multicolumn{3}{c}{Mean shot} \\
 &&	min	&	mean	&	max	&&	min	&	mean	&	max	&&	min	&	mean	&	max	\\
\cline{1}\cline{3-5}\cline{7-9}\cline{11-13}

Omniglot	&&	5	&	89.06	&	319	&&	5	&	19.04	&	45	&&	1	&	4.70	&	9.00	\\
Aircraft	&&	5	&	341.91	&	497	&&	5	&	9.92	&	15	&&	1	&	36.20	&	79.80	\\
Birds	&&	5	&	316.73	&	494	&&	5	&	17.40	&	30	&&	1	&	19.80	&	44.00	\\
Textures	&&	5	&	282.42	&	498	&&	5	&	6.02	&	7	&&	1	&	46.88	&	96.80	\\
Quick Draw 	&&	13	&	433.51	&	498	&&	5	&	28.80	&	50	&&	1	&	19.90	&	93.40	\\
Fungi	&&	5	&	361.61	&	495	&&	5	&	27.32	&	50	&&	1	&	16.11	&	60.60	\\
VGG Flower	&&	6	&	302.99	&	497	&&	5	&	10.73	&	16	&&	1	&	29.34	&	78.40	\\
Traffic Sign	&&	8	&	418.04	&	497	&&	5	&	23.55	&	43	&&	1	&	22.76	&	96.40	\\
MSCOCO	&&	17	&	419.84	&	498	&&	5	&	22.87	&	40	&&	1	&	23.66	&	94.60	\\
MNIST	&&	6	&	324.70	&	498	&&	5	&	7.51	&	10	&&	1	&	44.04	&	99.40	\\
CIFAR-10	&&	5	&	333.66	&	498	&&	5	&	7.52	&	10	&&	1	&	45.59	&	98.60	\\
CIFAR-100	&&	7	&	414.44	&	496	&&	5	&	28.01	&	50	&&	1	&	18.89	&	79.20	\\
CropDisease	&&	9	&	410.81	&	498	&&	5	&	21.86	&	38	&&	1	&	23.16	&	82.83	\\
EuroSAT	&&	5	&	330.78	&	498	&&	5	&	7.60	&	10	&&	1	&	44.42	&	98.40	\\
ISIC	&&	5	&	288.39	&	499	&&	5	&	6.02	&	7	&&	1	&	48.34	&	99.60	\\
ChestX	&&	5	&	285.32	&	498	&&	5	&	6.05	&	7	&&	1	&	47.13	&	99.60	\\
Food101	&&	6	&	417.28	&	498	&&	5	&	28.39	&	50	&&	1	&	19.21	&	85.60	\\
\bottomrule
\end{NiceTabular}
\caption{Summary statistics of the (600) test episodes for all MD+ datasets used in our main experiments.} \label{tab:data_summ}
\end{table}

Following prior studies~\citep{fes24, tsa22}, we adopt the original MD benchmark~\citep{md20} consisting of nine non-ILSVRC datasets\footnote{Similar to ~\citep{closerlook23}, we excluded ILSVRC-2012 from MD to avoid leaking information, as it may overlap with the off-the-shelf backbones' pretraining data.}, plus an additional eight from \citep{food14, cnaps19, mdmore20} also commonly used in FSC---thus 17 test datasets in total (henceforth referred to as ``MD+''). All the datasets are well-known benchmarks for few-shot classification---they cover a diverse range of target domains and are frequently used in the CDFSL literature. We used TSA's sampling schema instead of the original MD sampling procedure, to avoid a shuffling issue in MD (see Appendix~\ref{app:data}). We evaluate all algorithms on the same 600 test tasks (or ``episodes''), to facilitate comparisons using a paired {\em t}-test, minimizing the effects of data artifacts such as sampling randomness, differences in image pre-processing, etc. We tabulate the key summary statistics for the 600 test episodes in Table~\ref{tab:data_summ}.

We also use the same set of off-the-shelf feature extractors whose model weights were pretrained on ILSVRC---including the family of ResNet~\citep{resnet16} and ViT~\citep{vit21, deit21} backbones---for all algorithms. We used both supervised and self-supervised pretrained backbones; the supervised backbones were trained using cross-entropy loss. Appendix~\ref{app:backbone} provides details and links to the backbone model weights. While backbone-aligned comparisons between TSA and eTT are impossible due to backbone incompatibility, the MIV-head facilitates comparisons to either of them.

\subsubsection{Hyperparameters}

We re-implemented TSA and eTT using their default hyperparameters, see Appendix~\ref{app:hyper}, in which we also tabulate optimization settings of all algorithms. Most of the MIV-head hyperparameters are determined by the number ($D_n$) and spatial dimensions ($H_n\times W_n$) of the candidate patch-level representations derived by adaptive max-pooling in Component 1. The choice of values depends on the output shapes from different backbones' APIs. For example, the raw outputs from the last two blocks of ResNet-50 have the spatial dimensions $(14\times14)$ and $(7\times7)$, respectively. Hence, the permissible range\footnote{``Permissible range'' of a block means: (1) the coverage of this block's potential spatial dimensions; (2) non-overlapping with other blocks' potential spatial dimensions. Throughout this paper, we follow the convention of setting $H=W$ for all spatial dimensions and input resolutions.} of the hyperparameters for the second-to-last block is $D_{-2}=1,\ldots,7$ with $(H_{-2}\times W_{-2})=(8\times8),\ldots,(14\times14)$, and for the last block: $D_{-1}=1,\ldots,6$ with $(H_{-1}\times W_{-1})=(2\times2),\ldots,(7\times7)$. We set a fixed value of $D_n$ and ($H_n, W_n$), irrespective of the test domains: $D_n$ is typically between three and five, and values of ($H_n, W_n$) are selected at approximately equal intervals within the permissible range (\emph{cf.} Appendix~\ref{app:hyper}). Those choices are justified by the ablation analysis in Section~\ref{sec:abl_alone}. Furthermore, the ``number of output blocks of the backbone'', i.e., the hyperparameter ``$N$'' in Component 3, is chosen as two (out of a total of four) for ResNet and four (out of twelve) for ViT-small, following recommendations from the literature~\citep{tsa22, dino21}. In addition, to boost sample sizes for ``low-shot'' classes, we created distorted views of the support set using RandAugment~\citep{randaug20} and treated them as additional ``pseudo-queries'' during training, see Appendix~\ref{app:hyper} and \ref{app:abl_single}. Finally, we conducted all experiments on a single GPU.

In what follows, in Section \ref{sec:results}, we present key experimental results on accuracy and adaptation cost using the full MD+ benchmark. In Sections~\ref{sec:extra_results} and~\ref{sec:ablation}, we provide additional in-depth analysis and ablation using the smaller original MD benchmark.

\subsection{Backbone-aligned comparisons to state-of-the-art methods} \label{sec:results}

\begin{table}[t]
    \centering
    \begin{tabular}{@{}lccc@{}}
        Characteristics &  MIV-head (Ours) & TSA & eTT \\\toprule
        \rowcolor{lightgray!50!white}Image& i) Resize to $224\times224$& & \\
        \rowcolor{lightgray!50!white}pre-processing & ii) Normalize data using ``ImageNet &Identical to Ours&Identical to Ours\\
        \rowcolor{lightgray!50!white}& mean \& standard deviation'' & & \\
        \multirow{2}{*}{Backbones} & ResNet-50 (supervised \& self-supervised) &Identical to Ours&Identical to Ours\\
        & ViT-S/16(/8) (supervised \& self-supervised)& (ResNet-50) & (ViT-S/16) \\
        \rowcolor{lightgray!50!white}Modification & No &Yes (add& Yes (add\\
        \rowcolor{lightgray!50!white}of backbone & (Frozen weights \& architecture)& ``adapter'')&``adapter'')\\
        Trainable &\emph{Outside} backbone&\emph{Inside} backbone&\emph{Inside} backbone\\
        parameters & 15.7M(ResNet-50); 1.8M(ViT-S) & 5.5M(ResNet-50) & 2M(ViT-S)\\ 
        \rowcolor{lightgray!50!white}& &Training time:&Training time:\\
        \rowcolor{lightgray!50!white}Adaptation cost & See Figures~\ref{fig:ett_cost_ssp}, \ref{fig:tsa_cost_ssp}, \ref{fig:time_spv}, and Table~\ref{tab:latency} &$1.3\times$--$1.7\times$ of ours &$1.4\times$--$2\times$ of ours\\
        \rowcolor{lightgray!50!white} & &GFLOPs: $>2\times$&GFLOPs: $>2\times$\\
        \bottomrule
    \end{tabular}
    \caption{Summary of key model characteristics of the MIV-head (Ours) and the two baselines, TSA and eTT.}
    \label{tab:bl_summ}
\end{table}

Before presenting the main results, we contrast the MIV-head with the two baselines, TSA and eTT, in terms of key model characteristics, in Table~\ref{tab:bl_summ}. In particular, Table~\ref{tab:bl_summ} highlights an important factor affecting computational cost: whether the trainable parameters are inside or outside the backbone. 

When \emph{inside} the backbone, even a small number of trainable parameters may incur large computational cost. This is because, to update those parameters, the training procedure still needs to pass through the entire backbone, forward and backward, at \emph{every} training step, resulting in considerable computation that possibly overshadows other factors of adaptation cost. In contrast, trainable parameters \emph{outside} the backbone require passing through the backbone only \emph{once}---to retrieve the static embeddings from the backbone at the beginning of training.

Therefore, as shown in Table~\ref{tab:bl_summ}, although they have a smaller or similar number of trainable parameters compared to the MIV-head, the two baselines, TSA and eTT, which are among the most lightweight fine-tuning methods, still incur substantially higher adaptation cost than ours.

\begin{table}[ht]
\centering
\begin{NiceTabular}{@{}lccccccc@{}}
\toprule
\Block[c]{3-1}{Test\\dataset}& \multicolumn{7}{c}{Off-the-shelf pretrained backbone models for $224\times224$ input resolution} \\
& \multicolumn{3}{c}{Supervised ResNet-50} & & \multicolumn{3}{c}{Self-supervised(DINO) ResNet-50}\\ 
\cline{2-4}\cline{6-8}	
&NCC&TSA&Ours&&NCC&TSA&Ours\\
Omniglot	&	50.4$\pm$1.4	&	66.7$\pm$1.4	&\textbf{	76.5$\pm$1.2	}	&&	55.9$\pm$1.3	&	68.5$\pm$1.4	&\textbf{	74.4$\pm$1.2}\\ 
Aircraft	&	54.7$\pm$0.9	&	78.9$\pm$1.1	&\textbf{	84.4$\pm$1.0	}	&&	55.3$\pm$0.9	&	80.2$\pm$1.1	&\textbf{	84.4$\pm$1.0}\\ 
Birds	&	81.0$\pm$0.7	&	84.9$\pm$0.7	&\textbf{	87.2$\pm$0.8	}	&&	60.6$\pm$0.9	&	78.5$\pm$1.0	&\textbf{	81.8$\pm$1.1}\\ 
Textures	&	83.7$\pm$0.6	&	87.4$\pm$0.6	&\textbf{	89.0$\pm$0.6	}	&&	85.3$\pm$0.6	&	89.6$\pm$0.6	&89.6$\pm$0.6	\\ 
Quick Draw 	&	57.2$\pm$0.9	&	69.0$\pm$0.9	&\textbf{	71.7$\pm$0.8	}	&&	59.3$\pm$0.9	&	69.3$\pm$0.9	&\textbf{	70.1$\pm$0.8}\\ 
Fungi	&	44.9$\pm$1.1	&	55.7$\pm$1.1	&\textbf{	60.6$\pm$1.1	}	&&	52.7$\pm$1.1	&	\textbf{61.4$\pm$1.1}	&	60.5$\pm$1.1\\ 
VGG Flower	&	86.0$\pm$0.6	&	92.8$\pm$0.5	&\textbf{	96.0$\pm$0.4	}	&&	94.6$\pm$0.5	&	96.3$\pm$0.4	&\textbf{	96.7$\pm$0.4}\\ 
Traffic Sign	&	49.7$\pm$1.1	&	73.6$\pm$1.1	&\textbf{	78.6$\pm$1.0	}	&&	53.6$\pm$1.2	&	72.9$\pm$1.2	&\textbf{	81.8$\pm$1.0}\\ 
MSCOCO	&	57.0$\pm$1.0	&	\textbf{62.3$\pm$0.9}	&	60.8$\pm$1.0		&&	52.3$\pm$1.1	&	\textbf{62.0$\pm$1.0}	&	59.3$\pm$1.0\\ \midrule
Average (MD)	&	62.7	&	74.6	&\textbf{	78.3	} && 63.3	&	75.4	&\textbf{	77.6	}\\ \midrule
MNIST	&	77.1$\pm$0.8	&	92.0$\pm$0.7	&\textbf{	93.2$\pm$0.6	}	&&	79.2$\pm$0.7	&	91.0$\pm$0.7	&\textbf{	92.1$\pm$0.7}\\ 
CIFAR-10	&	82.0$\pm$0.6	&\textbf{	89.7}$\pm$0.6	&	84.7$\pm$0.7	&&	76.2$\pm$0.7	&	\textbf{84.9$\pm$0.8}	&80.8$\pm$0.8\\ 
CIFAR-100	&	69.1$\pm$0.9	&	\textbf{80.0$\pm$0.8}	&	74.9$\pm$0.8	&&	65.7$\pm$1.0	&	\textbf{75.2$\pm$0.9}	&	72.0$\pm$0.9\\ 
CropDisease	&	80.3$\pm$0.8	&	86.1$\pm$0.7	&\textbf{	91.2$\pm$0.5	}	&&	87.3$\pm$0.6	&	91.4$\pm$0.5	&\textbf{	92.3$\pm$0.5}\\ 
EuroSAT	&	83.8$\pm$0.5	&	90.2$\pm$0.5	&\textbf{	93.2$\pm$0.4	}	&&	89.2$\pm$0.5	&	92.6$\pm$0.5	&\textbf{	93.8$\pm$0.4}\\ 
ISIC	&	35.6$\pm$0.6	&	41.4$\pm$0.9	&\textbf{	43.1$\pm$0.9	}	&&	40.8$\pm$0.7	&	\textbf{45.2$\pm$0.9}	&	44.3$\pm$0.9	\\ 
ChestX	&	24.3$\pm$0.5	&	25.0$\pm$0.5	&\textbf{	25.9$\pm$0.5	}	&&	26.2$\pm$0.5	&	\textbf{28.2$\pm$0.6}	&	27.2$\pm$0.6\\ 
Food101	&	63.4$\pm$1.0	&	67.3$\pm$1.0	&\textbf{	69.5$\pm$1.0	}	&&	59.5$\pm$1.0	&	\textbf{67.2$\pm$1.0}	&	66.6$\pm$1.0	\\ \midrule
Average (MD+)	&	63.5	&	73.1	&\textbf{	75.3	}	&&	64.3	&	73.8	&\textbf{	74.6	} \\ 
\bottomrule 
\end{NiceTabular} 
\caption{``Varying-way varying-shot'': comparison of accuracy (in \%) $\pm$95\% confidence interval between TSA and the MIV-head (Ours) based on all non-ILSVRC datasets in the extended MD benchmark and the same off-the-shelf backbones. A NCC head with \emph{no} learnable parameters is included for reference, reflecting the classification capability purely from the backbone. The row of ``Average (MD)'' indicates the average accuracy across the nine original non-ILSVRC MD datasets whereas ``Average (MD+)'' is the average across the total 17 extended MD datasets. We also conducted a two-sided paired t-test between TSA and Ours, and bolded the higher accuracy when the $p$-value of the $t$-test is $<0.01$ (i.e., significant at $99\%$ confidence level).
} \label{tab:miv_acc_tsa}
\end{table}

\begin{table}[ht]
\centering
\begin{NiceTabular}{@{}lcccccccc@{}}\toprule
\Block[c]{4-1}{Test\\dataset}& \multicolumn{8}{c}{Off-the-shelf pretrained backbone models for $224\times224$ input resolution} \\
& \multicolumn{3}{c}{Supervised(DeiT) ViT-small} & & \multicolumn{4}{c}{Self-supervised(DINO) ViT-small}\\ 
\cline{2-4}\cline{6-9}
&NCC&eTT& Ours&&NCC&eTT\tabularnote{Due to the ``OOM'' issue caused by eTT when it was applied to the backbone taking $8\times8$ input patch-size (patch 8), the closest alternative is eTT based on the ``patch 16'' backbone.
}
& \multicolumn{2}{c}{Ours}\\
\cline{8-9}&&\multicolumn{2}{c}{(Patch 16)}&&&(Patch 16)& (Patch 8) &(Patch 16)\\
Omniglot&54.1$\pm$1.3&69.4$\pm$1.3&\textbf{74.5$\pm$1.2}&&61.8$\pm$1.3&77.0$\pm$1.3&\textbf{78.6$\pm$1.1}&77.0$\pm$1.2\\
Aircraft&54.9$\pm$0.9&78.4$\pm$1.0&\textbf{79.4$\pm$1.0}&&62.4$\pm$1.0&\underline{84.1$\pm$1.0}&\textbf{86.0$\pm$0.9}&83.4$\pm$1.0\\ 
Birds&84.8$\pm$0.6&88.9$\pm$0.6&88.7$\pm$0.7&&89.1$\pm$0.6&\textbf{92.2$\pm$0.6}&91.8$\pm$0.7&89.8$\pm$0.8\\ 
Textures&80.7$\pm$0.6&86.8$\pm$0.6&\textbf{88.3$\pm$0.6}&&86.0$\pm$0.5&89.3$\pm$0.6&\textbf{89.7$\pm$0.6}&89.5$\pm$0.6\\ 
Quick Draw&56.9$\pm$0.9&69.2$\pm$0.8&\textbf{70.9$\pm$0.8}&&62.3$\pm$0.9&\textbf{72.4$\pm$0.8}&71.8$\pm$0.8&71.6$\pm$0.8\\ 
Fungi&49.8$\pm$1.1&59.7$\pm$1.1&\textbf{60.9$\pm$1.1}&&59.6$\pm$1.1&\underline{65.1$\pm$1.1}&\textbf{65.6$\pm$1.0}&64.0$\pm$1.1\\ 
VGG Flower&88.3$\pm$0.6&93.6$\pm$0.5&\textbf{94.7$\pm$0.5}&&96.2$\pm$0.4&\underline{97.4$\pm$0.3}&97.3$\pm$0.4&97.0$\pm$0.4\\ 
Traffic Sign&48.1$\pm$1.2&\textbf{70.6$\pm$1.2}&68.2$\pm$1.2&&53.3$\pm$1.1&\textbf{81.3$\pm$1.1}&79.6$\pm$1.0&77.6$\pm$1.0\\ 
MSCOCO&61.9$\pm$0.9&65.2$\pm$0.9&\textbf{65.7$\pm$0.9}&&57.6$\pm$0.9&\textbf{64.4$\pm$0.9}&63.0$\pm$0.9&62.3$\pm$1.0\\ \midrule
Average (MD)	&	64.4	&	75.8	&\textbf{	76.8	}	&&	69.8	&\underline{	80.3	}&	80.4		&	79.1	\\
\midrule
MNIST&78.7$\pm$0.7&90.5$\pm$0.7&\textbf{91.5$\pm$0.6}&&79.8$\pm$0.7&\textbf{93.5$\pm$0.6}&92.4$\pm$0.5&91.6$\pm$0.6\\ 
CIFAR-10&89.3$\pm$0.5&\textbf{91.5$\pm$0.5}&89.2$\pm$0.5&&86.8$\pm$0.6&\textbf{92.4$\pm$0.5}&90.2$\pm$0.6&86.2$\pm$0.7\\ 
CIFAR-100&77.7$\pm$0.7&\textbf{83.4$\pm$0.7}&80.1$\pm$0.7&&76.2$\pm$0.8&\textbf{84.3$\pm$0.7}&80.6$\pm$0.8&76.5$\pm$0.8\\ 
CropDisease&80.4$\pm$0.8&88.5$\pm$0.7&\textbf{90.7$\pm$0.6}&&88.1$\pm$0.6&92.3$\pm$0.5&\textbf{92.7$\pm$0.5}&\textbf{92.7$\pm$0.5}\\ 
EuroSAT&82.4$\pm$0.6&89.1$\pm$0.6&\textbf{91.0$\pm$0.5}&&90.4$\pm$0.5&93.4$\pm$0.4&\textbf{93.8$\pm$0.4}&\textbf{94.0$\pm$0.4}\\ 
ISIC&40.6$\pm$0.8&42.3$\pm$0.9&\textbf{43.7$\pm$0.9}&&46.3$\pm$0.8&\textbf{48.0$\pm$1.0}&46.1$\pm$0.9&45.7$\pm$0.9\\ 
ChestX&23.8$\pm$0.5&23.5$\pm$0.5&\textbf{25.0$\pm$0.5}&&26.7$\pm$0.6&26.9$\pm$0.5&\textbf{27.7$\pm$0.6}&27.3$\pm$0.6\\ 
Food101&64.4$\pm$1.0&69.5$\pm$0.9&\textbf{70.6$\pm$0.9}&&69.3$\pm$0.9&\underline{72.6$\pm$0.9}&\textbf{75.5$\pm$0.9}&72.2$\pm$0.9\\
\midrule
Average (MD+)	&	65.7	&	74.1	&\textbf{	74.9	}	&&	70.1	&\textbf{	78.0	}&	77.8		&	76.4	\\ 
\bottomrule 
\end{NiceTabular}
\caption{``Varying-way varying-shot'': comparison of accuracy (in \%) $\pm$95\% confidence interval between eTT and the MIV-head (Ours) based on all non-ILSVRC datasets in the extended MD benchmark and the same off-the-shelf backbones. A NCC head with \emph{no} learnable parameters is included for reference, reflecting the classification capability purely from the backbone. The row labeled ``Average (MD)'' indicates the average accuracy across the nine original non-ILSVRC MD datasets whereas ``Average (MD+)'' is the average across the total 17 extended MD datasets. We also conducted a two-sided paired t-test between eTT and Ours, and bolded the higher accuracy when the $p$-value of the $t$-test is $<0.01$ (i.e., significant at $99\%$ confidence level). For the DINO ViT-small backbone, the eTT(Patch-16) results are underlined if they are no better than Ours(Patch-8) model but better than Ours(Patch-16) model.
} \label{tab:miv_acc_ett}
\end{table}

\subsubsection{Accuracy}\label{sec:results_acc}

Tables~\ref{tab:miv_acc_tsa} and \ref{tab:miv_acc_ett} report test results of the MIV-head compared to TSA (ResNet-50 backbone) and eTT (ViT backbone), respectively. 
All input images are resized to $224\times224$ resolution required by the backbones. While our approach's results on the original MD datasets (``Average (MD)'') are better than the reported accuracy of TSA in \citep{tsa22} (78.3$\%$ vs. 76.2$\%$) and eTT in \citep{ett22} (80.3$\%$ vs. 78.7$\%$), such comparisons (see Appendix~\ref{app:acc_more}) are arguably unfair because of different backbones along with data artifacts. Hence, we applied all algorithms based on the same off-the-shelf supervised and self-supervised backbones, and evaluated them on the same test tasks, yielding \emph{backbone-aligned}, unbiased results.

For supervised backbones, the MIV-head achieves higher accuracy than both baselines in the majority of test datasets and on average, across MD and MD+. For self-supervised backbones, the MIV-head outperforms TSA on average, and significantly outperforms TSA on slightly more datasets than those on which it significantly underperforms. As for eTT, we find it computationally infeasible with a backbone using $8\times8$ input patches (referred to as patch-8) due to ``OOM'' issues, although previous literature such as~\citep{closerlook23} showed the patch-8 backbone may be superior to the patch-16 variant. We therefore use patch-16 for eTT. The MIV-head can be applied with either backbone, producing the results in the rightmost two columns of Table~\ref{tab:miv_acc_ett}: ``MIV-head/patch-8'' is generally on par with eTT, while ``MIV-head/patch-16'' has lower accuracy, albeit the difference remains moderate. Given the MIV-head's advantage in adaptation cost demonstrated in Section~\ref{sec:results_cost}, it is a highly competitive alternative to the two baselines. Appendix~\ref{app:acc_more} lists results of other methods reported in the literature---which generally underperform TSA/eTT---to offer a broader perspective.

\begin{table}[t]
\hspace*{-1cm}
\centering
\begin{NiceTabular}{@{}lcccccccccccc@{}}
\Block[c]{5-1}{Test\\dataset}& \multicolumn{12}{c}{Off-the-shelf pretrained backbone models for $224\times224$ input resolution} \\\toprule
& \multicolumn{5}{c}{Supervised} & & \multicolumn{6}{c}{Self-supervised (DINO)}\\\cline{2-6}\cline{8-13}
& \multicolumn{2}{c}{ResNet-50} & & \multicolumn{2}{c}{DeiT-small} & & \multicolumn{2}{c}{ResNet-50} & & \multicolumn{3}{c}{ViT-small}\\ 
& TSA & Ours && eTT &Ours && TSA & Ours && eTT & \multicolumn{2}{c}{Ours}\\
\cline{12-13}& & &&\multicolumn{2}{c}{Patch-16}&& & && Patch-16 & Patch-8 & Patch-16\\
\cline{2-3}\cline{5-6}\cline{8-9}\cline{11-13}

Average (MD)	&	58.5	&\textbf{	60.6			}&&	59.9			&\textbf{	61.4			}&&	\textbf{	57.3			}&	56.1	&&	\underline{64.0}			&\textbf{	64.1			}&	62.1	\\
\midrule
Average (MD+)	&	57.2			&\textbf{	58.9			}&&	59.2			&\textbf{	60.6			}&&	\textbf{	56.5			}&	55.2&&	\underline{62.2}			&\textbf{	62.6			}&	60.3	\\ 

\bottomrule 
\end{NiceTabular}
\caption{``Five-way One-shot'': summary of accuracy (in \%), compared between TSA/eTT and the MIV-head (Ours) using the same setting as in Tables~\ref{tab:miv_acc_tsa} and \ref{tab:miv_acc_ett}, except that ``varying-way varying-shot'' test datasets are replaced by ``5-way 1-shot'' ones. ``Average (MD)'' indicates the average accuracy across the nine original non-ILSVRC MD datasets, and ``Average (MD+)'' across the 17 extended MD datasets. Higher accuracy is bolded within each comparison. See Appendix~\ref{app:acc1shot} for details.} \label{tab:summ1shot}
\end{table}

\begin{figure}[!ht]
    \centering
     \includegraphics[width=0.95\linewidth]{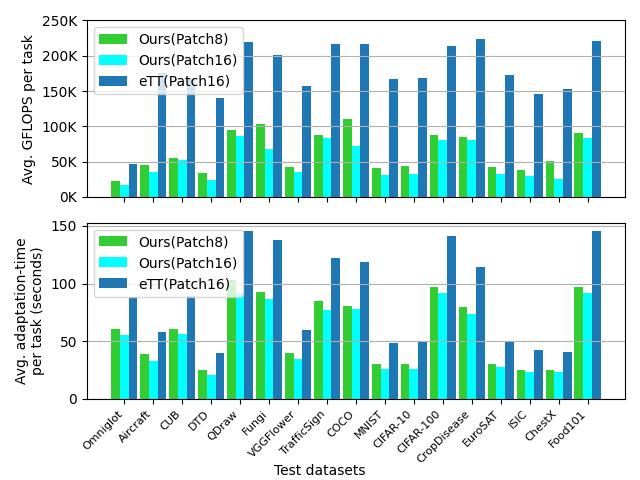}
     \caption{Comparison of adaptation cost between eTT and the MIV-head (Ours) based on non-ILSVRC datasets in the extended Meta-Dataset benchmark and the same self-supervised (DINO) backbone. The adaptation cost is measured by GFLOPs (upper panel) and training time (in seconds) per task using the same hardware (lower panel). Ours' adaptation cost is plotted by light-colored bars, showing that ours' GFLOPs is typically below $50\%$, and training time is $50\%$--$70\%$, of eTT's.}\label{fig:ett_cost_ssp}
\end{figure}

We additionally ran all algorithms in Tables~\ref{tab:miv_acc_tsa} and \ref{tab:miv_acc_ett}, aligned by the same backbones, in a ``five-way one-shot'' setting, following the standard practice of one-shot learning in this literature \citep{matchingnet16, tsa22, fes24}. The experimental results are summarized in Table~\ref{tab:summ1shot}, with details including 95\% confidence intervals reported in Table~\ref{tab:1shot} of Appendix~\ref{app:acc1shot}. Overall, the one-shot results exhibit a similar pattern as the ``varying-way varying-shot'' results: on average, ours are comparable to, and sometimes better than, TSA/eTT. In particular, when working with backbones pretrained using supervised learning, our approach consistently outperforms both TSA and eTT in all settings.

\subsubsection{Adaptation cost} \label{sec:results_cost}

A key factor impacting the adaptation cost is the number of passes over the backbone, possibly overshadowing other factors. In contrast to the baselines that pass through the backbone, forward and backward, at \emph{every} training step, our approach passes through only \emph{once} during training. To reflect this difference, we adopt two metrics to measure the computational cost of training during ``meta-testing'': ``Giga floating point operations (GFLOPs)'' and ``training time (in seconds)'' of adaptation. Note the latter is straightforward and particularly relevant to the near real-time requirements of the FSL use cases because it directly reflects how long users must wait for the adapted model. We compute the two metrics for all algorithms and find that all results exhibit a similar pattern. As such, we report the comparison between eTT and our approach based on self-supervised backbones in Figure \ref{fig:ett_cost_ssp} and display other results in Appendix~\ref{app:cost}. The MIV-head incurs substantially lower adaptation cost than eTT, typically $<50\%$ in GFLOPs and $50$--$70\%$ in training time.

\begin{sidewaystable}
\begin{NiceTabular}{@{\extracolsep{3pt}}l@{}c@{}c@{}cc@{}c@{}c@{\hskip3pt}c@{\hskip3pt}cc@{}c@{}c@{\hskip3pt}c@{\hskip3pt}c@{}}
\multicolumn{14}{c}{Self-supervised (DINO-pretrained) ViT-small backbones for $224\times224$ input resolution} \\ \toprule 
\Block[l]{3-1}{Test\\dataset} &\multicolumn{3}{c}{eTT (Patch 16)}& \multicolumn{5}{c}{Ours (Patch 8)}&\multicolumn{5}{c}{Ours (Patch 16)}\\
& Train & Infer & Total (3) & Train & Infer & Total (6) & Infer \% & \% of eTT & Train & Infer & Total (9) & Infer \% & \% of eTT\\
&(1) & (2) & =(1)+(2) & (4) & (5) & =(4)+(5) & =(5)/(6) & =(6)/(3) & (7) & (8) & =(7)+(8) & =(8)/(9) & =(9)/(3)\\ \cmidrule{2-4}\cmidrule{5-9}\cmidrule{10-14}

Omniglot	&	85.0	&	0.4	&	85.4	&	61.4	&	0.8	&	62.2	&	1.3\%	&	72.8\%	&	56.9	&	0.2	&	57.1	&	0.3\%	&	66.9\%	\\ 
Aircraft	&	55.6	&	0.2	&	55.8	&	36.9	&	0.4	&	37.3	&	1.1\%	&	66.8\%	&	34.6	&	0.2	&	34.8	&	0.5\%	&	62.3\%	\\ 
Birds	&	93.9	&	1.4	&	95.3	&	59.0	&	0.7	&	59.7	&	1.2\%	&	62.7\%	&	57.0	&	0.3	&	57.3	&	0.5\%	&	60.2\%	\\ 
Textures	&	38.0	&	0.1	&	38.1	&	24.1	&	0.3	&	24.3	&	1.1\%	&	63.8\%	&	21.7	&	0.1	&	21.8	&	0.3\%	&	57.1\%	\\ 
Quick Draw 	&	149.7	&	1.5	&	151.2	&	95.4	&	1.2	&	96.6	&	1.3\%	&	63.9\%	&	96.8	&	0.3	&	97.1	&	0.3\%	&	64.2\%	\\ 
Fungi	&	140.2	&	1.2	&	141.4	&	91.4	&	0.5	&	91.8	&	0.5\%	&	64.9\%	&	89.2	&	0.2	&	89.4	&	0.2\%	&	63.2\%	\\ 
VGG Flower	&	60.6	&	0.7	&	61.3	&	38.9	&	0.5	&	39.3	&	1.1\%	&	64.2\%	&	35.5	&	0.1	&	35.6	&	0.3\%	&	58.1\%	\\ 
Traffic Sign	&	124.1	&	1.3	&	125.4	&	86.1	&	1.0	&	87.1	&	1.2\%	&	69.5\%	&	80.3	&	0.2	&	80.6	&	0.3\%	&	64.2\%	\\ 
MSCOCO	&	121.1	&	1.3	&	122.4	&	78.7	&	1.0	&	79.7	&	1.2\%	&	65.1\%	&	76.1	&	0.2	&	76.4	&	0.3\%	&	62.4\%	\\ 
MNIST	&	45.9	&	0.2	&	46.0	&	29.3	&	0.3	&	29.6	&	1.1\%	&	64.3\%	&	26.3	&	0.1	&	26.3	&	0.3\%	&	57.2\%	\\ 
CIFAR-10	&	46.3	&	0.2	&	46.4	&	30.6	&	0.3	&	30.9	&	1.0\%	&	66.6\%	&28.8	&	0.1	&	28.9	&	0.3\%	&	62.1\%	\\ 
CIFAR-100	&	137.9	&	0.5	&	138.4	&	93.4	&	1.3	&	94.7	&	1.4\%	&	68.4\%	&94.3	&	0.3	&	94.6	&	0.3\%	&	68.3\% \\ 
CropDisease	&	110.5	&	0.5	&	111.0	&	77.6	&	0.9	&	78.5	&	1.2\%	&	70.7\%	&	73.5	&	0.2	&	73.7	&	0.3\%	&	66.4\%	\\ 
EuroSAT	&	51.8	&	0.6	&	52.4	&	31.8	&	0.3	&	32.2	&	1.1\%	&	61.4\%	&	27.6	&	0.1	&	27.7	&	0.3\%	&	52.9\%	\\ 
ISIC	&	43.3	&	0.3	&	43.5	&	25.1	&	0.3	&	25.3	&	1.0\%	&	58.2\%	&	22.2	&	0.1	&	22.2	&	0.3\%	&	51.1\%	\\ 
ChestX	&	42.8	&	0.4	&	43.2	&	24.0	&	0.3	&	24.2	&	1.1\%	&	56.1\%	&	21.7	&	0.1	&	21.8	&	0.3\%	&	50.5\%	\\ 
Food101	&	140.6	&	0.5	&	141.2	&	94.3	&	1.4	&	95.7	&	1.4\%	&	67.8\%	&95.4	&	0.3	&	95.7	&	0.3\%	&	67.8\%	\\ 

\bottomrule 
\end{NiceTabular}
\caption{Total latency time per task (in seconds), broken down by training (i.e., adaptation) and inference time, compared between eTT and the MIV-head (Ours) within the same setting as Figure~\ref{fig:ett_cost_ssp}. Columns labeled by ``Train'' (i.e., (1), (4), (7)) report the training time, ``Infer'' ((2), (5), (8)) reporting the inference time, ``Total'' ((3), (6), (9)) displaying the total latency-time combining both ``Train'' and ``Infer''. Columns labeled ``Infer \%'' (=(5)/(6) or =(8)/(9)) calculate the proportion of inference time in total latency-time. Columns labeled ``\% of eTT'' (=(6)/(3) or =(9)/(3)) present the ratio of total latency-time between ours and eTT.} \label{tab:latency}
\end{sidewaystable}

To consider total latency time that would be incurred in practice, which also includes inference time, we extend the analysis of Figure~\ref{fig:ett_cost_ssp} to the measure of average ``total latency time'' per task (in seconds)---combining both training and inference time---in Table~\ref{tab:latency}. The conclusion drawn from Table~\ref{tab:latency} is virtually unchanged from that of Figure~\ref{fig:ett_cost_ssp}: the MIV-head still incurs only $50$--$70\%$ of eTT's total latency time (see columns of ``\% of eTT''). This is unsurprising given that inference time is typically approximately $1\%$ or less of the total latency time, see columns labeled ``Infer \%'' in Table~\ref{tab:latency}.

\subsection{Additional analysis}\label{sec:extra_results}

\begin{table}[t]
\centering
\begin{NiceTabular}{lcccccc}
    \Block[l]{2-1}{Backbone\\architecture}
     & \Block[c]{2-1}{Pretraining\\data} & \Block[c]{2-1}{Pretraining\\algorithm} && \multicolumn{3}{c}{Average (MD)} \\
    \cmidrule{5-7}  &  &  && Ours && Baseline++ \\ \toprule

    DenseNet-161	&	ILSVRC	&	supervised	&&	77.2	&&	68.8	\\ 
    RegNetY-1.6GF	&	ILSVRC	&	supervised	&&	77.4	&&	64.3	\\ 
    ViT-B/16	&	ILSVRC	&	DeiT	&&	78.1	&&	71.0	\\ 
    Swin Transformer	&	ILSVRC	&	SimMIM	&&	78.2	&&	68.4	\\ 
    ViT-B/16	&	ILSVRC	&	DINO	&&	78.9	&&	73.8	\\ 
    ViT-L/14	&	WebImageText	&	CLIP	&&	83.7	&&	77.5	\\ \bottomrule
    
\end{NiceTabular}
\caption{Average accuracies (in \%) of the MIV-head (Ours) and Baseline++ with a diverse range of backbones, based on the (non-ILSVRC) MD benchmark. For ViT backbones, we follow the standard naming conventions to denote them as ``ViT-[size]/[patch]'' where [size] is either ``B''(Base) or ``L''(Large) representing model sizes, and [patch] $\in \{8,14,16\}$ representing the input patch sizes.}\label{tab:miv_diversebb}
\end{table}

\subsubsection{Additional backbones for our approach} \label{sec:more_bb}

To demonstrate that our approach is backbone-agnostic, we present experiments with the MIV-head on a diverse range of backbones, including CNNs (DenseNet~\citep{densenet17} and RegNet~\citep{regnet20}), ViTs, and the Swin Transformer~\citep{swin21}, with supervised pretraining (DeiT~\citep{deit21} and CNNs' standard supervised pretraining), self-supervised pretraining (DINO~\citep{dino21} and SimMIM~\citep{simmim22}) and contrastive pretraining (CLIP vision model~\citep{clip21}). The results are summarized in Table~\ref{tab:miv_diversebb}, with details in Table~\ref{tab:morebb} of Appendix~\ref{app:more_bb}. We also include the performance of Baseline++~\citep{closerfewshot19} on the same set of backbones for comparison.

While the backbones used in Table~\ref{tab:miv_diversebb} are among the best-performing ones, as shown by~\citep{closerlook23}, they \emph{cannot} be leveraged by TSA/eTT in the ``varying-way varying-shot'' setting, either due to a lack of adaptation recipes (for TSA) or computational infeasibility (for eTT\footnote{\citet{closerlook23} reported results of eTT on the relevant backbones only in the ``5-way 5-shot'' setting---such constraints on task type may not be practical in real-world use cases.}). Baseline++, a well-known ``head'' approach, supports all backbones but lags far behind in accuracy as shown by Table~\ref{tab:miv_diversebb}. In contrast, it is apparent that our approach works well with a diverse range of backbones, achieving performance similarly strong to that reported in Tables~\ref{tab:miv_acc_tsa} and\ref{tab:miv_acc_ett}---the results are comparable to, and sometimes better than, the SOTA.

\subsubsection{Another baseline: ``LN-Tune''}\label{sec:ln_tune}

\begin{table}[t]
\hspace*{-2cm}
\begin{minipage}{.66\textwidth}
\begin{NiceTabular}{@{}l@{}c@{\hskip3pt}c@{\hskip3pt}cc@{}c@{\hskip3pt}c@{\hskip3pt}c@{}}
\multicolumn{8}{c}{Off-the-shelf pretrained ViT-small backbones ($224\times224$)} \\ \toprule
\Block[l]{3-1}{Test\\dataset} & \multicolumn{3}{c}{Supervised(DeiT) } & & \multicolumn{3}{c}{Self-supervised(DINO)}\\ 
&LN-Tune&eTT& Ours&&LN-Tune&eTT & Ours\\
&\multicolumn{3}{c}{(Patch 16)}&&\multicolumn{2}{c}{(Patch 16)}& (Patch 8)\\\cmidrule{2-4}\cmidrule{6-8}

Omniglot	&	70.7$\pm$1.5	&	69.4	&\textbf{	74.5	}&&	76.3$\pm$1.2	&	77.0	&\textbf{	78.6	}\\ 
Aircraft	&	76.0$\pm$1.1	&	78.4	&\textbf{	79.4	}&&	78.9$\pm$1.0	&	84.1	&\textbf{	86.0	}\\ 
Birds	&	83.9$\pm$0.8	&\textbf{	88.9	}&	88.7	&&	91.2$\pm$0.6	&\textbf{	92.2	}&	91.8	\\ 
Textures	&	85.9$\pm$0.7	&	86.8	&\textbf{	88.3	}&&	88.7$\pm$0.6	&	89.3	&\textbf{	89.7	}\\ 
Quick Draw 	&		OOM		&	69.2	&\textbf{	70.9	}&&		OOM		&\textbf{	72.4	}&	71.8	\\ 
Fungi	&	55.7$\pm$1.1	&	59.7	&\textbf{	60.9	}&&	65.4$\pm$1.1	&	65.1	&\textbf{	65.6	}\\ 
VGG Flower	&	93.8$\pm$0.6	&	93.6	&\textbf{	94.7	}&&	96.8$\pm$0.3	&\textbf{	97.4	}&	97.3	\\ 
Traffic Sign	&\textbf{	77.0$\pm$1.3	}&	70.6	&	68.2	&&\textbf{	82.5$\pm$1.1	}&	81.3	&	79.6	\\ 
MSCOCO	&	64.7$\pm$0.9	&	65.2	&\textbf{	65.7	}&&\textbf{	67.2$\pm$0.9	}&	64.4	&	63.0	\\ \midrule 
Average	&	76.0			&	76.6	&\textbf{	77.5	}&&	80.9			&	81.3	&\textbf{	81.5	}\\ 

\bottomrule 
\end{NiceTabular}
\caption{\normalsize Comparison of accuracy (in \%), between LN-Tune, eTT and the MIV-head(Ours), based on both supervised (DeiT) and self-supervised (DINO) ViT-small backbones, and the (non-ILSVRC) MD benchmark. ``OOM'' indicates that we were unable to collect LN-Tune results due to OOM errors. For all algorithms, the calculation of average accuracy excludes the ``Quick Draw'' dataset where OOM occurred. Higher accuracy is bolded within each comparison.}\label{tab:lntune_acc}
\end{minipage}%
\hspace*{5mm}
\begin{minipage}{.5\textwidth}
\begin{NiceTabular}{@{}l@{}c@{\hskip3pt}c@{\hskip9pt}c@{\hskip6pt}c@{\hskip6pt}c@{}}
\multicolumn{6}{c}{Supervised (DeiT-small/16) backbones} \\ \toprule
\Block[l]{3-1}{Test\\dataset} & \Block[c]{2-1}{eTT} & LN- & \Block[c]{2-1}{Ours} & \% of  & \% of LN\\
& & Tune & & eTT= & -Tune=\\
& (1) & (2) & (3) & (3)/(1) & (3)/(2)\\\midrule

Omniglot	&	90.7	&	63.5	&	55.3	&	61\%	&	87\%	\\ 
Aircraft	&	60.1	&	64.3	&	33.7	&	56\%	&	52\%	\\ 
Birds	&	89.0	&	79.7	&	55.0	&	62\%	&	69\%	\\ 
Textures	&	40.3	&	48.9	&	21.5	&	53\%	&	44\%	\\ 
Quick Draw 	&	147.2	&	OOM	&	95.8	&	65\%	&	 ---	\\ 
Fungi	&	134.9	&	113.2	&	87.5	&	65\%	&	77\%	\\ 
VGG Flower	&	59.9	&	63.8	&	35.6	&	59\%	&	56\%	\\ 
Traffic Sign	&	120.6	&	109.7	&	78.3	&	65\%	&	71\%	\\ 
MSCOCO	&	116.9	&	108.8	&	76.2	&	65\%	&	70\%	\\ 

\bottomrule 
\end{NiceTabular}
\caption{\normalsize Comparison of adaptation (i.e., training) time, between LN-Tune, eTT, and the MIV-head(Ours), based on supervised (DeiT-small) backbones and the (non-ILSVRC) MD benchmark. Columns labeled ``\% of eTT'' and ``\% of LN-Tune'' compute the ratios of ours' adaptation time to that of eTT and LN-Tune, respectively. ``OOM'' indicates that we were unable to collect LN-Tune results due to OOM errors.
}\label{tab:lntune_cost}
\end{minipage}
\end{table}

For ViT backbones, several high-performing ``partial fine-tuning'' methods have recently been identified that are comparable to eTT. For example, \citet{vitbaseline24} showed that a simple baseline that only fine-tunes the ``Layer Normalization'' modules within the ViT backbones, referred to as ``LN-Tune'', is one of the best-performing ``parameter-efficient fine-tuning'' (PEFT) methods. To compare to this approach, we also experimented with LN-Tune using both supervised and self-supervised backbones on the (non-ILSVRC) MD benchmark, using the same data augmentation as in the MIV-head, subject to hardware constraints. The accuracy and adaptation cost (training time) of this method, tabulated in Tables~\ref{tab:lntune_acc} and \ref{tab:lntune_cost} respectively, are similar to those of eTT (slightly worse in accuracy). Importantly, despite the small number of parameters that need to be fine-tuned by LN-Tune, we find that this method is similarly costly to train compared to eTT---our training time is around 40\%-80\% of LN-Tune's, while being 50\%-70\% of eTT's, based on the same supervised backbones (\emph{cf.} Table~\ref{tab:lntune_cost}).

\subsection{Ablation study} \label{sec:ablation}

\subsubsection{What is the individual contribution of each core component to the MIV-head?} \label{sec:abl_alone}

\begin{minipage}{1.1\textwidth}
\centering
\hspace*{-2cm}
\begin{minipage}{.45\textwidth} 
\includegraphics[width=\textwidth]{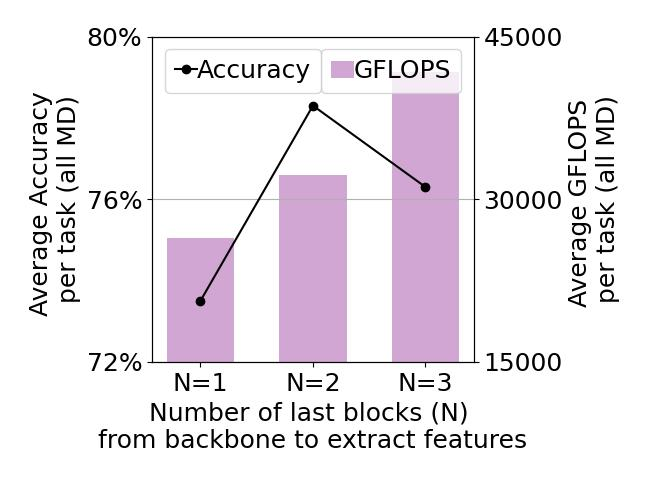}
\captionof{figure}{Accuracy (line on left y-axis) and GFLOPs (bar on right y-axis) by varying the number of output blocks ($N=1,2,3$) of the backbone. Both metrics are averaged over all test tasks, across all 9 (non-ILSVRC) datasets of the MD benchmark.} \label{fig:abl_blk}
\end{minipage}
\hfill
\begin{minipage}{.65\textwidth}
\begin{NiceTabular}{@{}l@{}c@{\hskip6pt}c@{\hskip9pt}c@{\hskip6pt}c@{\hskip3pt}c@{\hskip3pt}c@{\hskip1pt}c@{}} 
& 1 &2 &3 &4 &5 &6 &7 \\ \cmidrule{2-8}
\multirow{4}{1cm}{Test dataset} & & Replace & \multicolumn{5}{c}{Patch-level embeddings (D}\\
& Full & CAP & \multicolumn{5}{c}{candidates) to image-level}\\
\cline{4-8}
&  model & by avg- & GAP & \multicolumn{4}{c}{Pooling by attention}\\ \cline{5-8}
& & pooling & (D=1) & D=1 & D=2 & D=4 & D=6 \\ \toprule

Omniglot	&	76.5	&	68.9	&	73.5	&	74.6	&	75.1	&	75.6	&	75.8	\\
Aircraft	&	84.4	&	78.1	&	80.6	&	83.5	&	84.5	&	84.4	&	84.5	\\
Birds	&	87.2	&	87.2	&	82.0	&	85.0	&	86.0	&	87.2	&	87.2	\\
Textures	&	89.0	&	86.0	&	88.9	&	88.9	&	89.1	&	89.0	&	89.0	\\
Quick Draw 	&	71.7	&	66.5	&	70.2	&	70.9	&	71.4	&	71.7	&	71.8	\\
Fungi	&	60.6	&	57.8	&	56.6	&	59.2	&	60.0	&	60.5	&	60.5	\\
VGG Flower	&	96.0	&	95.6	&	94.9	&	95.4	&	95.7	&	96.1	&	96.1	\\
Traffic Sign	&	78.6	&	74.4	&	68.9	&	77.1	&	77.0	&	78.5	&	77.4	\\
MSCOCO	&	60.8	&	58.2	&	60.1	&	60.7	&	60.2	&	60.6	&	60.5	\\\midrule
Average	&	78.3	&	74.7	&	75.1	&	77.3	&	77.7	&	78.2	&	78.1	\\\bottomrule 
\end{NiceTabular}

\captionof{table}{Ablation analyses of the MIV-head based on the supervised ResNet-50 backbone and the (non-ILSVRC) MD benchmark, performed by replacing a component of interest or varying its hyperparameters.} \label{tab:abl} 
\end{minipage}
\end{minipage}

We disentangle the marginal effects of the three core components of the MIV-head based on the original MD benchmark (nine datasets) and the supervised ResNet-50 backbone. More concretely, from the ``full model'' used in Table~\ref{tab:miv_acc_tsa} we modify each component, one at a time, either replacing it by an alternative strategy or varying a key hyperparameter.

Figure~\ref{fig:abl_blk} plots the accuracy and GFLOPs (on the left and right axis, respectively) at different values of $N$, the hyperparameter specifying the ``number of output blocks of the backbone'' in Component 3. These results show that extracting features from the last 2 blocks ($N=2$) gives better accuracy than $N=1$, at the cost of additional computation (GFLOPs). $N=3$ adds no value while incurring unnecessarily higher cost. Therefore, we set $N=2$ for ResNet (see more details broken down by datasets in Appendix~\ref{app:abl_blk}). 

Given $N=2$, Table~\ref{tab:abl} analyzes Component 1, ``pooling by attention'' (columns 3--7) and Component 2, CAP (column 2), relative to the full model of the MIV-head (column 1). If we replace CAP with average-pooling---in other words, prototypes are created by averaging feature vectors of the same-class support samples---there is considerable deterioration in performance. Column 2 thus demonstrates the substantial contribution of CAP. Next, we vary the hyperparameter $D$\footnote{Here, $D$ is the same across different blocks, and thus has no subscript $n$.} in Component 1, i.e., the number of candidates for an image's patch-level representations. A special case is $D=1$ where we only need Equation~\eqref{ppool1} but not \eqref{ppool2}---in this case, we can pool the patch-level representations by Equation~\eqref{ppool1} (column 4), or instead by a global average pooling (GAP, column 3). Results for $D>1$ are in columns 5--7. Clearly, as part of ``pooling by attention'', Equation~\eqref{ppool1} contributes substantially (4 vs. 3), and Equation~\eqref{ppool2} also brings nontrivial benefit ($D>1$ vs. $D=1$). Given the presence of ``pooling by attention'', higher values of $D$ tend to add value, but returns diminish when $D>4$. Overall, performance is robust to varying $D$ when $D>1$.

\subsubsection{Does a component's contribution depend on the co-existence of other components?} \label{sec:abl_dependence}

While the above analysis shows substantial marginal effects from each individual component---assuming the existence of other components, it is also crucial to understand how such co-existence would impact the effects of the MIV-head's core components. To this end, we further investigate the dependence between multiple core components, by analyzing them jointly and offering insights into their interactions. Tables~\ref{tab:intx} and \ref{tab:intx2} present the results, revealing how a component behaves conditionally on the presence or absence of other components.

\begin{table}[t]
\begin{subtable}[t]{0.45\linewidth}
\centering
\begin{tabular}{@{}lcccc} 
 & 1 & 2 & 3 & 4 \\ \toprule
$N=2$ & &  & $\checkmark$ & $\checkmark$ \\
Pooling-by-attention & &  $\checkmark$  & & $\checkmark$ \\
\midrule
Omniglot	&	59.7	&	61.0	&	71.1	&	73.2	\\
Aircraft	&	71.8	&	73.3	&	79.9	&	83.6	\\
Birds	&	84.6	&	86.4	&	81.7	&	86.7	\\
Textures	&	86.1	&	86.7	&	88.7	&	88.8	\\
Quick Draw 	&	62.8	&	63.5	&	69.1	&	70.3	\\
Fungi	&	49.5	&	50.7	&	53.2	&	56.4	\\
VGG Flower	&	90.4	&	91.6	&	94.2	&	95.3	\\
Traffic Sign	&	61.0	&	62.9	&	66.0	&	74.8	\\
MSCOCO	&	55.8	&	56.2	&	56.9	&	57.1	\\\midrule
Average (MD)	&	69.1	&	70.3	&	73.4	&	76.3	\\

\end{tabular}
\caption{Accuracy considering different combinations of pooling-by-attention and $N$.} \label{tab:intx_blkpool}
\end{subtable}
\hfill
\begin{subtable}[t]{0.48\linewidth}
\vspace*{-2cm}
\centering
\begin{NiceTabular}{@{}l@{\hskip20pt}lc@{\hskip20pt}l@{}}
\toprule
 & \multicolumn{3}{c}{Pooling-by-attention} \\
 & Disabled & & \multirow{2}{*}{Enabled} \\
 & (i.e., GAP) & & \\ \cmidrule{2-4}
$N=1$ & 69.1\% & $\underrightarrow{(+1.2\%)}$ & 70.3\% \\
 & $\downarrow(+4.3\%)$ & & $\downarrow(+6\%)$ \\
$N=2$ & $73.4\%$ & $\underrightarrow{(+2.9\%)}$ & $76.3\%$ \\
\bottomrule 
\end{NiceTabular}
\caption{Summarization of Table~\ref{tab:intx_blkpool} about the interactions between pooling-by-attention (disabled, enabled) and $N$ ($N=1,2$). Disabling pooling-by-attention implies replacing it with global average pooling (GAP).} \label{tab:intx_blkpool_summ}
\end{subtable}
\caption{Interactions between the two core components of the MIV-head, ``pooling-by-attention'' and ``number of output blocks $N$'', based on the supervised ResNet-50 backbone and the (non-ILSVRC) MD benchmark. Note that in this analysis, CAP is enabled in the MIV-head but data augmentation is disabled.}\label{tab:intx}
\end{table}

\begin{table}[t]
\begin{subtable}[t]{0.45\linewidth}
\begin{tabular}{@{\extracolsep{2pt}}l@{}cccc} 
 & 1 & 2 & 3 & 4 \\ \toprule
$N=2$ & & & $\checkmark$ & $\checkmark$\\
Pooling-by-attention & & $\checkmark$ & & $\checkmark$ \\
CAP & & $\checkmark$ & & $\checkmark$  \\ \midrule

Omniglot	&	50.4	&	61.0	&	61.2	&	73.2	\\
Aircraft	&	54.7	&	73.3	&	56.9	&	83.6	\\
Birds	&	81.0	&	86.4	&	75.9	&	86.7	\\
Textures	&	83.7	&	86.7	&	86.1	&	88.8	\\
Quick Draw 	&	57.2	&	63.5	&	61.2	&	70.3	\\
Fungi	&	44.9	&	50.7	&	47.5	&	56.4	\\
VGG Flower	&	86.0	&	91.6	&	92.1	&	95.3	\\
Traffic Sign	&	49.7	&	62.9	&	49.3	&	74.8	\\
MSCOCO	&	57.0	&	56.2	&	52.1	&	57.1	\\\midrule
Average (MD)	&	62.7	&	70.3	&	64.7	&	76.3	\\

\end{tabular}
\caption{Accuracy considering different combinations of ``pooling-by-attention + CAP'' and $N$.}
\label{tab:intx_blkpoolcap}
\end{subtable}
\hfill
\begin{subtable}[t]{0.48\linewidth}
\vspace*{-2cm}
\begin{NiceTabular}{@{}l@{\hskip20pt}lc@{\hskip20pt}l@{}}
\toprule
 & \multicolumn{3}{c}{Pooling-by-attention + CAP} \\
 & Disabled & & \multirow{2}{*}{Enabled} \\
 & (i.e., GAPs) & & \\ \cmidrule{2-4}
$N=1$ & 62.7\% & $\underrightarrow{(+7.6\%)}$ & 70.3\% \\
 & $\downarrow(+2\%)$ & & $\downarrow(+6\%)$ \\
$N=2$ & $64.7\%$ & $\underrightarrow{(+11.6\%)}$ & $76.3\%$ \\
\bottomrule 
\end{NiceTabular}
\caption{Summary of Table~\ref{tab:intx_blkpoolcap} about the interactions between ``pooling-by-attention + CAP''(disabled, enabled) and $N$ ($N=1,2$). Disabling ``pooling-by-attention + CAP'' implies replacing each component with global average pooling (GAP).}\label{tab:intx_blkpoolcap_summ}
\end{subtable}
\caption{Interactions between the core components of the MIV-head, ``pooling-by-attention + CAP'' and ``number of output blocks $N$'', based on the supervised ResNet-50 backbone and the (non-ILSVRC) MD benchmark. Note that in this analysis, data augmentation is disabled.} \label{tab:intx2}
\end{table}

Table~\ref{tab:intx} considers the combination of two core components of the MIV-head, pooling-by-attention (disabled or enabled) and number of output blocks $N$ ($N=1,2$), while enabling CAP at all times. Disabling pooling-by-attention implies replacing it with GAP---same as column 3 of Table~\ref{tab:abl}. Table~\ref{tab:intx_blkpool} tabulates the accuracies of all four combinations in column 1--4 (with CAP enabled for all columns). Table~\ref{tab:intx_blkpool_summ} summarizes those accuracies to demonstrate the interaction effects. The results show that the benefit of a component depends strongly on the presence (absence) of another component. For example, when $N=1$, pooling-by-attention brings only a marginal increase ($+1.2\%$) of performance, whereas at $N=2$, there is a substantial increase ($+2.9\%$) attributed to pooling-by-attention. Likewise, conditional on the presence and absence of pooling-by-attention, the incremental accuracy caused by $N=2$ vs. $N=1$ varies considerably---$+6\%$ with pooling-by-attention and $+4.3\%$ without.

The interaction effects are more pronounced when we combine the two components ``pooling-by-attention and CAP'' together (rather than fixing CAP at the ``enabled'' state as in Table~\ref{tab:intx}). The interactions between ``pooling-by-attention + CAP'' (disabled or enabled) and $N$ ($N=1,2$) are shown in Table~\ref{tab:intx2}, which exhibits a similar tendency as Table~\ref{tab:intx}, with a greater magnitude. For example, when disabling both pooling-by-attention and CAP, there is only a minor contribution from $N=2$ (column 3 in~\ref{tab:intx_blkpoolcap}) vs. $N=1$ (column 1 in~\ref{tab:intx_blkpoolcap})---$+2\%$ on average---and the performance uplift is inconsistent across datasets. But the benefit becomes substantial and consistent (column 4 vs. 2 in Table~\ref{tab:intx_blkpoolcap}) when enabling ``pooling-by-attention + CAP'': $+6\%$ on average (Table~\ref{tab:intx_blkpoolcap_summ}). Similarly, the contribution from ``pooling-by-attention + CAP'' is higher at $N=2$ than at $N=1$.

Tables~\ref{tab:intx} and \ref{tab:intx2} demonstrate strong dependence between the core components of the MIV-head, and thus show their collective contribution to the efficacy of our approach. This analysis highlights the importance of employing the three core components together in our design of the MIV-head.

In summary, all the core components of the MIV-head collectively account for its good performance. Additional ablation analyses are reported in Appendix~\ref{app:abl}. In Appendix~\ref{app:abl_single}, we present more fine-grained ablation, including the mechanisms within Component 2 (CAP) and the data augmentation strategy. Although less significant, those elements improve performance. Appendix~\ref{app:abl_rn18} sheds light on the impact of off-the-shelf vs. specially-trained backbones on adaptation approaches.

\section{Conclusion, limitations and future work} \label{sec:concl}

This paper focuses on adaptation during meta-testing in CDFSL. Inspired by the representation of an FSC problem as a series of MIV tasks, we propose a novel cross-domain adaptation framework, named the ``MIV-head'', and implement it as a few-shot classification head. We demonstrate that, while enjoying the benefits of being a ``head'' approach---being backbone-agnostic and computationally efficient---the MIV-head is unusual among such approaches in being highly competitive with SOTA adapter (or fine-tuning) methods while incurring substantially lower adaptation cost. To the best of our knowledge, our study is the first to successfully tackle CDFSL when ``backbone fine-tuning'' is impossible, and the first to address such challenges using a ``verification with noise'' paradigm (MIV).

The backbone-agnostic nature of our design can also be viewed as an attempt, in CDFSL, to take embeddings as the starting point of learning. An embedding is clearly a more abstract representation than the raw input (i.e., an image) considered in the existing literature. This abstraction creates both benefits and challenges. The benefits include:
\begin{enumerate}[label=\roman*)]
    \item More practical deployment by obviating the need for knowledge of the backbone's model architecture---this is particularly appealing for users of cloud services, API-based models or vector databases; 
    \item Greater flexibility, allowing combinations of backbones, modalities, or input structures. This offers research opportunities that link to emerging areas like multi-modal few-shot learning \citep{multimodal24}, or new applications requiring complex representations of inputs (queries and/or support set)---for example, bags of objects, bags of bags, nested or composite structures.
    \item Easier generalization, leading to broader use cases or problems. This aspect facilitates extensions of CDFSL that, for example, can handle scenarios with greater differences between domains, such as different feature extraction procedures or different semantics of embeddings, between the target and source domains.
\end{enumerate}

Considering few-shot domain transfer in particular, learning directly from frozen embeddings introduces new challenges such as issues with the quality of the embeddings, misalignment of embedding semantics, effects on the explainability of the model's predictions, and diversity of problem representations and solution architectures. Despite our attempt to tackle some of the challenges by treating few-shot classification as ``verification with noise'', there is scope for further work.

Our work has several limitations. First, it does not explore meta-training---the well-known approach to tackle CDFSL through pre-training. However, as a native episodic learner, our approach is compatible with any meta-training pipeline. While we train the MIV-head from scratch during meta-testing, this practice may be suboptimal and carefully designed meta-training may add value to learning the MIV-head parameters. Second, the MIV-head may fail in some situations. In particular, it underperforms on low-quality images in our experiments---e.g., the CIFAR datasets that were selected from the ``80 million Tiny Images'' dataset, whose images were down-sampled to $32\times32$ resolution from their original high-resolution versions\footnote{See \url{https://cave.cs.toronto.edu/kriz/cifar.html} and \url{http://people.csail.mit.edu/torralba/tinyimages/}}. A plausible explanation is that the embeddings provided by a frozen backbone do not contain sufficient latent information to enable successful discrimination, so methods that adapt the backbone itself, such as TSA and eTT, may be necessary. Third, the design of the MIV-head is mostly empirically justified, and a fruitful avenue of future research would be to develop a strong theory. Fourth, the architectural complexity of the MIV-head is relatively high compared to a standard classification head. While higher complexity may be necessary to account for frozen/static embeddings, there could be alternative solutions that are simpler and more explainable. Fifth, in theory, the MIV-head has $O(|\mathcal{X}| \times |\mathcal{Y}|)$ inference complexity due to cross-attention, where $\mathcal{X},\mathcal{Y}$ are the query-set and support-set respectively. This could present a potential concern for inference scalability when $\mathcal{X}$ and $\mathcal{Y}$ become very large. 
In practice, however, there are several mitigating factors: (i) $O(|\mathcal{X}| \times |\mathcal{Y}|)$ is relevant only to the MIV-head---more precisely, Components 2 and 3---\emph{not} the backbone. Typically, as a ``classification head'', the MIV-head's inference time will be overshadowed by that of the backbone, for typical sizes of $\mathcal{X}$ and $\mathcal{Y}$. (ii) In real-world use cases, inference time needs to be considered in the context of total latency. In particular, the MIV-head's training and inference may re-use the same static embeddings retrieved from the backbone, and thus substantially decrease the total latency by obviating the need for expensive backbone evaluations. (iii) The fact that the MIV-head treats each query independently makes processing across queries embarrassingly parallel.

Finally, there are additional topics for future work: First, it may be interesting to explore or compare alternative problem representations and their solutions, in terms of complexity, performance, scalability to larger datasets, etc. Second, one may use multi-domain backbones, as opposed to single-domain ones (trained on ImageNet1K) used here. Third, the MIV-head's advantage of lower computational cost may need to be re-validated if more efficient fine-tuning techniques emerge.

\backmatter





\bmhead*{Acknowledgements}
We thank Hongyu Wang who kindly shared the data repository of the extended MD benchmark. We also thank the anonymous reviewers and the editor for their helpful comments that enabled us to improve our manuscript. 

\section*{Declarations}


\begin{itemize}
\item Funding: Xin Xu acknowledges financial support from the Machine Learning Group of the Computer Science Department at the University of Waikato.
\item Conflict of interest/Competing interests: On behalf of all authors, the corresponding author states that there is no conflict of interest.
\item Ethics approval and consent to participate: Not applicable
\item Consent for publication: Not applicable
\item Data availability: All data used can be acquired publicly via (1) \url{https://github.com/google-research/meta-dataset} for the official Meta-Dataset, (2) \url{https://github.com/cambridge-mlg/cnaps} for three additional target domains (MNIST, CIFAR-10, CIFAR-100), (3) \url{https://github.com/IBM/cdfsl-benchmark} for four additional target domains (EuroSAT, ISIC, CropDisease, ChestX), (4) \url{https://data.vision.ee.ethz.ch/cvl/datasets_extra/food-101/} for one additional target domain (Food101).
\item Materials availability: Not applicable
\item Code availability: The source code and instructions for executing the code (in README.md), needed to reproduce the main experimental results from this manuscript, are shared at: \url{https://github.com/xxweka/MIV-head}.
\item Author contribution: All authors contributed to the conception, methodology and design of the study. Eibe Frank first observed the similarity between the representations of FSC and MIV problems, leading to the key idea of this paper---solving CDFSL within a ``verification with noise'' paradigm (MIV). Material preparation, data collection, and analysis were performed by Xin Xu. The initial draft of this manuscript was prepared by Xu, and all authors reviewed, revised, and commented on subsequent versions of the manuscript. All authors read and approved the final manuscript.
\end{itemize}


\clearpage
\begin{appendices}
\appendixpage

\section{Formulation of Cross-Attention Pooling (CAP) (Section~\ref{sec:cap})} \label{app:cap_maths}
Here, we provide the formulation of CAP. For more details, see~\citep{miv25}. 

To recap our notation, for the $n^{th}$ block, we can write the CAP formulation as $v^{P_l}_n, v^Q_n = CAP_n(P^l_n, Q_n)$, where $P^l_n\in\mathbb{R}^{S^l\times C_n}$ stands for the bag of embeddings of $S^l$ images in the $l^{th}$ class. $Q_n\in\mathbb{R}^{1\times C_n}$ denotes a query image's feature-vector; $v^{P_l}_n, v^Q_n\in \mathbb{R}^{1\times C_n}$ denote the prototype (of the $l^{th}$-class) and the query representations, respectively. For ease of exposition, we omit subscripts (of $n$) and superscripts (of $l$), with the understanding that CAP is applied to the $n^{th}$ block and the $l^{th}$ class. 

Letting $LN(\cdot)$ be layer normalization~\citep{layernorm16}, $v^{query} = LN(Q)$, $v^{target} = LN(P)$, $cas(\cdot,\cdot)$ be the ``cross-attention score'' function, and $MHCE(\cdot)$ be ``multi-head co-excitation'', we model the two outputs of $CAP_n(\cdot, \cdot)$ as,
\begin{subequations} \label{perhead}
\begin{align}
v^P &= concat(O^P_1, O^P_2, \dots, O^P_h),\qquad v^Q = concat(O^Q_1, O^Q_2, \dots, O^Q_h), \nonumber \\
\end{align}
\begin{align}
O^P_j &= MatMul\Bigg[\underbrace{cas_j\Big(v^{query}W_j^K, v^{target}W_j^K\Big)}_{1\times S}, \underbrace{\Big(v^{target}W_j^V \odot MHCE_j(v^{query}) + v^{target}W_j^K\Big)}_{S\times d}\Bigg], \label{perhead_p} \\ 
O^Q_j &= \underbrace{v^{query}W_j^V \odot MHCE_j(v^{query}) + v^{query}W_j^K}_{1\times d}, \label{perhead_q} \qquad j = 1,2,\dots,h
\end{align}
\end{subequations}
where $h$ is the number of heads, $d=\frac{C}{h}$, $W_j^K, W_j^V \in \mathbb{R}^{C\times d}$ are two learnable weights of the multi-head linear projections, and the subscript $j$ of $cas_j(\cdot,\cdot)$ and $MHCE_j(\cdot)$ indicates the $j^{th}$ head. $MatMul$ and $\odot$ denote matrix and element-wise multiplication, respectively.

\subsection{Cross-attention, $cas(\cdot,\cdot)$} \label{sec:cas}

More precisely, omitting the subscript $j$ for brevity, we formulate $cas(\cdot,\cdot)$ as an L1-distance-based attention (DBA) function proposed by~\citet{miv25},
\begin{equation}\label{dba}
cas(Y, Z) = \underset{i\in\{1,\dots,S\}}{softmax}\Big(\frac{c - \sum_m^d \beta_m|Y_{0,i,m} - Z_{0,i,m}|}{s}\eta \Big),
\end{equation}
where $Y$ and $Z$ are broadcast to the same shape $(1\times S\times d)$, $|\cdot|$ is the element-wise absolute value, $\beta \in \mathbb{R}^{1\times d}$ is a constant vector of ones,\footnote{We found it sufficient to use constants instead of the learnable parameters in the original CAP, which has different $\beta$s for different heads.} $c=\sqrt{\frac{4}{\pi}}d$ and $s=\sqrt{(2-\frac{4}{\pi})d}$ are two constant scalars, broadcast as $(1\times S)$ to be compatible with Equation~\eqref{dba}, and $\eta$ is the ``down-scaling” hyperparameter, set to be a single value ($=0.1$ for all backbones) throughout this paper.\footnote{Another attention function proposed by~\citet{miv25}, ``variance-excited multiplicative attention'' (VEMA), is not applicable here, because channel-wise variance is undefined for 1-shot bags, which are often seen in the MD data. However, a simpler version of multiplicative attention---the original ``scaled dot-product attention'' (SDPA) popularized by the transformer \citep{transformer17, bert19, vit21}---is still applicable.
}

\subsection{Co-excitation, $MHCE(\cdot)$}\label{sec:coe}

Concretely, the ``multi-head co-excitation'' function, $MHCE(\cdot)$, is
\begin{equation}\label{sce}
MHCE(x) = sigmoid\Big(x\kappa\Big),
\end{equation}
where $\kappa\in\mathbb{R}^{C\times d}$ is a learnable parameter, and $sigmoid$ is the element-wise activation function. We emphasize that the same $x=v^{query}$, as well as the same learnable parameters, are shared between Equations~\eqref{perhead_p} and \eqref{perhead_q}, that is, in both Siamese twins, giving rise to the term ``co-excitation''. Also note that $MHCE(\cdot)$ is multi-headed, because $\kappa$ projects all the channels to a head. Equation~\eqref{sce} is a simplified version of MHCE in the original CAP~\citep{miv25}.

\subsection{``In-attention skip-connection''}\label{sec:inatt_skip}

The ``in-attention skip-connection'' mechanism corresponds to the summation terms in Equations~\eqref{perhead_p} and \eqref{perhead_q}, 
\begin{align}\label{full_skip}
    v^{target}W_j^V \odot MHCE_j(v^{query}) + v^{target}W_j^K,\\\nonumber
    v^{query}W_j^V \odot MHCE_j(v^{query}) + v^{query}W_j^K.
\end{align} 

Note that ``no skip-connection'' means these terms become the following ones, without summation:
\begin{align}\label{no_skip}
v^{target}W_j^V \odot MHCE_j(v^{query})\\ \nonumber
v^{query}W_j^V \odot MHCE_j(v^{query}).
\end{align}

Finally, if we disable the (multi-head) cross-attention, i.e., replace $cas(\cdot,\cdot)$ by ``average-pooling'' and exclude $W_j^K$ (as done in Table~\ref{tab:more_abl}), the ``in-attention skip-connection'' mechanism will be reduced to 
\begin{align}\label{simple_skip}
    v^{target}W_j^V \odot MHCE_j(v^{query}) + v^{target},\\\nonumber
    v^{query}W_j^V \odot MHCE_j(v^{query}) + v^{query}.
\end{align}

\section{Details of the experimental protocol (Section~\ref{sec:setup})} \label{app:exp}

\subsection{Implementation and evaluation}

We implemented all algorithms in our experiments using PyTorch~\citep{torch19}. In particular, we re-implemented TSA and eTT based on their GitHub repositories (\url{https://github.com/VICO-UoE/URL} and \url{https://github.com/chmxu/eTT_TMLR2022}). To verify that the results from our re-implementation approximately replicate the results in \citep{tsa22, ett22}, we compared them based on the original MD (nine datasets) and found them to be similar. More precisely, our re-implementation of TSA in Table~\ref{tab:rn18} is based on the same SDL-ResNet-18 backbone used in~\citep{tsa22}, and achieves an average MD accuracy of $69.9\%$ vs. $71.9\%$ as reported. The difference is likely due to the known (shuffle) issue of MD sampling that can cause inflation of accuracy. On the other hand, our re-implementation of eTT in Table~\ref{tab:miv_acc_ett} achieves $80.3\%$ (on average across MD), slightly better than the original results reported for eTT ($78.7\%$). The difference is possibly caused by how the backbone models were pretrained---the off-the-shelf backbone used in Table~\ref{tab:miv_acc_ett} was pretrained using the full ILSVRC-2012 dataset (with 1000 classes) whereas the backbone used in the original paper of eTT was pretrained on a subset of the ILSVRC dataset.   

The evaluation procedure we used is as follows, using the standard nomenclature of FSC. During inference, a batch of test queries known as a ``query set'', with the same classes as the support set, is paired with the support set. Such a pair, involving a query set and a support set, is usually called an ``episode'' or a ``task''---we use both terms (``episode'' and ``task'') interchangeably. A task (or episode) is evaluated when the corresponding query set is classified using any algorithm and evaluated against the ground truth.

The MIV-head evaluates each query independently, to prevent leaking information from other queries within the same task---this is because prototypes created by the MIV-head depend on both support set and query, and we need to ensure prototypes produced for one query \emph{cannot} be used by another. To this end, during evaluation of the MIV-head, we repeated the support set by $\mathcal{X}$ times, where $\mathcal{X}$ is the size of the query set.

We used a single Nvidia A100 GPU (40GB GPU-memory) with 16 vCPUs to produce all main experimental results. 

\subsection{Data} \label{app:data}
Meta-Dataset (MD)~\citep{md20} is a few-shot classification benchmark that initially consisted of ten datasets: ILSVRC-2012~\citep{imagenet_cvpr09, ILSVRC15}, Omniglot~\citep{omni15}, FGVC-Aircraft (Aircraft)~\citep{aircraft13}, CUB-200-2011 (Birds)~\citep{cub11}, Describable Textures (DTD)~\citep{dtd14}, QuickDraw (QDraw)~\citep{qdraw16}, FGVCx Fungi (Fungi)~\citep{fungi18}, VGG Flower~\citep{flower08}, Traffic Signs~\citep{sign11} and MSCOCO (COCO)~\citep{coco14}. It was then further expanded with MNIST~\citep{mnist98}, CIFAR-10 and CIFAR-100~\citep{cifar09}. For an even more comprehensive evaluation, we follow~\citep{fes24} to add Food101~\citep{food14} and four datasets from the CDFSL benchmark~\citep{mdmore20}—CropDisease, EuroSAT, ISIC and ChestX. Similarly to recent studies~\citep{closerlook23, vitbaseline24}, we excluded ILSVRC-2012 from the MD benchmark for our cross-domain evaluations because most of the off-the-shelf backbones were pretrained on the full ILSVRC-2012 dataset including the entire 1000 classes. Therefore, the test data in our experiments comprises 17 datasets in total, leading to the extended MD (MD+) benchmark.

We generated 600 tasks per test dataset, by sampling from the test-split of each MD dataset. However, due to the shuffling issue in MD (mentioned in~\citep{tsa22}), we retrieved only their sampling \emph{schema} for each task, i.e., the sizes and classes of support and query sets in each task, \emph{not} the exact samples. We then conducted our own random sampling from the MD data repository based on this schema to create test tasks for all experiments.

\subsection{The backbone models and links to their weights} \label{app:backbone}
For supervised backbones, we downloaded ResNet weights from PyTorch Hub libraries (or, if specially trained, from the GitHub repository of \citep{url21}) and used DeiT (no distillation, ~\cite{deit21}) as ViT weights. For self-supervised backbones for both ResNet and ViT, we employed weights from DINO~\citep{dino21}. The links to the backbone model weights are as follows:
\begin{itemize}
    \item DINO-ViT and DINO-ResNet50 backbones' pretrained model weights can be downloaded from the DINO repository: \url{https://github.com/facebookresearch/dino}
    \item DeiT(ViT) backbone's pretrained model weights can be downloaded from the DeiT repository: \url{https://github.com/facebookresearch/deit/blob/main/README_deit.md}
    \item SDL-ResNet18 backbone's pretrained model weights can be downloaded from the ``URL'' repository: \url{https://github.com/VICO-UoE/URL}
    \item ResNet backbones' pretrained model weights from PyTorch Hub (\url{https://pytorch.org/docs/stable/hub.html}) should be automatically downloaded when calling the relevant API
\end{itemize}

\subsection{Hyperparameters and optimization}\label{app:hyper_optim}

\begin{table}[h]
\centering
\caption{Optimization settings} \label{tab:opt_setting}
\begin{tabular}{@{}lcccc@{}} \toprule
 & MIV-head & TSA & eTT & ``FiT Head'' / Baseline++ \\ \cmidrule{2-5}
 & SGD, & Adadelta, & AdamW, & Adam, \\
 Optimizer & momentum=0.9, &  $\rho=0.9$, & $\epsilon=1\mathrm{e}{-4}$, & $\epsilon=1\mathrm{e}{-8}$, \\
 & no weight-decay & no weight-decay & weight-decay=0.01 & no weight-decay \\  \midrule
\multirow{3}{50pt}{Learning\\rate} & $LR$=0.3, & \multirow{3}{50pt}{$LR_\alpha=0.5$,\\[.2\baselineskip] $LR_\beta=1$} & $LR=1\mathrm{e}{-3},$ & \multirow{3}{75pt}{FiT Head: $0.0035$,\\[.2\baselineskip]Baseline++: $0.03$}\\
 & Component 2: $LR$,  & & feature adaptation: $LR$,& \\
 & Component 1: $0.05LR$ & & otherwise: $0.5LR$ & \\ \midrule
Iterations & 40 & 40 & 40 & 400 \\
\bottomrule 
\end{tabular}
\end{table}

\subsubsection{Optimization}\label{app:optim}

Table~\ref{tab:opt_setting} lists the optimizers, learning rate (LR) and number of iterations (or steps) for optimization adopted by all algorithms during adaptations, following their published settings.

\subsubsection{Hyperparameters of the MIV-head} \label{app:hyper}

The considerations leading to the choice of hyperparameter settings include computational constraints (e.g., GPU memory), backbone architecture constraints, values used by standard practices, and recommendations from the relevant literature. We emphasize that all hyperparameter values were fixed before experimentation on the extended MD benchmark. The values of all hyperparameters are listed as follows. 

\textbf{Component 1.} 
\begin{enumerate}
    \item ViT backbones
    \begin{enumerate}[label*=\arabic*.]
        \item $\tau = 200$ (in Equations~\eqref{ppool1} and~\eqref{ppool2})
        \item Following recommendations of the DINO paper~\citep{dino21}, we used the last 4 blocks of ViT backbones.
        \item Adaptive max-pooling shapes for patch 8 (``[CLS]'' is the ``prefix token'' embedding vector, \emph{cf.} Section~\ref{sec:arch}):
        \begin{description}
            \item[Last block:] ``[CLS]''
            \item[Second-to-last block:] ``[CLS]''
            \item[Third last block:] ($8\times8,\quad 12\times12,\quad$ ``[CLS]'')
            \item[Fourth last block:] ($16\times16,\quad 20\times20, \quad 24\times24,\quad$ ``[CLS]'')
        \end{description}
        \item Adaptive max-pooling shapes for patch 16:
        \begin{description}
            \item[Last block:] ``[CLS]''
            \item[Second-to-last block:] ``[CLS]''
            \item[Third last block:] ($7\times7,\quad$ ``[CLS]'')
            \item[Fourth last block:] ($10\times10,\quad 13\times13,\quad$ ``[CLS]'')
        \end{description}
    \end{enumerate}
    \item ResNet backbones:
    \begin{enumerate}[label*=\arabic*.]
        \item $\tau = 500$ (in Equations~\eqref{ppool1} and~\eqref{ppool2})
        \item We used the last 2 blocks except for (off-the-shelf) ResNet-18, in which case we used the last 3 blocks. 
        Note that SDL-ResNet-18's architecture is quite different from that of the off-the-shelf ResNet-18---the output shapes from its last two blocks are similar to the shapes from the second and third last blocks of the off-the-shelf ResNet-18. Therefore, we only used SDL-ResNet-18's last 2 blocks. 
        \item Adaptive max-pooling shapes for ResNet-50/34:
        \begin{description}
            \item[Last block:] $(4\times4,\quad 5\times5,\quad 6\times6,\quad 7\times7)$
            \item[Second-to-last block:] $(8\times8,\quad 9\times9, \quad 11\times11,\quad 13\times13, \quad 14\times14)$
        \end{description}
        \item Adaptive max-pooling shapes for off-the-shelf ResNet-18:
        \begin{description}
            \item[Last block:] $(3\times3)$
            \item[Second-to-last block:] $(4\times4,\quad 5\times5,\quad6\times6)$
            \item[Third last block:] $(7\times7,\quad8\times8,\quad9\times9,\quad10\times10,\quad11\times11)$
        \end{description}
        \item Adaptive max-pooling shapes for SDL-ResNet-18:
        \begin{description}
            \item[Last block:] $(3\times3,\quad4\times4,\quad5\times5,\quad6\times6)$
            \item[Second-to-last block:] $(7\times7,\quad8\times8,\quad9\times9,\quad10\times10,\quad11\times11)$
        \end{description}
    \end{enumerate}
\end{enumerate}

\textbf{Component 2.}
\begin{itemize}
    \item Number of heads = output channel-size (of the backbone's specific block) / 64
    \item Attention function: we used the ``distance-based attention'' (DBA) function specified by Equation~\eqref{dba} in Section~\ref{sec:cas}. 
    \item ``Down-scaling” hyperparameter (in Equation~\eqref{dba}
    ), $\eta=0.1$
\end{itemize}

In addition, to boost sample sizes for ``low-shot'' classes, we created distorted views of the support set using RandAugment and treated them as extra ``pseudo-queries'' during training. This treatment is similar to that in recent FSC studies~\citep{pmf22, vitbaseline24}. Unlike these other studies, which used custom data augmentation methods, we adopted a standard one. More precisely, the transforms of the support set include:
\begin{itemize}
    \item RandAugment~\citep{randaug20}
    \item Randomly convert image to grayscale with a probability of 0.2
    \item Horizontally flip image randomly with a probability of 0.5
\end{itemize}
The threshold to trigger the above data augmentation was set between 15 and 30 depending on the backbone and input resolution and was reduced where necessary to avoid OOM errors. When augmentation is triggered for a class, the number of augmented (or ``distorted'') views of the support set will be $max(1,\lfloor\frac{T}{S}\rfloor)$, where $T$ denotes the threshold and $S$ denotes the original ``number of shots'' in the class. Note that we treat $T=0$ as \emph{no} augmentation, see Table~\ref{tab:da_abl} in Appendix~\ref{app:abl_single}.

\subsubsection{TSA hyperparameters}
\begin{itemize}
    \item Adapter type: ``residual''
    \item Adapter form: ``matrix''
    \item Adaptation option: ``alpha+beta''
    \item Initialization for adapter: identity matrix scaled by 0.0001
\end{itemize}

\subsubsection{``FiT Head'' hyperparameters}
There is one hyperparameter: the method to estimate covariance, which leads to FiT-head variations employing either ``Linear Discriminant Analysis'' (LDA), ``Quadratic Discriminant Analysis'' (QDA) or ``ProtoNets'' (diagonal matrix as covariance matrix). We set this hyperparameter to be LDA, because QDA cannot be used (i.e., is undefined) for 1-shot support sets commonly seen in the MD benchmark. ``ProtoNets'' is shown by~\cite{fit23} to be inferior to LDA/QDA.

eTT and Baseline++ have no obvious hyperparameters apart from the ones for optimization (Table~\ref{tab:opt_setting}). 

\section{Additional experimental results for Section~\ref{sec:results}}

\subsection{More results on accuracy (Section~\ref{sec:results_acc})}\label{app:acc}

\begin{table}[hbp]
\centering
\begin{subtable}{\linewidth}
\centering
\begin{NiceTabular}{@{}lccccc@{}}
\toprule
\Block[c]{3-1}{Test\\dataset}& \multicolumn{5}{c}{Off-the-shelf backbones for $224\times224$ input resolution} \\
& \multicolumn{2}{c}{Supervised ResNet-50} & & \multicolumn{2}{c}{Self-supervised(DINO) ResNet-50}\\ 
\cline{2-3}\cline{5-6}	
&TSA&Ours&&TSA&Ours\\

Omniglot	&	59.3	$\pm$	1.1	&\textbf{	68.3	$\pm$	1.1	}&&	60.4	$\pm$	1.1	&\textbf{	67.5	$\pm$	1.1	}\\
Aircraft	&	40.6	$\pm$	0.8	&\textbf{	42.9	$\pm$	0.8	}&&	37.5	$\pm$	0.8	&\textbf{	37.7	$\pm$	0.8	}\\
Birds	&\textbf{	75.3	$\pm$	1.0	}&	68.6	$\pm$	1.0	&&\textbf{	54.4	$\pm$	1.0	}&	49.6	$\pm$	0.9	\\
Textures	&	61.4	$\pm$	0.8	&\textbf{	63.2	$\pm$	0.8	}&&\textbf{	60.0	$\pm$	0.8	}&	58.7	$\pm$	0.8	\\
Quick Draw 	&	58.7	$\pm$	1.0	&\textbf{	63.0	$\pm$	1.0	}&&\textbf{	59.9	$\pm$	0.9	}&	59.6	$\pm$	0.9	\\
Fungi	&	51.1	$\pm$	1.1	&\textbf{	51.8	$\pm$	1.0	}&&\textbf{	54.8	$\pm$	1.0	}&	49.3	$\pm$	1.0	\\
VGG Flower	&	72.1	$\pm$	0.9	&\textbf{	79.9	$\pm$	0.8	}&&\textbf{	82.3	$\pm$	0.7	}&	79.5	$\pm$	0.8	\\
Traffic Sign	&	53.1	$\pm$	0.9	&\textbf{	55.4	$\pm$	0.9	}&&\textbf{	56.2	$\pm$	0.9	}&	55.9	$\pm$	0.9	\\
MSCOCO	&\textbf{	55.0	$\pm$	1.0	}&	52.3	$\pm$	1.0	&&\textbf{	50.2	$\pm$	1.0	}&	46.8	$\pm$	0.9	\\ \midrule
Average (MD)	&	58.5			&\textbf{	60.6			}&&\textbf{	57.3			}&	56.1			\\ \midrule
MNIST	&	55.3	$\pm$	0.9	&\textbf{	63.1	$\pm$	0.9	}&&	55.5	$\pm$	0.9	&\textbf{	58.9	$\pm$	0.9	}\\
CIFAR-10	&\textbf{	63.2	$\pm$	0.8	}&	57.2	$\pm$	0.9	&&\textbf{	55.1	$\pm$	0.8	}&	51.8	$\pm$	0.8	\\
CIFAR-100	&\textbf{	72.5	$\pm$	0.9	}&	66.0	$\pm$	0.9	&&\textbf{	66.0	$\pm$	0.9	}&	61.5	$\pm$	0.9	\\
CropDisease	&	74.7	$\pm$	0.9	&\textbf{	83.5	$\pm$	0.8	}&&	83.7	$\pm$	0.8	&\textbf{	84.0	$\pm$	0.8	}\\
EuroSAT	&	66.8	$\pm$	0.9	&\textbf{	73.9	$\pm$	0.8	}&&	72.6	$\pm$	0.8	&\textbf{	73.1	$\pm$	0.8	}\\
ISIC	&	27.9	$\pm$	0.6	&\textbf{	29.8	$\pm$	0.6	}&&\textbf{	31.8	$\pm$	0.6	}&	31.2	$\pm$	0.6	\\
ChestX	&	22.4	$\pm$	0.5	&\textbf{	22.8	$\pm$	0.5	}&&	22.8	$\pm$	0.5	&	22.9	$\pm$	0.5	\\
Food101	&\textbf{	63.9	$\pm$	1.0	}&	59.1	$\pm$	0.9	&&\textbf{	56.8	$\pm$	0.9	}&	50.9	$\pm$	0.9	\\ \midrule
Average (MD+)	&	57.2			&\textbf{	58.9			}&&\textbf{	56.5			}&	55.2			\\

\bottomrule 
\end{NiceTabular} 
\caption{\normalsize ``5-way 1-shot'': comparison between TSA and the MIV-head (Ours).}
\end{subtable}

\begin{subtable}{\linewidth}
\begin{NiceTabular}{@{}lcccccc@{}}\toprule
\Block[c]{4-1}{Test\\dataset}& \multicolumn{6}{c}{Off-the-shelf pretrained backbones for $224\times224$ input resolution} \\
& \multicolumn{2}{c}{Supervised(DeiT) ViT-small} & & \multicolumn{3}{c}{Self-supervised(DINO) ViT-small}\\ 
\cline{2-3}\cline{5-7}
&eTT& Ours&&eTT& \multicolumn{2}{c}{Ours}\\
\cline{6-7}&\multicolumn{2}{c}{(Patch 16)}&&(Patch 16)& (Patch 8) &(Patch 16)\\

Omniglot	&	62.7	$\pm$	1.1	&\textbf{	69.5	$\pm$	1.0	}&&	71.5	$\pm$	1.1	&\textbf{	73.9	$\pm$	1.0	}&	70.9	$\pm$	1.1	\\
Aircraft	&\textbf{	41.7	$\pm$	0.8	}&	41.2	$\pm$	0.9	&&	40.2	$\pm$	0.9	&\textbf{	42.8	$\pm$	0.9	}&	39.0	$\pm$	0.8	\\
Birds	&\textbf{	79.5	$\pm$	0.9	}&	74.2	$\pm$	0.9	&&\textbf{	78.4	$\pm$	0.9	}&	74.3	$\pm$	0.9	&	72.1	$\pm$	0.9	\\
Textures	&	57.4	$\pm$	0.8	&\textbf{	59.7	$\pm$	0.8	}&&\textbf{	62.5	$\pm$	0.8	}&	61.0	$\pm$	0.7	&	61.7	$\pm$	0.8	\\
Quick Draw 	&	58.7	$\pm$	0.9	&\textbf{	62.6	$\pm$	0.9	}&&	63.6	$\pm$	0.9	&\textbf{	64.0	$\pm$	1.0	}&	62.9	$\pm$	1.0	\\
Fungi	&\textbf{	55.3	$\pm$	1.0	}&	54.8	$\pm$	1.0	&&\textbf{	59.5	$\pm$	1.0	}&	59.2	$\pm$	1.0	&	58.1	$\pm$	1.0	\\
VGG Flower	&	74.3	$\pm$	0.9	&\textbf{	77.6	$\pm$	0.8	}&&\textbf{	86.4	$\pm$	0.7	}&	84.9	$\pm$	0.7	&	85.1	$\pm$	0.7	\\
Traffic Sign	&	50.8	$\pm$	0.8	&\textbf{	55.0	$\pm$	0.8	}&&	57.4	$\pm$	1.0	&\textbf{	62.6	$\pm$	0.9	}&	57.9	$\pm$	0.9	\\
MSCOCO	&\textbf{	58.2	$\pm$	1.0	}&	57.6	$\pm$	1.0	&&\textbf{	56.3	$\pm$	0.9	}&	54.5	$\pm$	1.0	&	51.5	$\pm$	0.9	\\ \midrule
Average (MD)	&	59.9			&\textbf{	61.4			}&&	64.0			&\textbf{	64.1			}&	62.1			\\ \midrule
MNIST	&	60.8	$\pm$	0.9	&\textbf{	64.3	$\pm$	0.9	}&&	59.4	$\pm$	0.8	&\textbf{	65.2	$\pm$	0.9	}&	60.4	$\pm$	0.9	\\
CIFAR-10	&\textbf{	73.1	$\pm$	0.8	}&	69.0	$\pm$	0.8	&&\textbf{	68.6	$\pm$	0.8	}&	65.2	$\pm$	0.8	&	59.7	$\pm$	0.8	\\
CIFAR-100	&\textbf{	78.8	$\pm$	0.8	}&	74.6	$\pm$	0.9	&&\textbf{	76.7	$\pm$	0.8	}&	74.4	$\pm$	0.8	&	69.1	$\pm$	0.9	\\
CropDisease	&	75.0	$\pm$	0.9	&\textbf{	81.3	$\pm$	0.8	}&&	83.6	$\pm$	0.8	&\textbf{	84.5	$\pm$	0.8	}&	84.3	$\pm$	0.8	\\
EuroSAT	&	62.6	$\pm$	0.9	&\textbf{	69.5	$\pm$	0.9	}&&	73.8	$\pm$	0.8	&	75.0	$\pm$	0.8	&\textbf{	75.2	$\pm$	0.8	}\\
ISIC	&	31.1	$\pm$	0.6	&\textbf{	32.2	$\pm$	0.6	}&&\textbf{	34.0	$\pm$	0.7	}&	33.7	$\pm$	0.6	&	33.5	$\pm$	0.6	\\
ChestX	&	21.8	$\pm$	0.5	&\textbf{	22.5	$\pm$	0.5	}&&	22.9	$\pm$	0.5	&\textbf{	23.3	$\pm$	0.5	}&\textbf{	23.3	$\pm$	0.5	}\\
Food101	&\textbf{	64.9	$\pm$	1.0	}&	64.6	$\pm$	0.9	&&	63.1	$\pm$	0.9	&\textbf{	65.1	$\pm$	0.9	}&	60.7	$\pm$	0.9	\\ \midrule
Average (MD+)	&	59.2			&\textbf{	60.6			}&&	62.2			&\textbf{	62.6			}&	60.3			\\

\bottomrule 
\end{NiceTabular}
\caption{\normalsize ``5-way 1-shot'': comparison between eTT and the MIV-head (Ours).}
\end{subtable}
\caption{``Five-way One-shot'' setting: comparison of accuracy (\%) $\pm$95\% confidence interval between TSA/eTT and Ours based on all non-ILSVRC datasets in an extended MD benchmark, aligned by the same backbones. Higher accuracy is bolded within each comparison.} \label{tab:1shot}
\end{table}

\subsubsection{``Five-way One-shot'' setting}\label{app:acc1shot}

Table~\ref{tab:1shot} reports the details, including 95\% confidence intervals, of the summarized results under the ``Five-way One-shot'' setting in Table~\ref{tab:summ1shot} of Section~\ref{sec:results_acc}.

\subsubsection{Comparisons with ``FiT Head'' and Baseline++} \label{app:fh}

\begin{sidewaystable}[hbp]
\centering
\begin{NiceTabular}{@{\extracolsep{2pt}}l@{}c@{}c@{}c@{\hskip2pt}c@{}c@{}c@{\hskip2pt}c@{}c@{}c@{\hskip2pt}c@{}c@{}c@{}}
\toprule
\Block[c]{3-1}{Test\\dataset}& \multicolumn{12}{c}{Off-the-shelf pretrained backbone models for $224\times224$ input resolution} \\
& \multicolumn{3}{c}{ResNet-50} & \multicolumn{3}{c}{DINO ResNet-50} & \multicolumn{3}{c}{DeiT-small/16} & \multicolumn{3}{c}{DINO ViT-small/8}\\ 
\cmidrule{2-4}\cmidrule{5-7}\cmidrule{8-10}\cmidrule{11-13}	
&FiT-head&Baseline++&Ours&FiT-head&Baseline++&Ours&FiT-head&Baseline++&Ours&FiT-head&Baseline++&Ours\\
Omniglot	&	58.1$\pm$1.5	&	56.4$\pm$1.4	&\textbf{	76.5$\pm$1.2	}	&	60.3$\pm$1.5	&	60.8$\pm$1.4	&\textbf{	74.4$\pm$1.2	}	&	61.9$\pm$1.5	&	61.3$\pm$1.4	&\textbf{	74.5$\pm$1.2	}	&	72.0$\pm$1.3	&	68.1$\pm$1.3	&\textbf{	78.6$\pm$1.1	}\\ 
Aircraft	&	64.9$\pm$1.1	&	67.9$\pm$1.0	&\textbf{	84.4$\pm$1.0	}	&	81.1$\pm$1.1	&	74.8$\pm$1.1	&\textbf{	84.4$\pm$1.0	}	&	52.8$\pm$1.1	&	69.9$\pm$1.0	&\textbf{	79.4$\pm$1.0	}	&	83.6$\pm$1.1	&	82.0$\pm$1.0	&\textbf{	86.0$\pm$0.9	}\\ 
Birds	&	80.4$\pm$0.8	&	84.1$\pm$0.7	&\textbf{	87.2$\pm$0.8	}	&	72.5$\pm$1.1	&	71.2$\pm$1.1	&\textbf{	81.8$\pm$1.1	}	&	79.2$\pm$1.0	&	87.2$\pm$0.6	&\textbf{	88.7$\pm$0.7	}	&	91.0$\pm$0.6	&	92.4$\pm$0.6	&\textbf{	91.8$\pm$0.7	}\\ 
Textures	&	80.8$\pm$0.7	&	86.6$\pm$0.6	&\textbf{	89.0$\pm$0.6	}	&	86.6$\pm$0.7	&	88.7$\pm$0.6	&\textbf{	89.6$\pm$0.6	}	&	68.6$\pm$1.1	&	85.7$\pm$0.7	&\textbf{	88.3$\pm$0.6	}	&	82.0$\pm$0.9	&	89.3$\pm$0.6	&\textbf{	89.7$\pm$0.6	}\\ 
Quick Draw 	&	54.9$\pm$1.0	&	60.8$\pm$1.0	&\textbf{	71.7$\pm$0.8	}	&	57.2$\pm$1.2	&	63.4$\pm$1.0	&\textbf{	70.1$\pm$0.8	}	&	45.2$\pm$1.1	&	63.7$\pm$0.9	&\textbf{	70.9$\pm$0.8	}	&	58.1$\pm$1.0	&	68.0$\pm$0.9	&\textbf{	71.8$\pm$0.8	}\\ 
Fungi	&	38.9$\pm$1.2	&	46.4$\pm$1.1	&\textbf{	60.6$\pm$1.1	}	&	48.1$\pm$1.2	&	50.4$\pm$1.1	&\textbf{	60.5$\pm$1.1	}	&	34.0$\pm$1.3	&	49.8$\pm$1.1	&\textbf{	60.9$\pm$1.1	}	&	54.9$\pm$1.3	&	59.5$\pm$1.1	&\textbf{	65.6$\pm$1.0	}\\ 
VGG Flower	&	88.4$\pm$0.6	&	89.1$\pm$0.6	&\textbf{	96.0$\pm$0.4	}	&	94.9$\pm$0.6	&	94.7$\pm$0.5	&\textbf{	96.7$\pm$0.4	}	&	84.0$\pm$1.0	&	90.3$\pm$0.6	&\textbf{	94.7$\pm$0.5	}	&	95.8$\pm$0.5	&	96.3$\pm$0.4	&\textbf{	97.3$\pm$0.4	}\\ 
Traffic Sign	&	55.7$\pm$1.4	&	56.0$\pm$1.3	&\textbf{	78.6$\pm$1.0	}	&	63.4$\pm$1.4	&	60.7$\pm$1.3	&\textbf{	81.8$\pm$1.0	}	&	42.7$\pm$1.4	&	53.5$\pm$1.3	&\textbf{	68.2$\pm$1.2	}	&	65.3$\pm$1.4	&	66.5$\pm$1.2	&\textbf{	79.6$\pm$1.0	}\\ 
MSCOCO	&	50.0$\pm$1.1	&	53.8$\pm$1.1	&\textbf{	60.8$\pm$1.0	}	&	45.3$\pm$1.3	&	50.2$\pm$1.2	&\textbf{	59.3$\pm$1.0	}	&	42.5$\pm$1.0	&	58.8$\pm$1.0	&\textbf{	65.7$\pm$0.9	}	&	43.3$\pm$1.1	&	59.5$\pm$1.0	&\textbf{	63.0$\pm$0.9	}\\\midrule
Average (MD)	&	63.6			&	66.8			&\textbf{	78.3			}	&	67.7			&	68.3			&\textbf{	77.6			}	&	56.7			&	68.9			&\textbf{	76.8			}	&	71.8			&	75.7			&\textbf{	80.4			}\\\midrule
MNIST	&	86.5$\pm$0.8	&	85.3$\pm$0.8	&\textbf{	93.2$\pm$0.6	}	&	88.9$\pm$0.7	&	86.1$\pm$0.7	&\textbf{	92.1$\pm$0.7	}	&	78.7$\pm$0.8	&	86.6$\pm$0.7	&\textbf{	91.5$\pm$0.6	}	&	84.3$\pm$0.9	&	89.3$\pm$0.6	&\textbf{	92.4$\pm$0.5	}\\ 
CIFAR-10	&	82.4$\pm$0.6	&	83.8$\pm$0.7	&\textbf{	84.7$\pm$0.7	}	&	78.4$\pm$0.8	&	79.9$\pm$0.8	&\textbf{	80.8$\pm$0.8	}	&	78.7$\pm$0.9	&	89.0$\pm$0.5	&\textbf{	89.2$\pm$0.5	}	&	88.6$\pm$0.5	&	91.6$\pm$0.5	&\textbf{	90.2$\pm$0.6	}\\ 
CIFAR-100	&	71.9$\pm$0.9	&	71.6$\pm$0.9	&\textbf{	74.9$\pm$0.8	}	&	64.6$\pm$1.1	&	68.6$\pm$1.0	&\textbf{	72.0$\pm$0.9	}	&	69.5$\pm$1.0	&	77.5$\pm$0.7	&\textbf{	80.1$\pm$0.7	}	&	78.7$\pm$0.8	&	82.3$\pm$0.7	&\textbf{	80.6$\pm$0.8	}\\ 
CropDisease	&	74.7$\pm$1.0	&	80.0$\pm$0.8	&\textbf{	91.2$\pm$0.5	}	&	84.7$\pm$0.8	&	85.9$\pm$0.7	&\textbf{92.3$\pm$0.5	}	&	73.5$\pm$1.2	&	81.2$\pm$0.8	&\textbf{	90.7$\pm$0.6	}	&	87.0$\pm$0.8	&	88.0$\pm$0.6	&\textbf{	92.7$\pm$0.5	}\\ 
EuroSAT	&	87.6$\pm$0.6	&	88.7$\pm$0.6	&\textbf{	93.2$\pm$0.4	}	&	90.8$\pm$0.6	&	91.7$\pm$0.5	&\textbf{	93.8$\pm$0.4	}	&	70.4$\pm$1.1	&	86.9$\pm$0.6	&\textbf{	91.0$\pm$0.5	}	&	89.0$\pm$0.6	&	92.5$\pm$0.5	&\textbf{	93.8$\pm$0.4	}\\ 
ISIC	&	32.2$\pm$0.7	&	30.1$\pm$0.7	&\textbf{	43.1$\pm$0.9	}	&	35.3$\pm$0.8	&	32.6$\pm$0.8	&\textbf{	44.3$\pm$0.9	}	&	30.6$\pm$0.6	&	32.7$\pm$0.8	&\textbf{	43.7$\pm$0.9	}	&	35.4$\pm$0.8	&	36.9$\pm$0.9	&\textbf{	46.1$\pm$0.9	}\\ 
ChestX	&	21.9$\pm$0.5	&	22.8$\pm$0.5	&\textbf{	25.9$\pm$0.5	}	&	24.0$\pm$0.5	&	24.5$\pm$0.5	&\textbf{	27.2$\pm$0.6	}	&	20.2$\pm$0.5	&	22.7$\pm$0.5	&\textbf{	25.0$\pm$0.5	}	&	22.6$\pm$0.5	&	26.0$\pm$0.5	&\textbf{	27.7$\pm$0.6	}\\ 
Food101	&	55.0$\pm$1.1	&	64.4$\pm$1.0	&\textbf{	69.5$\pm$1.0	}	&	59.0$\pm$1.1	&	62.7$\pm$1.1	&\textbf{	66.6$\pm$1.0	}	&	48.0$\pm$1.2	&	66.3$\pm$1.0	&\textbf{	70.6$\pm$0.9	}	&	68.8$\pm$1.0	&	75.6$\pm$0.8	&\textbf{	75.5$\pm$0.9	}\\\midrule
Average (MD+)	&	63.8			&	66.3			&\textbf{	75.3			}	&	66.8			&	67.5			&\textbf{	74.6			}	&	57.7			&	68.4			&\textbf{	74.9			}	&	70.6			&	74.3			&\textbf{	77.8			}\\ 
\bottomrule 
\end{NiceTabular}
\caption{Comparison of accuracy (in \%) $\pm$95\% confidence interval between ``FiT Head''(LDA), Baseline++ and the MIV-head (Ours) based on all non-ILSVRC datasets in an extended MD benchmark, and aligned with the same backbones. The row labeled ``Average (MD)'' indicates the average accuracy across nine original (non-ILSVRC) MD datasets whereas the row labeled ``Average (MD+)'' is the average across the total 17 extended MD datasets. We also conducted a two-sided paired t-test between FiT-head/Baseline++ and Ours, and bolded the higher accuracy when the p-value of a paired t-test is $<0.01$.} \label{tab:miv_fh}
\end{sidewaystable}

We also compared the MIV-head with two of the best-performing classification heads, namely Baseline++~\citep{closerfewshot19} and ``FiT Head'' (FiT-head) proposed as part of the ``FiT'' method~\citep{fit23}. The FiT-head is a Gaussian Naive Bayes classifier, with two variants, LDA or QDA, depending on how the covariance matrix is estimated. Although the two classifiers are usually used in conjunction with other methods like meta-training or fine-tuning methods, to compare with our approach, we treated them as standalone adaptation methods, on top of black-box backbones. In this sense, they can be seen as proxies for up-to-date ``head'' approaches.

The results of the comparison, between the FiT-head (LDA variant), Baseline++, and our approach aligned to the same backbones and using the same test datasets as in Tables~\ref{tab:miv_acc_tsa}, \ref{tab:miv_acc_ett} (Section~\ref{sec:results_acc}), are presented in Table~\ref{tab:miv_fh}. As demonstrated, although the accuracy of both methods appears better than that of the plain NCC classifier (\emph{cf.} Tables~\ref{tab:miv_acc_tsa}, \ref{tab:miv_acc_ett}), neither is competitive with the MIV-head.

\begin{table}[t]
\centering
\begin{tabular}{@{}lc@{\hskip6pt}c@{\hskip6pt}c@{\hskip6pt}c@{\hskip6pt}c@{\hskip6pt}c@{\hskip6pt}c@{}}
Algorithm & TSA & eTT & CTX & ALFA & ProtoNet & BOHB & Ours \\
Backbone & ResNet-34 & ViT-small/16 & ResNet-34 & 4CONV & ResNet-34 & ResNet-18 & ViT-small/8 \\
Test dataset & & & & & & & \\\toprule 
Omniglot	&	82.6$\pm$1.1	&	78.1$\pm$1.2	&	82.2$\pm$1.0	&	61.9$\pm$1.5	&	68.5$\pm$1.3	&	67.6$\pm$1.2	&	78.6$\pm$1.1	\\
Aircraft	&	80.1$\pm$1.0	&	79.9$\pm$1.1	&	79.5$\pm$0.9	&	63.4$\pm$1.1	&	58.0$\pm$1.0	&	54.1$\pm$0.9	&	86.0$\pm$0.9	\\
CUB	&	83.4$\pm$0.8 &	85.9$\pm$0.9	&	80.6$\pm$0.9	&	69.8$\pm$1.1	&	74.1$\pm$0.9	&	70.7$\pm$0.9	&	91.8$\pm$0.7	\\
DTD	&	79.6$\pm$0.7	&	87.6$\pm$0.6	&	75.6$\pm$0.6	&	70.8$\pm$0.9	&	68.8$\pm$0.8	&	68.3$\pm$0.8	&	89.7$\pm$0.6	\\
QDraw 	&	71.0$\pm$0.8	&	71.3$\pm$0.9	&	72.7$\pm$0.8	&	59.2$\pm$1.2	&	53.3$\pm$1.1	&	50.3$\pm$1.0	&	71.8$\pm$0.8	\\
Fungi	&	51.4$\pm$1.2	&	61.8$\pm$1.1	&	51.6$\pm$1.1	&	41.5$\pm$1.2	&	40.7$\pm$1.2	&	41.4$\pm$1.1	&	65.6$\pm$1.0	\\
VGGFlower	&	94.1$\pm$0.5	&	96.6$\pm$0.5	&	95.3$\pm$0.4	&	86.0$\pm$0.8	&	87.0$\pm$0.7	&	87.3$\pm$0.6	&	97.3$\pm$0.4	\\
TrafficSign	&	81.7$\pm$1.0	&	85.1$\pm$0.9	&	82.7$\pm$0.8	&	60.8$\pm$1.3	&	58.1$\pm$1.1	&	51.8$\pm$1.0	&	79.6$\pm$1.0	\\
COCO	&	61.7$\pm$1.0	&	62.3$\pm$1.0	&	59.9$\pm$1.0	&	48.1$\pm$1.1	&	41.7$\pm$1.1	&	48.0$\pm$1.0	&	63.0$\pm$0.9	\\\midrule
Average (MD)	&	76.2			&	78.7			&	75.6			&	62.4			&	61.1			&	60.0			&	80.4			\\ \bottomrule
\end{tabular}
\caption{More results reported from the CDFSL literature, based on non-ILSVRC datasets in Meta-Dataset (MD). Accuracies(in \%) are reported, along with $\pm$95\% confidence interval. All methods together with their backbones are included in the first and second rows. Our approach's results on an off-the-shelf self-supervised (DINO) ViT-small/8 backbone are in the rightmost column. Particularly, in addition to TSA \citep{tsa22} and eTT \citep{ett22}, ``CTX'' and ``ProtoNet'' are from ~\citep{ctx20}, ``ALFA'' is the ``ALFA+fo-Proto-MAML'' model in~\citep{alfa20}, ``BOHB'' is from~\citep{bohb20}, ``/16'' and ``/8'' denote patch-sizes of 16 and 8, respectively, of the input images taken by the corresponding backbones. All approaches except ours use their specially-trained backbones.} \label{tab:moreacc}
\end{table}

\subsubsection{Comparisons with previous literature}\label{app:acc_more}

We tabulate more results reported from the CDFSL literature in Table~\ref{tab:moreacc}, primarily sourced from the leaderboard published at the MD website: \url{https://github.com/google-research/meta-dataset}. All results are based on non-ILSVRC datasets in MD, and all methods together with their backbones are included in the first and second rows of Table~\ref{tab:moreacc}, where we also included our approach's results on a self-supervised (DINO) ViT-small backbone in the rightmost column. All approaches except ours use their specially-trained backbones, whereas ours utilizes an off-the-shelf backbone.

Strictly speaking, the comparison in Table~\ref{tab:moreacc} is unfair due to the difference in backbones used by the algorithms, unlike the backbone-aligned comparisons in Section~\ref{sec:results_acc}. Nonetheless, Table~\ref{tab:moreacc} does provide a broader perspective, showcasing what the MIV-head is able to achieve with off-the-shelf, black-box backbones.

\subsection{More results on adaptation cost (Section~\ref{sec:results_cost})}\label{app:cost}

\begin{figure}[t]
\centering
\includegraphics[width=\textwidth]{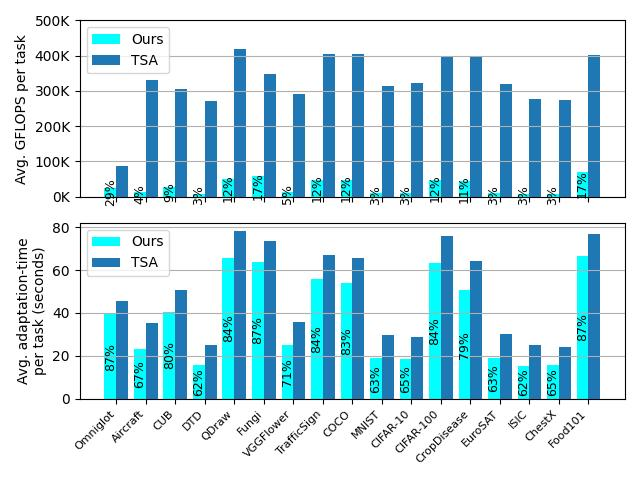}
\caption{Comparison of adaptation cost between TSA and the MIV-head (Ours), based on non-ILSVRC datasets in the extended Meta-Dataset and the same self-supervised (DINO) backbone. The adaptation cost is measured in GFLOPs (upper panel) and average training time per task, in seconds, using the same hardware (lower panel). Ours' adaptation cost is plotted by light-colored bars in which a percentage relative to the baselines' cost is also shown. Ours' GFLOPs is typically below $20\%$, and training time is $60\%$--$80\%$, of TSA's.}
\label{fig:tsa_cost_ssp}
\end{figure}

\begin{figure}[t]
\centering
\includegraphics[width=\textwidth]{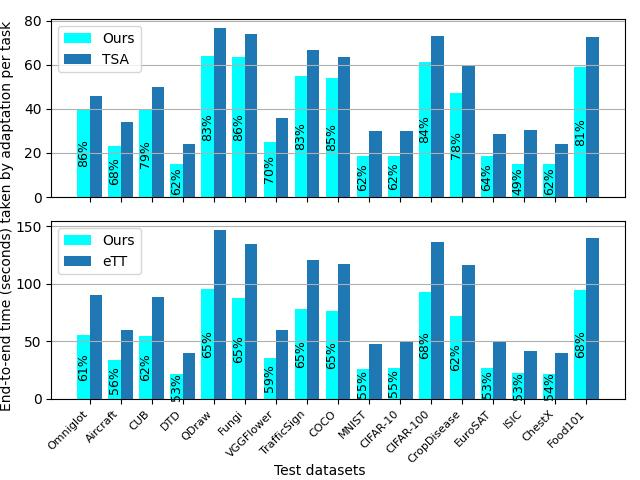}
\caption{Comparison of adaptation cost between TSA (upper panel), eTT (lower panel), and our MIV-head approach (``Ours''), based on non-ILSVRC datasets in the extended Meta-Dataset and based on the same supervised backbones. The adaptation cost is measured by average training time per task, in seconds, using the same hardware. Our method's adaptation cost is plotted by light-colored bars in which a percentage relative to the baselines' cost is also shown---our method's cost is typically $50\%$--$70\%$ and $60\%$--$80\%$, respectively, of that of eTT and TSA.}
\label{fig:time_spv}
\end{figure}

We first describe the protocols to calculate the adaptation cost metrics for all algorithms. We calculate ``GFLOPs'' using the ``fvcore'' library maintained by a computer vision team in FAIR, see \url{https://github.com/facebookresearch/fvcore/tree/main}. More precisely, fvcore counts the forward-pass GFLOPs; To estimate the GFLOPs of a single training step, we used the common rule of thumb that the backward pass, if needed, requires twice the GFLOPs of the forward pass. We then multiply the single-step GFLOPs by the number of iterations to calculate the GFLOPs of the entire training procedure. For ``training time'', we started timing all algorithms just \emph{before} feeding any data into a backbone and stopped measuring time as soon as training iterations were completed.

Figure~\ref{fig:tsa_cost_ssp} plots the two metrics of adaptation cost for TSA and the MIV-head, based on self-supervised backbones. It exhibits a similar pattern as Figure~\ref{fig:ett_cost_ssp}: the MIV-head incurs substantially lower adaptation cost than TSA, typically below $20\%$ in GFLOPs and $60\%$--$80\%$ in training time.

Figure~\ref{fig:time_spv} displays the adaptation time of all algorithms based on supervised backbones. Because simply changing the pre-training protocol (from self-supervised to supervised) of backbone weights, without architecture changes, would not impact adaptation GFLOPs, we omit the GFLOPs results here. Figure~\ref{fig:time_spv} shows similarly that, compared to the baselines, our approach incurs substantially lower adaptation cost, typically $50\%$--$70\%$ and $60\%$--$80\%$, respectively, of eTT's and TSA's.

\section{Additional experimental results and details for Section~\ref{sec:extra_results}}

\subsection{Details of results in Section~\ref{sec:more_bb}} \label{app:more_bb}

\begin{table}[t]
\begin{subtable}{\linewidth}
\centering
    \begin{NiceTabular}{@{}l@{}c@{}cc@{\hskip2pt}c@{\hskip2pt}c@{\hskip2pt}c@{\hskip2pt}c@{\hskip2pt}c@{\hskip2pt}c@{\hskip2pt}c@{\hskip2pt}c@{\hskip2pt}c@{\hskip2pt}}
         Backbone &  Pretraining &  Pretraining &  \Block[c]{2-1}{Omni}&  \Block[c]{2-1}{Acraft}&  \Block[c]{2-1}{CUB}&  \Block[c]{2-1}{DTD} &  \Block[c]{2-1}{QDraw} &  \Block[c]{2-1}{Fungi} &  \Block[c]{2-1}{Flower} & \Block[c]{2-1}{Sign} & \Block[c]{2-1}{COCO} & \Block[c]{2-1}{Avg} \\
         architecture & data & algorithm &&&&&&&&&& \\ \toprule
 DenseNet-161 & ILSVRC & supervised & 74.3&	82.5&	84.2&	87.8&	71.2&	58.8&	94.9&	79.9&	60.9&	77.2\\
 RegNetY-1.6GF & ILSVRC & supervised & 71.0&	81.5&	87.9&	89.3&	69.8&	60.9&	95.4&	75.6&	65.1&	77.4\\
 ViT-B/16 & ILSVRC & DeiT & 76.3&	81.5&	89.1&	88.9&	71.9&	62.0&	95.5&	70.5&	67.4&	78.1\\
  Swin Transformer & ILSVRC & SimMIM & 70.8& 80.0&	91.4&	88.8&	70.9&	64.3&	96.1&	74.3&	67.5&	78.2\\
 ViT-B/16 & ILSVRC & DINO & 75.0&	84.5&	89.8&	89.2&	71.3&	64.2&	97.2&	75.7&	63.1&	78.9\\
 ViT-L/14 & WebImageText & CLIP& 83.1	&85.0	&94.6	&91.7	&76.2&	66.3&	99.0&	83.3&	73.8&	83.7\\ \bottomrule
    \end{NiceTabular}
    \caption{MIV-head with a diverse range of backbones, based on the (non-ILSVRC) MD benchmark.}  
    \label{tab:miv_morebb}
\end{subtable}
\begin{subtable}{\linewidth}
\centering
    \begin{NiceTabular}{@{}l@{}c@{}cc@{\hskip2pt}c@{\hskip2pt}c@{\hskip2pt}c@{\hskip2pt}c@{\hskip2pt}c@{\hskip2pt}c@{\hskip2pt}c@{\hskip2pt}c@{\hskip2pt}c@{\hskip2pt}}
    Backbone &  Pretraining &  Pretraining &  \Block[c]{2-1}{Omni}&  \Block[c]{2-1}{Acraft}&  \Block[c]{2-1}{CUB}&  \Block[c]{2-1}{DTD} &  \Block[c]{2-1}{QDraw} &  \Block[c]{2-1}{Fungi} &  \Block[c]{2-1}{Flower} & \Block[c]{2-1}{Sign} & \Block[c]{2-1}{COCO} & \Block[c]{2-1}{Avg} \\
    architecture & data & algorithm &&&&&&&&&& \\ \toprule

    DenseNet-161	&	ILSVRC	&	supervised	&	59.7	&	76.3	&	84.7	&	85.8	&	62.0	&	47.7	&	91.6	&	59.0	&	52.2	&	68.8	\\ 
    RegNetY-1.6GF	&	ILSVRC	&	supervised	&	51.1	&	59.2	&	84.7	&	85.1	&	58.8	&	43.7	&	87.7	&	52.3	&	56.2	&	64.3	\\ 
    ViT-B/16	&	ILSVRC	&	DeiT	&	66.6	&	72.5	&	88.2	&	87.4	&	66.2	&	51.0	&	93.1	&	55.5	&	58.8	&	71.0	\\ 
    Swin Transformer	&	ILSVRC	&	SimMIM	&	57.2	&	63.3	&	86.5	&	87.0	&	63.8	&	49.9	&	91.5	&	54.7	&	61.4	&	68.4	\\ 
    ViT-B/16	&	ILSVRC	&	DINO	&	65.3	&	78.9	&	88.7	&	89.0	&	66.7	&	56.4	&	96.4	&	64.7	&	57.9	&	73.8	\\ 
    ViT-L/14	&	WebImageText	&	CLIP	&	74.9	&	79.8	&	95.8	&	90.1	&	68.8	&	54.7	&	98.5	&	73.4	&	61.3	&	77.5	\\ \bottomrule
    
    \end{NiceTabular}
    \caption{Baseline++ with the same backbones as in~\ref{tab:miv_morebb}, based on the same (non-ILSVRC) MD benchmark.}  
    \label{tab:blpp_morebb}
\end{subtable}
\caption{Accuracy (in \%) of the MIV-head (Table~\ref{tab:miv_morebb}) and Baseline++ (Table~\ref{tab:blpp_morebb}), with a diverse range of backbones, based on the (non-ILSVRC) MD benchmark. For ViT backbones, we follow the standard naming convention to denote them as ``ViT-[size]/[patch]'' where [size] is either ``B''(Base) or ``L''(Large) representing model sizes, and [patch] $\in \{8,14,16\}$ representing patch sizes.} \label{tab:morebb}
\end{table}

Detailed results for Table~\ref{tab:miv_diversebb} in Section~\ref{sec:more_bb}, of the MIV-head and Baseline++ respectively, are provided in Table~\ref{tab:morebb}. In particular, for the MIV-head, we retrieved the last 2 blocks for all backbones used in Table~\ref{tab:miv_morebb}, with key hyperparameters\footnote{If any hyperparameter value is unspecified here, it is the same as that in Appendix~\ref{app:hyper}.} and links to backbone weights listed as follows:

\begin{enumerate}
    \item Backbone: DenseNet-161
    \begin{enumerate}[label*=\arabic*.]
        \item Model weights: \url{https://huggingface.co/timm/densenet161.tv_in1k/tree/main}
        \item $\tau = 500$
        \item Adaptive max-pooling shapes:
        \begin{description}
            \item[Last block:] $(5\times5,\quad 7\times7)$
            \item[Second-to-last block:] $(9\times9, \quad 11\times11,\quad 14\times14)$
        \end{description}
    \end{enumerate}

    \item Backbone: RegNetY-1.6GF
    \begin{enumerate}[label*=\arabic*.]
        \item Model weights: \url{https://huggingface.co/timm/regnety_016.tv2_in1k/tree/main}
        \item $\tau = 500$
        \item Adaptive max-pooling shapes:
        \begin{description}
            \item[Last block:] $(4\times4,\quad 5\times5,\quad 6\times6,\quad 7\times7)$
            \item[Second-to-last block:] $(8\times8,\quad 9\times9, \quad 11\times11,\quad 13\times13, \quad 14\times14)$
        \end{description}
    \end{enumerate}
    
    \item Backbone: ViT-B/16, pretrained by DeiT and DINO
    \begin{enumerate}[label*=\arabic*.]
        \item Model weights:\\
        DeiT: \url{https://github.com/facebookresearch/deit/blob/main/README_deit.md}\\
        DINO: \url{https://github.com/facebookresearch/dino}
        \item $\tau = 200$
        \item Adaptive max-pooling shapes (``[CLS]'' is the ``prefix token'' embedding vector):
        \begin{description}
            \item[Last block:] (``[CLS]'')
            \item[Second-to-last block:] ($7\times7,\quad10\times10,\quad 14\times14,\quad$ ``[CLS]'')
        \end{description}
    \end{enumerate}

    \item Backbone: Swin Transformer-Base, pretrained by SimMIM
    \begin{enumerate}[label*=\arabic*.]
        \item Model weights: \url{https://drive.google.com/file/d/1xEKyfMTsdh6TfnYhk5vbw0Yz7a-viZ0w/view}
        \item $\tau = 200$
        \item Adaptive max-pooling shapes:
        \begin{description}
            \item[Last block:] ($4\times4$)
            \item[Second-to-last block:] ($7\times7$)
        \end{description}
    \end{enumerate}

    \item Backbone: ViT-L/14, pretrained by CLIP (vision model)
    \begin{enumerate}[label*=\arabic*.]
        \item Model weights: \url{https://huggingface.co/openai/clip-vit-large-patch14/tree/main}
        \item $\tau = 200$
        \item Adaptive max-pooling shapes (``[CLS]'' is the ``prefix token'' embedding vector):
        \begin{description}
            \item[Last block:] ($7\times7,\quad$ ``[CLS]'')
            \item[Second-to-last block:] ($11\times11,\quad 15\times15,\quad$ ``[CLS]'')
        \end{description}
    \end{enumerate}
\end{enumerate}

\section{Further ablation analysis and details for Section~\ref{sec:ablation}} \label{app:abl}

\subsection{Details of Figure~\ref{fig:abl_blk}} \label{app:abl_blk}

\begin{table}[t]
\centering
\begin{NiceTabular}{@{}lcccccccc@{}}
\toprule
\Block[c]{2-1}{Test\\dataset} & \multicolumn{2}{c}{$N=1$} && \multicolumn{2}{c}{$N=2$} && \multicolumn{2}{c}{$N=3$} \\
& GFLOPs & Accuracy && GFLOPs & Accuracy && GFLOPs & Accuracy \\ \cline{2-3} \cline{5-6} \cline{8-9}

Omniglot	&	20440.9	&	65.4	&&	25110.5	&	76.5	&&	30683.2	&	75.4	\\
Aircraft	&	10665.3	&	75.8	&&	12693.0	&	84.4	&&	26093.2	&	84.6	\\
CUB	&	21958.6	&	88.1	&&	26716.7	&	87.2	&&	43511.6	&	82.8	\\
DTD	&	6152.9	&	87.2	&&	7174.0	&	89.0	&&	16695.1	&	88.5	\\
QDraw 	&	40831.6	&	65.3	&&	50101.2	&	71.7	&&	55908.8	&	70.0	\\
Fungi	&	47591.1	&	56.9	&&	58619.4	&	60.6	&&	51605.2	&	56.5	\\
VGGFlower	&	11452.8	&	93.3	&&	13708.0	&	96.0	&&	26268.4	&	95.1	\\
TrafficSign	&	39299.6	&	69.1	&&	48106.7	&	78.6	&&	63215.3	&	76.4	\\
COCO	&	39099.9	&	60.5	&&	47860.8	&	60.8	&&	61364.1	&	57.7	\\\midrule
Average (MD)	&	26388.1	&	73.5	&&	32232.3	&	78.3	&&	41705.0	&	76.3	\\

\bottomrule 
\end{NiceTabular}
\caption{GFLOPs and accuracy(in \%), broken down by MD datasets, used to plot Figure~\ref{fig:abl_blk} in Section~\ref{sec:abl_alone}.} \label{tab:abl_blk}
\end{table}

Table~\ref{tab:abl_blk} shows the details by datasets, including both GFLOPs and accuracy, used to plot Figure~\ref{fig:abl_blk} in Section~\ref{sec:abl_alone}.

\subsection{More fine-grained results on individual mechanisms' marginal effect}\label{app:abl_single}

\begin{table}[t]
\centering
\begin{NiceTabular}{@{}lcccc@{}} 
& 1 &2 &3 &4 \\\toprule
\Block[c]{3-1}{Test\\dataset}& Full & Excluding  & Excluding  & Excluding\\
& CAP & cross-attention & ``in-attention skip-& co-excitation\\
& module & mechanism & connection''(IASC) & (and IASC)\\ \midrule

Omniglot	&	78.6	&	77.1	&	77.8	&	77.9	\\
Aircraft	&	86.0	&	84.6	&	86.1	&	86.0	\\
Birds	&	91.8	&	91.6	&	91.8	&	91.9	\\
Textures	&	89.7	&	89.0	&	89.7	&	89.6	\\
Quick Draw 	&	71.8	&	70.8	&	71.7	&	71.7	\\
Fungi	&	65.6	&	65.9	&	65.7	&	65.9	\\
VGG Flower	&	97.3	&	97.1	&	97.4	&	97.4	\\
Traffic Sign	&	79.6	&	78.5	&	78.5	&	79.5	\\
MSCOCO	&	63.0	&	65.3	&	62.9	&	63.0	\\\midrule
Average (MD)	&	80.4	&	80.0	&	80.2	&	80.3	\\

\bottomrule 
\end{NiceTabular}
\caption{Ablation study of Component 2 (CAP) of the MIV-head based on self-supervised (DINO) ViT-small/8 backbone, by removing some of the three mechanisms within CAP (columns 2--4) from this component (column 1). See detailed description of each column in Appendix~\ref{app:abl_single}} \label{tab:more_abl}
\end{table}

\begin{table}[t]
\centering
\begin{tabular}{@{}lccccc@{}} 
\multicolumn{6}{c}{Utilizing augmented support samples?} \\\toprule
Test & \multicolumn{2}{c}{Self-supervised(DINO)} && 
\multicolumn{2}{c}{Supervised} \\
dataset & \multicolumn{2}{c}{ViT-small/8 backbone} && \multicolumn{2}{c}{ResNet-50 backbone} \\
& Yes  & No  && Yes & No \\ \toprule

Omniglot	&	78.6	&	75.8	&&	76.5	&	73.2	\\
Aircraft	&	86.0	&	85.5	&&	84.4	&	83.6	\\
Birds	&	91.8	&	92.0	&&	87.2	&	86.7	\\
Textures	&	89.7	&	89.5	&&	89.0	&	88.8	\\
Quick Draw 	&	71.8	&	71.4	&&	71.7	&	70.5	\\
Fungi	&	65.6	&	63.3	&&	60.6	&	56.4	\\
VGG Flower	&	97.3	&	97.3	&&	96.0	&	95.3	\\
Traffic Sign	&	79.6	&	76.2	&&	78.6	&	74.8	\\
MSCOCO	&	63.0	&	60.5	&&	60.8	&	57.1	\\\midrule
Average (MD)	&	80.4	&	79.1	&&	78.3	&	76.3	\\

\bottomrule 
\end{tabular}
\caption{Study of the impact of data augmentation on the performance of our approach based on supervised ResNet-50 and self-supervised (DINO) ViT-small/8 backbones, by comparing performance of the MIV-head with and without data augmentation.}\label{tab:da_abl}
\end{table}

We conducted more fine-grained ablation analysis on Component 2 (CAP) of the MIV-head, based on the original non-ILSVRC MD (nine datasets) and a self-supervised (DINO) ViT backbone, similar to Table~\ref{tab:abl} (Section~\ref{sec:abl_alone}), that is, we remove one mechanism at a time from the ``full CAP module''---noting that removal of ``co-excitation'' will also remove ``in-attention skip-connection'', see \emph{Column 4} description below---to demonstrate their marginal effect. The results are shown in Table~\ref{tab:more_abl}---Column 1 shows the results with the full CAP module, and the other columns list the results by excluding mechanisms from CAP. Those exclusions are described as follows:

\begin{description}
    \item[Column 2:] We excluded the ``cross-attention'' mechanism (described by Appendix~\ref{sec:cas}), replacing it by the standard average-pooling (without any attention score). We still kept the co-excitation (described by Appendix~\ref{sec:coe}) and ``in-attention skip-connection'' mechanisms (specified by Equation~\eqref{simple_skip} in Appendix~\ref{sec:inatt_skip}) in this case;
    \item[Column 3:] We removed the ``in-attention skip-connection'' mechanism, while maintaining the cross-attention and co-excitation mechanisms (see Equation~\eqref{no_skip} in Appendix~\ref{sec:inatt_skip});
    \item[Column 4:] We removed the ``co-excitation'' mechanism (and ``in-attention skip-connection'' that, by design, co-exists alongside ``co-excitation''), while keeping the cross-attention mechanisms. Under this scenario, Equation~\eqref{full_skip} becomes $v^{target}W_j^V,\ v^{query}W_j^V$.
\end{description} 

As shown in Table~\ref{tab:more_abl}, removing each mechanism generally reduces accuracy, on average and across the nine datasets, although the effect is small. Hence, we include all of them in Component 2 (CAP). 

In addition, we evaluate the effect of augmentation using distorted support samples during training (\emph{cf.} Appendix~\ref{app:hyper}). The impact of distorted views of the support set (i.e., ``augmented support samples''), based on two backbones, supervised ResNet-50 and self-supervised (DINO) ViT-small/8, is shown in Table~\ref{tab:da_abl}. It indicates that the accuracy with support-set augmentation is higher than that without (\emph{cf.} columns ``Yes'' vs. ``No'' under the corresponding backbones), albeit in different magnitudes depending on the backbones. This demonstrates that ``low-shot'' tasks can benefit from more training data by using augmented views of the support set.

Notably, even without the support-set augmentation, our approach (\emph{cf.} the rightmost column in Table~\ref{tab:da_abl}) is still better than TSA, across most MD datasets and on average (76.3\% vs. 74.6\%), demonstrating the robustness of our approach. Moreover, based on the hardware used in our experiments, we found the computational cost incurred by such augmentation may be too high for the baselines, leading to either OOM or excessively long training time. This is an important difference from lightweight approaches like ours.

\subsection{Off-the-shelf vs. specially-trained ResNet-18 backbone} \label{app:abl_rn18}

\begin{table}[t]
\centering
\begin{NiceTabular}{lccccc} \toprule
\Block[c]{2-1}{Test\\dataset}& \multicolumn{5}{c}{ResNet-18 backbones for $84\times84$ input resolution} \\
& \multicolumn{2}{c}{SDL-ResNet-18} && \multicolumn{2}{c}{Off-the-shelf ResNet-18}\\ 
\cline{2-3}\cline{5-6}
&TSA&Ours&&TSA&Ours\\
Omniglot	&	75.8	$\pm$	1.2	&\textbf{	79.1	$\pm$	1.1	}	&&	72.8	$\pm$	1.3	&\textbf{	78.5	$\pm$	1.1	}\\ 
Aircraft	&\textbf{	71.9	$\pm$	1.1	}&	71.5	$\pm$	1.0		&&	68.3	$\pm$	1.1	&	68.7	$\pm$	1.0	\\ 
Birds	&\textbf{	73.2	$\pm$	0.9	}&	68.9	$\pm$	1.1		&&\textbf{	62.6	$\pm$	1.1	}&	61.1	$\pm$	1.1	\\ 
Textures	&\textbf{	75.8	$\pm$	0.8	}&	74.6	$\pm$	0.8		&&\textbf{	78.4	$\pm$	0.7	}&	77.3	$\pm$	0.7	\\ 
Quick Draw 	&\textbf{	66.4	$\pm$	0.9	}&	64.9	$\pm$	0.9		&&	67.2	$\pm$	0.9	&\textbf{	69.5	$\pm$	0.8	}\\ 
Fungi	&\textbf{	43.9	$\pm$	1.2	}&	43.0	$\pm$	1.1		&&	39.7	$\pm$	1.1	&\textbf{	42.0	$\pm$	1.1	}\\ 
VGG Flower	&	89.9	$\pm$	0.6	&	89.6	$\pm$	0.7		&&	88.5	$\pm$	0.6	&\textbf{	89.4	$\pm$	0.7	}\\ 
Traffic Sign	&\textbf{	80.0	$\pm$	1.0	}&	77.9	$\pm$	1.0		&&	74.4	$\pm$	1.2	&\textbf{	84.3	$\pm$	0.9	}\\ 
MSCOCO	&\textbf{	52.5	$\pm$	1.1	}&	48.6	$\pm$	1.1		&&\textbf{	54.8	$\pm$	1.1	}&	53.4	$\pm$	1.1	\\ \midrule
Average (MD)	&\textbf{	69.9			}&	68.7				&&	67.4			&\textbf{	69.4			}\\ \midrule
MNIST	&\textbf{	93.7	$\pm$	0.6	}&	92.7	$\pm$	0.6		&&	94.0	$\pm$	0.6	&\textbf{	94.6	$\pm$	0.5	}\\ 
CIFAR-10	&\textbf{	77.2	$\pm$	0.8	}&	68.2	$\pm$	0.8		&&\textbf{	80.3	$\pm$	0.8	}&	75.3	$\pm$	0.8	\\ 
CIFAR-100	&\textbf{	67.7	$\pm$	1.0	}&	58.3	$\pm$	1.1		&&\textbf{	68.7	$\pm$	1.0	}&	66.2	$\pm$	1.0	\\ 
CropDisease	&	81.8	$\pm$	0.8	&\textbf{	84.2	$\pm$	0.8	}	&&	78.8	$\pm$	1.0	&\textbf{	85.0	$\pm$	0.8	}\\ 
EuroSAT	&\textbf{	89.3	$\pm$	0.6	}&	88.4	$\pm$	0.6		&&	89.4	$\pm$	0.6	&\textbf{	90.5	$\pm$	0.5	}\\ 
ISIC	&\textbf{	45.8	$\pm$	0.9	}&	42.1	$\pm$	0.8		&&\textbf{	45.1	$\pm$	0.9	}&	44.2	$\pm$	0.8	\\ 
ChestX	&\textbf{	26.1	$\pm$	0.5	}&	24.4	$\pm$	0.5		&&	24.2	$\pm$	0.5	&\textbf{	25.0	$\pm$	0.5	}\\ 
Food101	&\textbf{	48.8	$\pm$	1.2	}&	47.2	$\pm$	1.1		&&	42.9	$\pm$	1.2	&\textbf{	45.1	$\pm$	1.1	}\\ \midrule
Average (MD+)	&\textbf{	68.2			}&	66.1				&&	66.5			&\textbf{	67.7			}\\ 
\bottomrule 
\end{NiceTabular}
\caption{Comparison of accuracy ($\pm$95\% confidence interval) between TSA and the MIV-head (Ours) based on all non-ILSVRC datasets in extended MD benchmark, and based on the same ResNet-18 backbones. The row labeled ``Average (MD)'' indicates the average accuracy across the nine original (non-ILSVRC) MD datasets whereas the row labeled ``Average (MD+)'' is the average across the total 17 extended MD datasets. We also conducted a two-sided paired t-test between TSA and Ours, and bolded the higher accuracy when the p-value of the t-test is $<0.01$.} \label{tab:rn18}
\end{table}

We also used lower input resolution ($84\times84$) based on ResNet-18 backbones, another popular setting in FSC, in our experiments. As demonstrated in Table~\ref{tab:rn18}, the results for an off-the-shelf ResNet-18 (pretrained on ILSVRC-2012) backbone exhibit a similar pattern to that in Table~\ref{tab:miv_acc_tsa}---the MIV-head outperforms TSA on average and in most of the test datasets. Nevertheless, when we adopt ``SDL-ResNet-18''~\citep{url21}, another ResNet-18 backbone specially pretrained using a carefully-designed meta-training procedure on a subset of ILSVRC, the results show the opposite, i.e., TSA's accuracy is mostly better than that of the MIV-head. This shows the impact of special-purpose vs. off-the-shelf backbones on adaptation performance of different approaches.

\section{Further qualitative illustrations using t-SNE visualization of embeddings} \label{app:tsne}

\begin{figure}[t]
    \centering
     \begin{subfigure}[t]{0.66\linewidth}
          \includegraphics[width=0.49\textwidth]{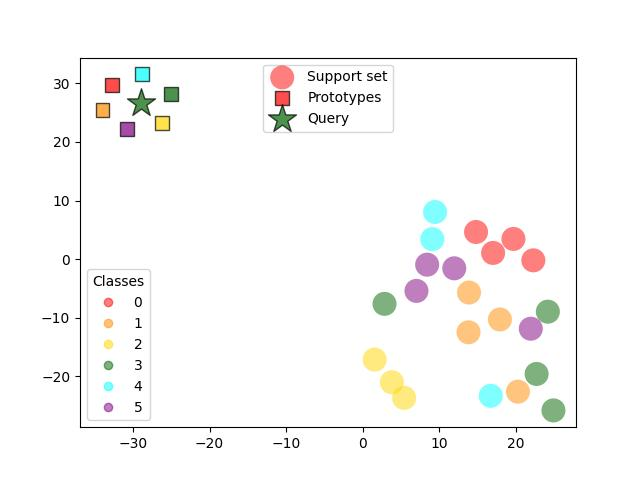}
          \includegraphics[width=0.49\textwidth]{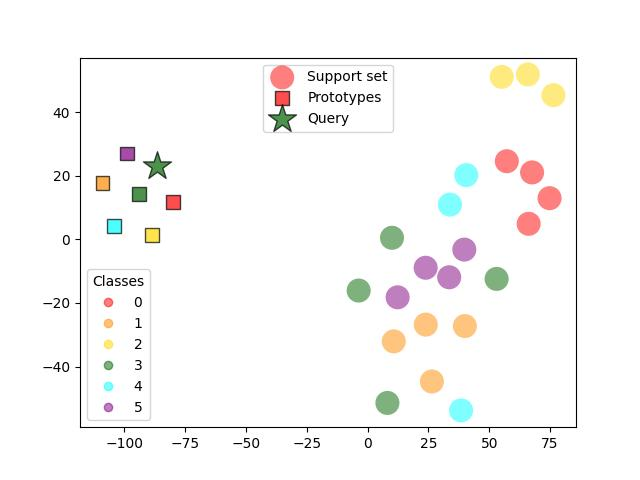}
        \caption{MIV-head (Aircraft)}\label{fig:tsne_miv_a2}
     \end{subfigure}
     \hfill
     \begin{subfigure}[t]{0.33\linewidth}
        \includegraphics[width=\textwidth]{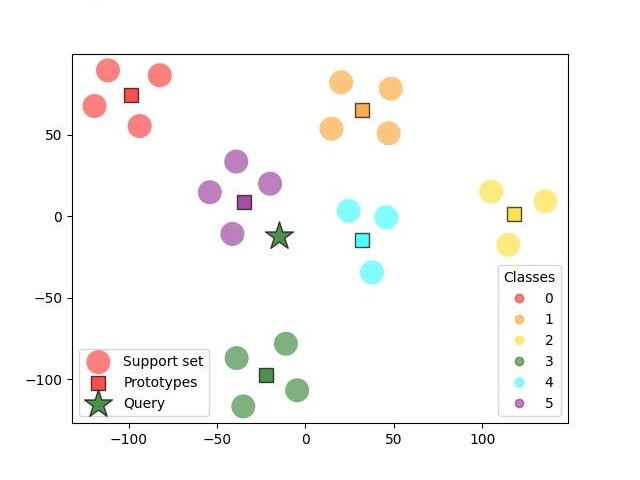}
        \caption{TSA (Aircraft)}\label{fig:tsne_tsa_a2}
     \end{subfigure}
     \hfill
     \begin{subfigure}[t]{0.66\linewidth}
          \includegraphics[width=0.49\textwidth]{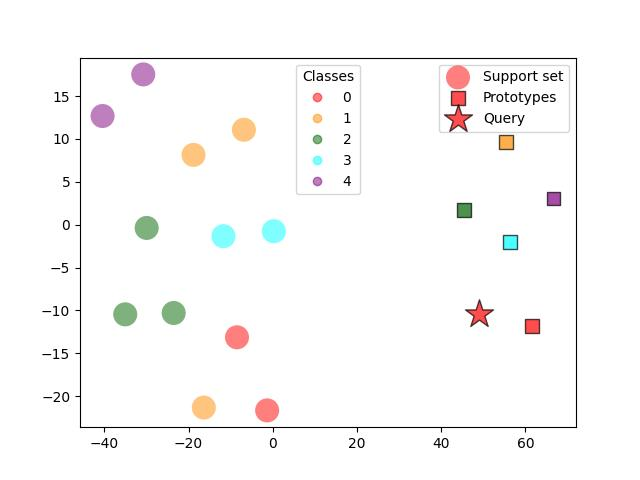}
          \includegraphics[width=0.49\textwidth]{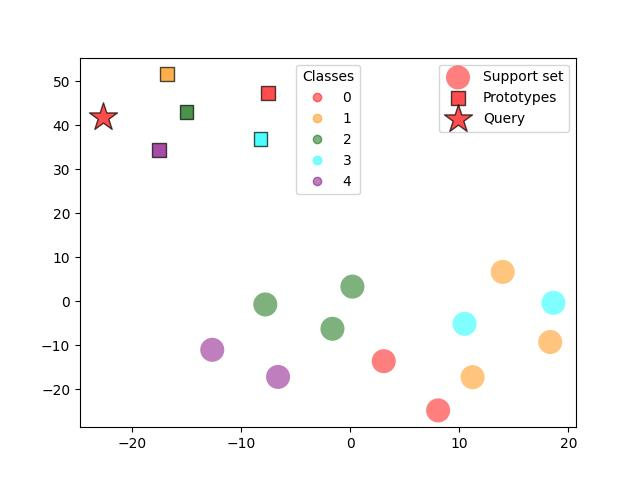}
        \caption{MIV-head (Omniglot)}\label{fig:tsne_miv_o1}
     \end{subfigure}
     \hfill
        \begin{subfigure}[t]{0.33\linewidth}
        \includegraphics[width=\textwidth]{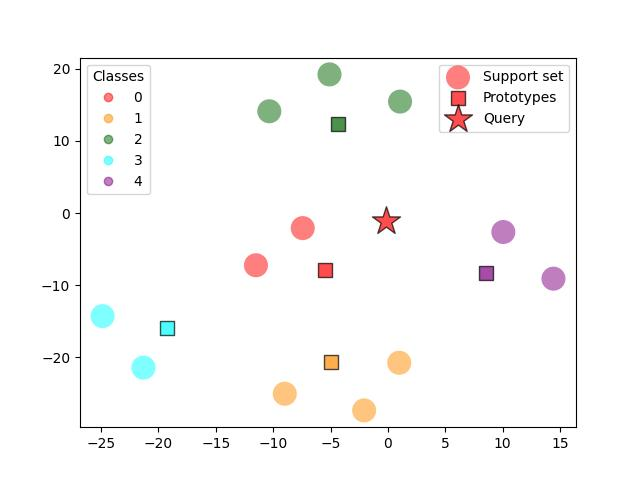}
        \caption{TSA (Omniglot)}\label{fig:tsne_tsa_o1}
     \end{subfigure}
     \hfill
     \begin{subfigure}[t]{0.66\linewidth}
          \includegraphics[width=0.49\textwidth]{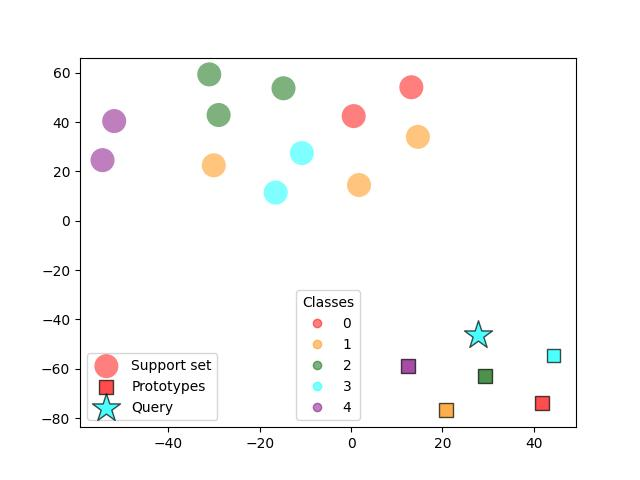}
          \includegraphics[width=0.49\textwidth]{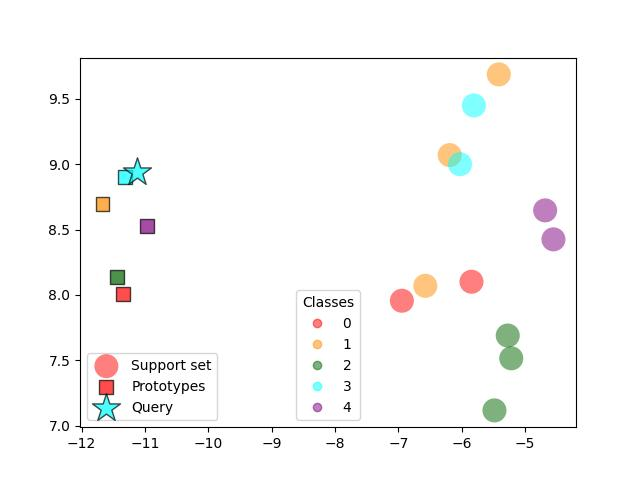}
          \caption{MIV-head (Omniglot)}\label{fig:tsne_miv_o2}
     \end{subfigure}
     \hfill
     \begin{subfigure}[t]{0.33\linewidth}
        \includegraphics[width=\textwidth]{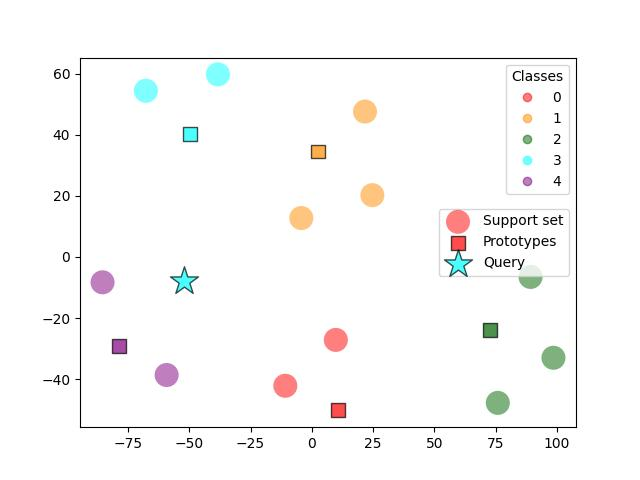}
        \caption{TSA (Omniglot)}\label{fig:tsne_tsa_o2}
     \end{subfigure}
\caption{Embedding visualizations with t-SNE of the support set (circles), prototype (squares) and query (star) produced by the MIV-head vs. TSA, based on ResNet-50 backbone and the same task from the Aircraft and Omniglot datasets. 
For the MIV-head, embeddings from the last and second-to-last blocks of the backbone are visualized in the left and right panels, respectively, in~\ref{fig:tsne_miv_a2},~\ref{fig:tsne_miv_o1},~\ref{fig:tsne_miv_o2}. Embeddings produced by TSA are visualized in~\ref{fig:tsne_tsa_a2},~\ref{fig:tsne_tsa_o1},~\ref{fig:tsne_tsa_o2}. All embeddings are colored according to their ground-truth class labels, with colors specified by the legend. Best viewed in color.}\label{fig:tsne_more}
\end{figure}

Figure~\ref{fig:tsne_more} plots more t-SNE visualizations of the embeddings of support set, prototype, and query, produced by the MIV-head and TSA. They are based on the same test episode of the Aircraft or Omniglot dataset, and a ResNet-50 backbone similar to Figure~\ref{fig:tsne_full} (Section~\ref{sec:intro}). The embeddings of the support set created by adapter methods like TSA are well-clustered based on their ground-truth class labels. Consequently, the centroids of clusters (or classes) can be used as prototypes to classify the query, see Figures~\ref{fig:tsne_tsa_a1}, \ref{fig:tsne_tsa_a2}, \ref{fig:tsne_tsa_o1}, \ref{fig:tsne_tsa_o2}. 

In stark contrast to TSA, given the ``frozen'' embeddings retrieved from the black-box backbone,\footnote{To produce visualizations in Figures~\ref{fig:tsne_miv_a1},~\ref{fig:tsne_miv_a2},~\ref{fig:tsne_miv_o1},~\ref{fig:tsne_miv_o2}, the embeddings from the black-box backbone were processed by the patch-level ``pooling-by-attention'' mechanism of the MIV-head.} the support-set embeddings are less useful for classification---it is clear, from Figures~\ref{fig:tsne_miv_a1},~\ref{fig:tsne_miv_a2},~\ref{fig:tsne_miv_o1},~\ref{fig:tsne_miv_o2}, that they are less clustered w.r.t. their ground-truth classes, whether retrieved from the last or second-to-last block of the backbone (\emph{cf.} left and right panels). However, CAP in the MIV-head creates the prototype of each class as a bag representation induced by the query, which is less impacted by the quality of the support-set embeddings. Indeed, as illustrated by the visualizations of the MIV-head's embeddings, all prototypes induced by a query are ``projected'' close to the query, rather than the support set. CAP uses the support set to learn how to effectively pull together or push apart embeddings of the prototypes relative to the query, based on their ground-truth class labels. Figures~\ref{fig:tsne_miv_a1},~\ref{fig:tsne_miv_a2},~\ref{fig:tsne_miv_o1},~\ref{fig:tsne_miv_o2} demonstrate that such transformations by CAP frequently succeed in pulling the prototype of the ``query class'' \emph{closest} to the query, leading to improved classification in spite of the low-quality embeddings of the support set.

Note that embeddings created from different blocks of the backbone (in this case the last and second-to-last blocks) compete to generate logits through the $logsumexp$ function. Therefore, the classification is usually dominated by the block whose prototypes are closer to the query---a situation illustrated in Figures~\ref{fig:tsne_miv_o1} and ~\ref{fig:tsne_miv_o2}.

\end{appendices}
\clearpage
\bibliography{references}
\end{document}